\documentclass{article}
\usepackage{arxiv}
\usepackage[utf8]{inputenc}
\usepackage[T1]{fontenc}
\usepackage{hyperref}
\usepackage{url}
\usepackage{booktabs}
\usepackage{amsfonts}
\usepackage{nicefrac}
\usepackage{microtype}
\usepackage{xcolor}
\usepackage{amsmath, mathtools}
\usepackage{graphicx, subcaption}
\usepackage{multirow, rotating}
\usepackage{chngcntr}
\usepackage{algorithm, algpseudocode}
\usepackage{amsthm}
\usepackage{tocloft, titletoc}

\newtheorem{theorem}{Theorem}

\newtheorem{corollary}{Corollary}

\newtheorem{remark}{Remark}
\newtheorem{assumption}{Assumption}

\usepackage{amsopn}
\DeclareMathOperator{\p}{p}
\DeclareMathOperator{\Data}{Data}
\DeclareMathOperator{\Res}{Res}
\DeclareMathOperator{\IC}{IC}
\DeclareMathOperator{\SI}{SI}
\DeclareMathOperator{\LS}{LS}
\DeclareMathOperator{\tra}{tra}
\DeclareMathOperator{\Split}{Split}

\usepackage{array}
\usepackage{tabularx}
\newcolumntype{Y}{>{\raggedright\arraybackslash}X} 
\usepackage{algorithm}

\algrenewcommand\alglinenumber[1]{}
\newcommand{\algindent}{\hspace*{1em}}
\newcommand{\algindenta}{\hspace*{2em}}

\title{A Physics-Informed Neural Networks-Based Model Predictive Control Framework for $SIR$ Epidemics}

\author{
	Aiping Zhong \\
	\textit{School of Automation Science and Technology} \\
	\textit{South China University of Technology, Guangzhou, China} \\
	\And
	Baike She \\
	\textit{School of Electrical and Computer Engineering} \\
	\textit{Georgia Institute of Technology, Atlanta, GA 30332 USA} \\
	\And
	Philip E. Paré\thanks{Corresponding author: \href{mailto:philpare@purdue.edu}{philpare@purdue.edu}. Funding: This work was partially supported by US National Science Foundation (NSF--ECCS \#2238388).} \\
	\textit{Elmore Family School of Electrical and Computer Engineering} \\
	\textit{Purdue University, West Lafayette, IN 47907 USA}\\
        \textit{philpare@purdue.edu}
}

\date{}
\makeatletter
\renewcommand{\@date}{}      
\makeatother

\begin{document}

\twocolumn[
\maketitle
\begin{abstract}
This work introduces a physics-informed neural networks (PINNs)-based model predictive control (MPC) framework for susceptible-infected-recovered ($SIR$) spreading models. Existing studies in MPC design for epidemic control often assume either 1) measurable states of the dynamics, where the parameters are learned, or 2) known parameters of the model, where the states are learned. In this work, we address the joint real-time estimation of states and parameters within the MPC framework using only noisy infected states, under the assumption that 1) only the recovery rate is known, or 2) only the basic reproduction number is known. 
Under the first assumption, we propose MPC-PINNs and two novel PINNs algorithms, all of which are integrated into the MPC framework. First, we introduce MPC-PINNs, which are designed for $SIR$ models with control. We then propose log-scaled PINNs (MPC-LS-PINNs), which incorporate a log-scaled loss function to improve robustness against noise. Next, we present split-integral PINNs (MPC-SI-PINNs), which leverage integral operators and state coupling in the neural network training process to effectively reconstruct the complete epidemic state information. 
Building upon these methods, we further extend our framework for the second assumption. We establish the necessary conditions and extend our PINNs algorithms, where MPC-SI-PINNs are simplified as split-PINNs (MPC-S-PINNs). By incorporating these algorithms into the MPC framework, we simultaneously estimate the epidemic states and parameters while generating optimal control strategies. Experiment results demonstrate the effectiveness of the proposed methods under different settings.
\end{abstract}
]  

\section{Introduction}

\label{Introduction}
Model predictive control (MPC) excels in real-world control applications~\cite{grune2017nonlinear, kouvaritakis2016model}. It solves optimal control problems iteratively, balancing future performance with constraints. MPC's robustness to model and parameter uncertainty makes it valuable for infectious disease control~\cite{wang2023model, kohler2021robust, carli2020model, sereno2023switched, sauerteig2022model, she2022learning, morato2020optimal, peni2021convex, liu2021use}. 
For example,~\cite{carli2020model} applies MPC to devise multi-region control strategies. Similarly,~\cite{sereno2023switched} manages infection peaks with switched nonlinear MPC. However, most studies rely on having either fully measurable states to estimate parameters or complete parameter information to estimate states in order to solve the MPC problem~\cite{wang2023model, kohler2021robust, carli2020model, sereno2023switched, sauerteig2022model, she2022learning, morato2020optimal, peni2021convex, liu2021use}.

We categorize the parameter and state estimation methods for infectious disease spread into three types. The first estimates unknown parameters using fully measurable states~\cite{she2022learning, hadeler2011parameter, chowell2017fitting, magal2018parameter, samsuzzoha2013parameter}. For instance,~\cite{she2022learning} uses generalized least squares  in an MPC framework to estimate transmission and recovery rates and design controls concurrently. 
The second estimates unmeasured states assuming fully known parameters~\cite{niazi2022observer, mehra2019observer, ibeas2015stability, alonso2012observer, azimi2022state, degue2018interval}. For example,~\cite{azimi2022state} applies a variant of the extended Kalman filter to estimate the epidemic states. Likewise,~\cite{mehra2019observer} estimates states from infected states for sliding mode control. The third reconstructs states and parameters from partially observed states and parameters~\cite{hota2021closed, hespanha2021forecasting, blima2018class, rajaei2021state, gomez2021monitoring, capaldi2012parameter}. For instance,~\cite{hespanha2021forecasting} uses confirmed case and death data for a stochastic $SIR$ model to estimate the parameters and states.
However, none of these methods can both 1) simultaneously estimate the states and parameters of the spreading model using only partially observed data, and 2) integrate this process into an MPC framework.


To address these two challenges, we propose a physics-informed neural networks (PINNs)-based MPC framework. We demonstrate the framework on an $SIR$ model in which only noisy infected states are available. Conventional model estimation methods such as least squares~\cite{she2022learning, chowell2017fitting, samsuzzoha2013parameter, capaldi2012parameter} and extended Kalman filter~\cite{azimi2022state, rajaei2021state, gomez2021monitoring} often rely on prior knowledge of the noise characteristics in the observed data, such as the covariance matrix. In contrast, PINNs do not require such explicit noise modeling~\cite{raissi2019physics}.
Further,
PINNs excel in solving inverse problems for ordinary differential equations (ODEs)~\cite{raissi2019physics}, positioning them as a promising tool for solving the parameter and state estimation simultaneously. 
Unlike traditional deep learning methods for spreading models~\cite{chimmula2020time, qin2019data, arora2020prediction, kirbacs2020comparative, moftakhar2020exponentially, ayyoubzadeh2020predicting, tomar2020prediction, kolozsvari2021predicting, ibrahim2021variational}, which demand extensive training data without considering the laws of physics, PINNs embed physical constraints in the training process~\cite{raissi2019physics}. This property enhances interpretability and substantially reduces the amount of data, making it well-suited for infectious disease modeling where data are limited~\cite{grimm2022estimating, han2024approaching, shaier2022data, kim2025physics, long2021identification, kharazmi2021identifiability, schiassi2021physics, rodriguez2023einns, millevoi2024physics}. The study~\cite{kharazmi2021identifiability} examines how parameter uncertainty and model structure affect predictions with an integer- or fractional-order PINN, while~\cite{kim2025physics} proves that variable scaling improves PINN training via neural tangent kernel (NTK).
However, existing techniques for PINNs rely on full state observations to estimate parameters, except for~\cite{rodriguez2023einns, millevoi2024physics}. 
The work~\cite{millevoi2024physics} proposes a split PINN framework that separates the data regression process from physical training, using only infected states to reconstruct both the states and parameters of $SIR$ models. However, this approach fails to learn the states and parameters when integrated into an MPC framework where external control inputs are introduced. 

In our proposed PINNs-based MPC framework, PINNs are employed to infer the unknown states and parameters from noisy infected states at each MPC iteration. Integrating PINNs into an MPC framework introduces distinct challenges. Studies such as~\cite{krishnapriyan2021characterizing, wang2022and} indicate that the high frequencies in ODE models, such as steep gradients or rapid oscillations, may disrupt the regularization inherent in the {ODEs' residual term~\cite{raissi2019physics}, potentially weakening robustness to noise. Incorporating external control inputs, especially when they exhibit periodic or stepwise changes due to practical constraints and the MPC design, introduces high-frequency components into the spreading models. Furthermore, using MPC-derived control inputs during PINNs training creates a coupled dependency between the PINNs and the MPC components.

Thus, we aim to develop a novel PINN approach that simultaneously estimates the states and parameters from noisy infected states and integrates these estimates into an MPC framework for control design in $SIR$ models. Assuming access to noisy infected states and either 1) the recovery rate or 2) the basic reproduction number of the $SIR$ model, the contributions of this paper are as follows:
\begin{itemize}
    \item Assuming the recovery rate is known, we develop  a set of single-input and single-output neural networks (NNs) for the $SIR$ models with control inputs, called MPC-PINNs. This design enables simultaneous modeling of the $SIR$ dynamics, estimation of the transmission rate and states, and solving the optimal control problem through the MPC framework.
    \item Assuming the recovery rate is known, we design a PINNs-based MPC framework where PINNs provide the estimated states and transmission rate for the MPC. The MPC computes a control strategy for spreading dynamics. The $SIR$ models generate noisy infected states fed back to the PINNs, creating a closed-loop system with iterative interactions.
    \item Assuming the recovery rate is known, we introduce a log-scaled PINNs (MPC-LS-PINNs) algorithm that incorporates a log-relative error term in the loss function to improve the robustness of the proposed PINNs against substantial noise in the infected states.
    \item Assuming the recovery rate is known, we propose a split-integral PINNs (MPC-SI-PINNs) algorithm that decouples data regression from physics-informed training and enhances the latter using integral-derived state estimates.
    \item Without knowing any parameters, we show analytically that, given the basic reproduction number, we can uniquely identify the states and parameters of the $SIR$ dynamics using infected states. Based on this condition, we adapt our proposed PINNs and MPC-LS-PINNs, and modify the MPC-SI-PINNs into a split-PINNs (MPC-S-PINNs) algorithm for use within the MPC framework.
\end{itemize}
The remainder of this paper is structured as follows. In Sec.~\ref{sec:preliminaries}, we introduce the preliminaries and formulate the problems, including the $SIR$ models and the MPC framework. Sec.~\ref{Sec3} proposes a PINN-based MPC framework and further introduces PINNs algorithms. Sec.~\ref{results} provides experimental results, and Sec.~\ref{section5} concludes with future directions.
\section{PRELIMINARIES AND OBJECTIVES} \label{sec:preliminaries}
In this section, we first introduce the epidemic spreading model. We then propose an MPC framework. We also outline the problems addressed in this work.
\subsection{$SIR$ MODELS WITH CONTROL} \label{Sec_SIR_Model}
We consider the following closed-loop susceptible-infected-recovered ($SIR$) models:
\begin{subequations} \label{1}
\begin{align}
\dot{S}(t) &= -\beta I(t) S(t) \label{1a} \\
\dot{I}(t) &= \beta I(t) S(t) - (\gamma + u(t)) I(t) \label{1b} \\
\dot{R}(t) &= (\gamma + u(t)) I(t), \label{1c}
\end{align}
\end{subequations}
where $S(t)$, $I(t)$, and $R(t) \in [0,1]$ are susceptible, infected, and recovered proportions at time $t$, respectively, with  $S(t) + I(t) + R(t) = 1$, for all $t \geq 0$; $\beta \in (0,1)$ and $\gamma \in (0,1)$ denote the transmission and recovery rates, respectively; and the control $u(t) \in [0,1]$ reflects the intensity of the intervention measures aimed at removing infected individuals from the transmission chain. For $SIR$ models without any control input, the herd immunity point $S^\star = \frac{\gamma}{\beta}$ acts as a threshold condition, which is the reciprocal of the basic reproduction number $\mathcal{R}_0 = \frac{\beta} {\gamma}$. When the susceptible proportion falls below $S^\star$, the infected proportion $I(t)$ converges exponentially to the disease-free equilibrium~\cite{hethcote2000mathematics}. Consider the situation where we set the initial conditions to
\begin{equation}
S(0) = 1 - I(0), R(0) = 0, \label{initial2}
\end{equation}
assuming $R(0) = 0$ at the very early stage of the spreading process. Moreover, we denote the noisy infected states as $\tilde{I}(t)\neq I(t)$, which follows
\begin{equation}
\tilde{I}(t)= \frac{\text{Pois}(I(t)\times \kappa N)}{\kappa N}.  \label{Poisson}
\end{equation}
$\text{Pois}(I(t) \times \kappa N)$ is a Poisson distribution with mean $I(t) \times \kappa N$, where $N$ is the total population and $\kappa$ is a noise regulation factor that adjusts the noise intensity relative to the true values. This formulation is consistent with the Poisson noise modeling adopted in \cite{millevoi2024physics}. The construction in~\eqref{Poisson} yields an expectation equal to $I(t)$ and variance $\frac{I(t)}{\kappa N}$ so that smaller $\kappa$ results in larger noise. After introducing the $SIR$ spreading models, we give the following two assumptions.


\begin{assumption}
The initial conditions given in \eqref{initial2}, observed infected states $\tilde{I}(t)$ defined in \eqref{Poisson}, and the recovery rate $\gamma$ are known. \label{assumption1} 
\end{assumption}
\begin{assumption}
The initial conditions given in \eqref{initial2}, observed infected states $\tilde{I}(t)$ defined in \eqref{Poisson}, and the basic reproduction number $\mathcal{R}_0=\frac{\beta}{\gamma}$ are known. \label{assumption2} 
\end{assumption}
Regarding Assumptions \ref{assumption1} and \ref{assumption2}, the observed infected states $\tilde{I}(t)$ can be acquired from publicly available epidemiological surveillance data. The initial condition \eqref{initial2} can be approximated based on reported cases at the onset of the epidemic. Furthermore, the recovery rate $\gamma$ can be obtained from the average recovery time from the virus. The basic reproduction number $\mathcal{R}_0$ arises from epidemiological studies.  
Under Assumption \ref{assumption1} and Assumption \ref{assumption2}, we design a PINNs-based MPC framework that leverages noisy infected states $\tilde{I}(t)$ to iteratively 1) learn the transmission rate (and recovery rate), 2) estimate the susceptible, infected, and recovered states, and 3) determine the control input $u(t)$ to steer the infectious spreading dynamics toward the herd immunity threshold while simultaneously satisfying the control objectives.
\subsection{AN MPC FRAMEWORK} \label{subsection C}
We first present an MPC framework for $SIR$ models.
We discretize the continuous-time system in~\eqref{1} using Euler's method with a sampling period $T_s= 1$ day, to match the frequency of daily disease tracking. We then consider $x_{k+1} = f(x_k, u_k)$ at each time step $k$, where
\begin{equation}
    f(x_k, u_k) = \begin{bmatrix}
        S_{k+1}\\
        I_{k+1}\\
        R_{k+1}\\
    \end{bmatrix}=
    \begin{bmatrix}
        (-\beta I_k+1)S_k\\
       (\beta S_k -(\gamma + u_k)+1 )I_k\\
        (\gamma + u_k)I_k+R_k\\
    \end{bmatrix}, \label{f()}
\end{equation} for all $k\in\mathbb{N}_{>0}$; $x_k\coloneqq x(kT_s) = [S_k, I_k, R_k]^\top$ is the state vector; and $u_k \coloneqq u(kT_s)$ is the control input during the $k^{\text{th}}$ sampling interval. Hence, the optimization problem becomes 
\begin{subequations}
\label{Eq_MPC_OP}
\begin{equation}
\begin{aligned}
\min_{\mathbf{u}_k\in[0,u_{\max}],\, I_k\in [0, I^\star_{\max}]} \mathcal{J}(\mathbf{x}_k, \mathbf{u}_k) 
= \\\sum_{j=k}^{k+N_{\p}-1} \left(q_1 \|S_{j+1} - S^\star \|^2 \right. 
&\left. +\; q_2  \|u_j\|^2 \right),
\end{aligned}
\end{equation}
\begin{equation}
\text{s.t.}\,x_{j+1} = f(x_j, u_j), j = k, \cdots, k+N_{\p} -1,\label{prediction_model}
\end{equation}
\begin{equation}
u_{j} \in [0,u_{\max}],  x_{j} \in [0,1]^3,  j = k, \cdots, k+N_{\p}-1,
\end{equation}\end{subequations}
where the cost function is defined as $\mathcal{J}(\mathbf{x}_k, \mathbf{u}_k)$, the prediction horizon spans $N_{\p}\in\mathbb{N}_{>0}$ steps, $\mathbf{x}_k = [x_k, \dots, x_{k + N_{\p} - 1}]^\top$ is the state sequence, and $\mathbf{u}_k = [u_k, \dots, u_{k + N_{\p} - 1}]^\top$ is the control sequence over the $k^{\text{th}}$ prediction horizon. 
Note that $q_1$ and $q_2$, both non-negative, balance the trade-off between hitting the herd immunity target and cutting the control costs. The infection peak is capped by $I^\star_{\max}$, and the control input $u$ is restricted to $[0, u_{\max}]$, reflecting resource constraints.
Over the $k^{\text{th}}$ prediction horizon $[k, k + N_{\p} - 1]$, MPC computes an optimal control strategy by solving the optimization problem in~\eqref{Eq_MPC_OP}. It steers the spreading process toward the herd immunity point, measured by $\|S_{j+1}-S^{\star}\|^2$ while ensuring that the peak infection level remains below $I^\star_{\max}$. The strategy also respects input constraints and minimizes economic costs, which are quantified by $\|u_j\|^2$.

Solving the optimization problem in~\eqref{Eq_MPC_OP} requires knowledge of the system dynamics (transmission rate $\beta$ and recovery rate $\gamma$) and the full state information (susceptible, infected, and recovered states) at each iteration. However, Sec.~\ref{Sec_SIR_Model} states that only the noisy infected states $\tilde{I}(t)$ and recovery rate $\gamma$ (or basic reproduction number $\mathcal{R}_0$) are available. Thus, we integrate PINNs into the MPC framework, which estimate $\beta$ (and $\gamma$) and all the states, in order to solve~\eqref{Eq_MPC_OP}.

\subsection{OBJECTIVES}
\label{problem formulation}
After introducing the $SIR$ dynamics with control inputs in~\eqref{1} and an MPC framework utilizing the estimated states and parameters as inputs, we address the following challenges in this work:
\begin{enumerate}
\item What is an appropriate way to design the neural network architecture for integrating the MPC control input $u(t)$ into the PINNs framework for joint training? How can we define PINNs within the proposed architecture? \label{prb1}
\item Under Assumption \ref{assumption1}, how are the PINNs constructed within the MPC framework to estimate the full states $S,I,R$, and the transmission rate $\beta$, for the optimization problem in~\eqref{Eq_MPC_OP}?\label{prb2} 
\item Under Assumption \ref{assumption1} and the proposed PINNs, how can robustness be enhanced when using noisy infected states $\tilde{I}$, especially in the presence of significant measurement noise?\label{prb3}
\item Under Assumption \ref{assumption1}, how can we refine the implementation of the proposed  PINNs by exploiting the couplings among the $S$, $I$, and $R$ states to achieve faster convergence during training and enhance the overall performance? \label{prb4}
\item Under Assumption \ref{assumption2}, what additional conditions are necessary for estimating the full states  $S,I,R$ and parameters $\beta,\gamma$, and how should the proposed PINNs be adapted accordingly?\label{prb5}
\end{enumerate}
We address the above challenges in the remainder of this paper: Sec.~\ref{Sec3.1} addresses Problem~\ref{prb1}; Sec.~\ref{Sec3.2} addresses Problem~\ref{prb2}; Sec.~\ref{Sec4} addresses Problem~\ref{prb3}; Sec.~\ref{Sec5} addresses Problem~\ref{prb4}; and Sec.~\ref{Sec6} addresses Problem~\ref{prb5}. To validate the effectiveness of our methods, we compare them with proposed PINNs in Sec.~\ref{Sec3.2} as a baseline and demonstrate significant improvements via experiments in Sec.~\ref{results}.

\section{A PINNS-BASED MPC FRAMEWORK}  
\label{Sec3}
In this section, we develop a PINNs-based MPC framework for the 
$SIR$ models in~\eqref{1}. We first introduce the neural network architecture for joint state–parameter estimation. Next, we incorporate the external control input in the case where the transmission rate is known but the recovery rate is unknown. We then present two enhanced algorithms, one designed to improve noise robustness and the other to increase training efficiency. Finally, we extend these methods to cases where both the transmission and recovery rates are unknown and present the generalized approach.

\subsection{NEURAL NETWORK ARCHITECTURE} \label{Sec3.1}
As a prerequisite for constructing our PINNs, we integrate a feedforward neural network, defined as
\begin{equation}
    \Sigma^{(m)}(\mathbf{z}^{(m)}) = \Lambda^{(m)}\left(\mathbf{W}^{(m)} \mathbf{z}^{(m)} + \mathbf{b}^{(m)}\right) ,
    \label{one NN}
\end{equation}
where $\Sigma^{(m)}(\cdot):\mathbb{R}^{n_{m-1}} \to \mathbb{R}^{n_m}$ and $\Lambda^{(m)}(\cdot): \mathbb{R}^{n_m} \to \mathbb{R}^{n_m}$ are the mapping and activation functions of the $m^{th}$ layer, respectively, with $n_m \in\mathbb{N}_{>0}$ neurons; the network has a total of $M\in\mathbb{N}_{>0}$ hidden layers; the weight matrix and bias vector are represented by $\mathbf{W}^{(m)} \in \mathbb{R}^{n_m \times n_{m-1}}$ and $\mathbf{b}^{(m)} \in \mathbb{R}^{n_m}$, respectively; the input $\mathbf{z}^{(m)}$ is transformed recursively across layers.

Inspired by \cite{millevoi2024physics}, we introduce five separate SISO neural networks. We denote these networks as $NN_{\hat{S}}$, $NN_{\hat{I}}$, $NN_{\hat{u}}$, $NN_{\hat \beta}$, and $NN_{\hat \gamma}$. Each network $NN_{\hat{v}}$ takes time $t$ as its sole input with $v\in\{S,  I, \beta, \gamma, u\}$. The outputs of these networks are denoted as $\hat{S}$, $\hat{I}$, $\hat{\beta}$, $\hat{\gamma}$ and $\hat{u}$. 
Under Assumption~\ref{assumption1},
we leverage four neural networks $NN_{\hat{v}}$, where $v\in\{S,  I, \beta, u\}$, to generate their corresponding outputs, and
the estimated recovered state $\hat{R}$ is then computed as ${\hat{R}}= 1 -{\hat{S}} - {\hat{I}}$. Under Assumption~\ref{assumption2}, 
we leverage a different set of four neural networks $NN_{\hat{v}}$, with $v\in\{S,  I, \gamma, u\}$, to generate their corresponding estimations, where
the estimated transmission rate $\hat{\beta}$ is derived as ${\hat{\beta}} = \mathcal{R}_0{\hat{\gamma}}$. Based on \eqref{one NN}, for any scalar function $NN_{\hat{v}}(t)$, the final layer of the neural network is designed as a single-node output with one-dimensional input time $t$, given by
\begin{equation}
NN_{\hat{v}}(t) = \Sigma^{(M+1)} \circ \Sigma^{(M)} \circ \cdots \circ \Sigma^{(1)}(t).    \label{NN}
\end{equation}

\begin{remark}
Our separate SISO neural network architecture is tailored to the challenges of epidemic modeling and control for the following reasons:  
1) Most related PINN-based studies adopt a single-input, multi-output (SIMO) neural network~\cite{han2024approaching, shaier2022data, kim2025physics, long2021identification, kharazmi2021identifiability, schiassi2021physics, rodriguez2023einns}, where each output corresponds to one estimation. In contrast, we assign four separate SISO networks to estimate $\hat{S}$, $\hat{I}$, $\hat{\beta}$, and $\hat{u}$. This modular design improves training stability and computational efficiency, demonstrated by the experimental results in Sec.~\ref{results}. It also supports the MPC-SI-PINNs framework in Sec.~\ref{Sec5}, where the data regression step directly trains $NN_{\hat{I}}$ and $NN_{\hat{u}}$.
2) Instead of treating the implemented control $u$ as an external input to $NN_{\hat{v}}$ along with time $t$, we design a separate neural network $NN_{\hat{u}}$ with input $t$ only. This design avoids generalization issues caused by a two-dimensional input, i.e., $t$ and $u$, where the sparsity of control data trajectories may hinder effective training. It also enables $NN_{\hat{u}}$ to participate in joint training through the backpropagation of gradients from the PINNs loss. Experimental results in the supporting material further illustrate the advantages of this architecture.
\end{remark}




\subsection{PHYSICS-INFORMED NEURAL NETWORKS}\label{Sec3.2}
In this section, we develop the PINNs under Assumption~\ref{assumption1} and designate the resulting framework as \textit{MPC-PINNs}. The multi-objective nature of PINNs requires a well-designed loss function. Using the outputs of the neural networks $NN_{\hat{S}}$, $NN_{\hat{I}}$, $NN_{\hat{u}}$, and $NN_{\hat{\beta}}$ with input~$t$, we define the total loss for MPC-PINNs as 
\begin{equation}
\begin{aligned}
   \text{Loss}({\hat{S}}, {\hat{I}}, {\hat{u}}, {\hat{\beta}},\tilde{I}_i,u_j) = \text{Loss}_{\Data} + 
   \text{Loss}_{\Res} + \text{Loss}_{\IC}.
\end{aligned}
 \label{PINN_LOSS}
\end{equation}We further explain these terms in detail. Starting from time step $0$ to $k \in \mathbb{N}_{>0}$, we have the total number of $k + 1$ noisy observed infected states $\tilde{I}_i \coloneqq \tilde{I}(i)$, $i = 0, 1, \dots, k$, where $\tilde{I}(t)$ is defined in \eqref{Poisson}. 
Meanwhile, we also have a sequence of $k$ implemented control inputs $u_j \coloneqq u(j)$, $j = 0, 1, \dots, k-1$ generated by MPC, as shown in Fig.~\ref{fig:time_axis}. We employ the mean squared error (MSE) as the loss metric and enforce ${\hat{R}} = 1 - {\hat{S}} - {\hat{I}}$ during the training.

Now we define the data term $\text{Loss}_{\Data}$ as
\begin{subequations}
\begin{align}
\text{Loss}_{\Data}({\hat{I}},{ \hat{u}},\tilde{I}_i,u_j) 
&= \lambda_1 \text{MSE}^I_{\Data} + \lambda_2 \text{MSE}^u_{\Data}, \label{data1} \\
\intertext{\text{where}}  
\text{MSE}^I_{\Data} 
&= \frac{1}{k+1} \sum_{i=0}^{k} \left| \hat{I}(i) - \tilde{I}_i \right|^2, \label{data1b}\\
\text{MSE}^u_{\Data} 
&= \frac{1}{k} \sum_{j=0}^{k-1} \left| \hat{u}(j) - u_j \right|^2. \label{data1c}
\end{align}
\end{subequations}
We use $\lambda_\ast$ as tuning parameters throughout this work. 
Both $\text{MSE}^I_{\Data}$ and $\text{MSE}^u_{\Data}$ serve as training terms for the noisy infected states $\tilde{I}_i$ and the real-world control input $u_j$, respectively.

For the $SIR$ residual term $\text{Loss}_{\Res}$, we select $N_c$ collocation points $t_j$ sampled over the interval $[0, k]$, where $j = 1, \dots, N_c$, $N_c \in \mathbb{N}_{>0}$, and typically $N_c \gg k$. Based on~\eqref{1}, we have that 
\begin{subequations}
\begin{equation}
\begin{aligned}
\text{Loss}_{\Res}({\hat{S}},{\hat{I}}, {\hat{u}},{\hat{\beta}}) \!\!
=\!\! \lambda_3\, \text{MSE}^{S}_{\Res} \!\!
+\!\! \lambda_4\, \text{MSE}^{I}_{\Res} \!\!
+\!\! \lambda_5\, \text{MSE}^{R}_{\Res},
\label{Eq_Loss_Res}
\end{aligned}
\end{equation}
\vspace{0.05cm}where
\begin{equation}
\begin{aligned}
&\text{MSE}^S_{\Res} = \frac{1}{N_c} \sum_{j=1}^{N_c} 
\left| \frac{d\hat{S}(t_j)}{dt}  
+ \hat{\beta} \hat{I}(t_j) \hat{S}(t_j) \right|^2, \\
&\text{MSE}^I_{\Res}\! = \!\frac{1}{N_c}\!\! \sum_{j=1}^{N_c} \!
\left| \frac{d\hat{I}(t_j)\!}{dt}
\!-\! \hat{\beta} \hat{I}(t_j) \hat{S}(t_j) \!
+\! (\gamma\! +\! \hat{u}(t_j)) \hat{I}(t_j)\! \right|^2\!\!, \\
&\text{MSE}^R_{\Res} = \frac{1}{N_c} \sum_{j=1}^{N_c} 
\left| \frac{d\hat{R}(t_j)}{dt} 
- (\gamma + \hat{u}(t_j)) \hat{I}(t_j) \right|^2,
\end{aligned}
\label{RES_MSE}
\end{equation}
\end{subequations}
with the derivative $\frac{d\hat{v}(t_j)}{dt}$, for $v\in\{S, I, R\}$, obtained via automatic differentiation \cite{raissi2019physics}. These terms,  $\text{MSE}^S_{\Res}, \text{MSE}^I_{\Res}$ and $\text{MSE}^R_{\Res}$, embed the $SIR$ dynamics into training and ensure physical consistency, forming the basis of our PINNs approach.

The last term in \eqref{PINN_LOSS} incorporates the initial conditions in~\eqref{initial2} as
\begin{subequations}
\begin{equation}
\text{Loss}_{\IC}({\hat{S}},{\hat{I}}) = \lambda_6 \text{MSE}^S_{\IC} + \lambda_7 \text{MSE}^I_{\IC} + \lambda_8 \text{MSE}^R_{\IC}, \label{IC1}
\end{equation}
\text{where}
\begin{equation}
\begin{aligned}
&\text{MSE}^S_{\IC} = \left| \hat{S}(0) - (1 - I(0)) \right|^2, \\
&\text{MSE}^I_{\IC} = \left| \hat{I}(0) - I(0) \right|^2, 
\text{MSE}^R_{\IC} = \left| \hat{R}(0) \right|^2.
\end{aligned} \label{PINN_IC}
\end{equation}
\end{subequations}
The terms $\text{MSE}^S_{\IC}$, $\text{MSE}^I_{\IC}$, and $\text{MSE}^R_{\IC}$ incorporate the initial conditions into the training process.  Based on the Picard–Lindel\"{o}f Theorem~\cite{hartman2002ordinary}, these conditions ensure a unique solution to the $SIR$ dynamics. The uniqueness of the ODEs' solution is a critical factor in guaranteeing the training accuracy of PINNs~\cite{raissi2019physics}. 
By minimizing~\eqref{PINN_LOSS}, the MPC-PINNs simultaneously train all the SISO NNs, including $NN_{\hat{S}}$, $NN_{\hat{I}}$, $NN_{\hat{\beta}}$, and $NN_{\hat{u}}$.
With the inclusion of $\text{Loss}_{\Res}$, the MPC-PINNs evolve beyond the standard neural networks, enabling the state and parameter estimation while capturing their physical behavior.


After designing the MPC-PINNs with a set of single-input, single-output neural network architectures and the loss function in \eqref{PINN_LOSS}, we introduce the PINNs-based MPC framework in Fig.~\ref{fig:flowdiagram_main}. At the $k^{\text{th}}$ time step, MPC-PINNs train the neural networks $NN_{\hat{S}}$, $NN_{\hat{I}}$, $NN_{\hat{\beta}}$, and $NN_{\hat{u}}$ with the noisy infected states $\tilde{I}_i,i=0,\dots,k$ and the known control inputs $u_j,j=0,\dots,k-1$. These neural networks then provide the estimates $\hat{S}(k)$, $\hat{I}(k)$, $\hat{R}(k)$ (computed as $1-{\hat{S}(k)}-{\hat{I}(k)}$), and $\hat{\beta}(k)$ by setting $t=k$, which are fed into the MPC. Note that we do not use the output $\hat u$ of $NN_{\hat{u}}$ for the MPC, and $NN_{\hat{u}}$ is set only for the PINNs training.
MPC then solves the optimization problem in~\eqref{Eq_MPC_OP} to compute the optimal control input $u_k$, which is applied to $SIR$ models to generate new states for updating the MPC-PINNs. At the ${k + 1}^{\text{th}}$ time step, new state $\tilde{I}_{k+1}$ updates the MPC-PINNs with $u_k$ for the next iteration.  This real-time data updating mechanism supports iterative refinement of the prediction model, enabling continuous adaptation and improved accuracy of the estimation, which is further validated through experiments in Sec.~\ref{Sec7.1}.


\begin{figure}
    \centering
    \begin{minipage}{0.7\linewidth}
        \centering
        \includegraphics[width=0.96\linewidth]{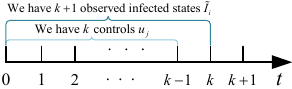}
        \subcaption{Time axis illustration.}
        \label{fig:time_axis}
    \end{minipage}
    \hfill
    \begin{minipage}{1.0\linewidth}
        \centering
        \includegraphics[width=0.96\linewidth]{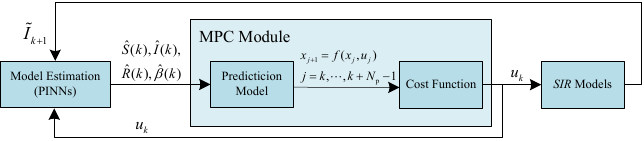}
        \subcaption{PINNs-based MPC closed-loop framework.}
        \label{fig:flowdiagram_main}
    \end{minipage}
    \caption{Schematic of the physics-informed neural networks-based MPC closed-loop framework: (a) illustrates the availability of observed infected states and controls over time and (b) shows the interaction between the PINNs, MPC, and $SIR$ models for real-time estimation and control.}
    \label{fig:flowdiagram}

\end{figure}

\subsection{LOG-SCALED PHYSICS-INFORMED NEURAL NETWORKS}
\label{Sec4}
Though $\text{Loss}_{\Res}$ in \eqref{PINN_LOSS} provides regularization~\cite{raissi2019physics} for the physics-informed training, our experiments show that the high observation noise used in $\text{Loss}_{\Data}$ deteriorates the training performance for all the SISO NNs, i.e., $NN_{\hat{S}}$, $NN_{\hat{I}}$, $NN_{\hat{\beta}}$, and $NN_{\hat{u}}$ (See Sec.~\ref{Sec7.1}).
This disruption occurs after we incorporate the control variable. 
Following the framework introduced in Sec.~\ref{Sec3.2}, and under Assumption~\ref{assumption1}, we introduce log-scaled PINNs for MPC (MPC-LS-PINNs) to replace the MPC-PINNs in the PINNs-based MPC framework. MPC-LS-PINNs modify $\text{Loss}_{\Data}({\hat{I}},{\hat{u}},\tilde{I}_i,u_j)$ in~\eqref{data1} by replacing $\text{MSE}^I_{\Data}$ with the log-relative error (LRE), while the other terms remain unchanged. We define LRE as
\begin{equation}
\text{LRE}^I_{\Data}\! =\! \frac{1}{k+1}\!\! \sum_{i=0}^{k}\! \left| \log\!{( \max[\hat{I}(i), \epsilon] )}\!\! -\! \log\!{( \max[\tilde{I}_i, \epsilon] )}\! \right|^2\!\!, \label{LRE}
\end{equation}
with $0 < \epsilon \ll 1$ ensuring the logarithmic terms remain well defined.
\begin{remark}
Without loss of generality, if we omit~$\epsilon$,~\eqref{LRE} becomes $\left| \log \frac{\hat{I}(i)}{\tilde{I}_i} \right|^2$. Unlike the traditional MSE used in~\eqref{data1}, which focuses on absolute errors, this logarithmic form emphasizes relative errors. Absolute errors can bias PINNs toward larger values and cause them to neglect the early and late epidemic stages when the true $I(t)$ is small. This issue is exacerbated especially when the noise is correlated with the actual data, impairing the learning of the physical system. Therefore, by focusing on the relative error in~\eqref{LRE}, the proposed loss function overcomes these limitations and enhances the PINNs’ ability to capture all the dynamics.
\end{remark}
After introducing the LRE term in \eqref{LRE}, we define the LRE-based data loss as
\begin{equation}
\text{Loss}^{\LS}_{\Data}({\hat I},{\hat u},\tilde{I}_i,u_j) = \lambda_1 \text{LRE}^I_{\Data} + \lambda_2 \text{MSE}^u_{\Data}, \label{dataLRE}
\end{equation}
where  $\text{MSE}^u_{\Data}$ is from \eqref{data1c}.
Then, we define the loss function in MPC-LS-PINNs as
\begin{equation}
\text{Loss}^{\LS}({\hat S}, {\hat I},{\hat u},{\hat \beta},\tilde{I}_i,u_j) = \text{Loss}^{\LS}_{\Data}+ \text{Loss}_{\Res} + \text{Loss}_{\IC} .
\label{LS_PINN_loss}
\end{equation}
At the $k^{\text{th}}$ time step of the PINNs-based MPC framework, the MPC-LS-PINNs module estimates the current states and parameters by minimizing the loss function in~\eqref{LS_PINN_loss}, as summarized in Algorithm~\ref{algorithmic1}.

\begin{algorithm}
\caption{MPC-LS-PINNs for the $k^{\text{th}}$ time step in the PINNs-based MPC framework}
\label{algorithmic1}
\begin{algorithmic}
\State \textbf{Input:} Noisy infected states $\tilde{I}_i$ for $i=0,1,\dots,k$, implemented control $u_j$ for $j=0,1,\dots,k-1$, and the known recovery rate $\gamma$.
\State \textbf{Output:} Estimated current states $\hat{S}(k)$, $\hat{I}(k)$, $\hat{R}(k)$, transmission rate $\hat{\beta}(k)$, and the optimal control $u(k)$.
\State {Initialization:} Load the saved weights of $NN_{\hat{S}}, NN_{\hat{I}},$ and $NN_{\hat{\beta}}$ from the $k-1^{\text{th}}$ time step.
\State \textbf{Loss function construction:} To jointly train $NN_{\hat{S}}, NN_{\hat{I}}, NN_{\hat{\beta}},$ and $NN_{\hat{u}}$, compute the total loss $\text{Loss}^{\mathrm{LS}}(\hat{S}, \hat{I}, \hat{u}, \hat{\beta}, \tilde{I}_i, u_j)$ in~\eqref{LS_PINN_loss} as the weighted sum of the data-fitting, physics-residual, and initial-condition terms. Save the trained weights of all NNs.

\State \textbf{MPC solver:}
\Statex \algindent $NN_{\hat{S}}$ and $NN_{\hat{I}}$ generate $\hat{S}(k)$ and $\hat{I}(k)$.
\Statex \algindent Compute $\hat{R}(k) = 1 - \hat{S}(k) - \hat{I}(k)$.
\Statex \algindent $NN_{\hat{\beta}}$ produces $\hat{\beta}(k)$.
\Statex \algindent MPC uses:
\Statex \algindenta $\hat{S}(k)$, $\hat{I}(k)$, and $\hat{R}(k)$ as current states.
\Statex \algindenta $\hat{\beta}(k)$ and the known $\gamma$ as current parameters.
\Statex \algindent Solve \eqref{Eq_MPC_OP} to obtain:
\Statex \algindenta Optimal control input $u(k)$.
\Statex \algindenta Predicted states from $k$ to $k + N_{\mathrm{p}} - 1$.

\State This process is repeated at each time step $k$ within the MPC framework.
\end{algorithmic}
\end{algorithm}

Algorithm~\ref{algorithmic1} explains how MPC-LS-PINNs estimate the states and parameters with only the observed infected states at the $k^{\text{th}}$ time step. 
These SISO NNs, which take $t$ as input and output the states ($\hat{S}$ and $\hat I$) and parameter ($\hat{\beta}$), are used to construct an $SIR$ model. This model is then incorporated into the MPC framework to determine $u(k)$. By initializing the NNs with the weights obtained from the previous training, we provide a well-informed starting point for the optimization, which reduces the training cost and enables more efficient real-time deployment.

\subsection{SPLIT-INTEGRAL PHYSICS-INFORMED NEURAL NETWORKS}
\label{Sec5}
As presented in Sec.~\ref{Sec3.2} and~\ref{Sec4}, MPC-PINNs and MPC-LS-PINNs calibrate the states and transmission rate jointly by minimizing the data loss function and $SIR$ residuals. However, this inverse problem requires multiple iterations to converge~\cite{millevoi2024physics}. 
Building on the direct link between noisy infected states and model states~\cite{millevoi2024physics}, we propose a split-integral PINNs (\textit{MPC-SI-PINNs}) algorithm to replace MPC-PINNs within the same PINNs-based MPC framework. At the $k^{\text{th}}$ time step, with known $\tilde{I}_i$, for $i=0,\dots,k$ and $u_j$, for $j=0,\dots,k-1$, MPC-SI-PINNs decouple the data regression from the physics-informed training.
This method first fits the noisy infected states and then enhances the physics-informed training with the integral-derived state estimates.

We introduce MPC-SI-PINNs in three steps. 
First, to train the SISO neural networks $NN_{\hat{I}}$ and $NN_{\hat{u}}$, the data regression step uses the original data loss term as
\begin{equation}
\text{Loss}^{\SI}_{\Data1}({\hat I},{\hat u},\tilde{I}_i,u_j) = \alpha_1 \text{MSE}^I_{\Data} + \alpha_2 \text{MSE}^u_{\Data},
\label{Eq_Loss_SI_Data}
\end{equation}
where $\text{MSE}^I_{\Data}$ and $\text{MSE}^u_{\Data}$ are defined in~\eqref{data1c}; and $\alpha_\ast$ tunes the loss function terms. This independent training, which uses $\tilde{I}_i$ and $u_j$, is based on the separate SISO NN architecture described in Sec. \ref{Sec3.1}.
Without the control term, this step is equivalent to the split method from~\cite{millevoi2024physics}, which separates the data regression from the physics-informed training.  
Unlike~\cite{millevoi2024physics}, where the neural network for the infected states is trained independently to reduce the computational load of physics-informed training, our approach aims to enrich the information available for physics-informed training. We achieve this goal by first using the observed infected states $\tilde{I}$ and control inputs u, which facilitates the estimation of additional states through their corresponding SISO NNs. Building on this property, we introduce the next step: the integral operation.

Based on the output $\hat{I}$ of the neural network $NN_{\hat{I}}$, the integral operation discretizes~\eqref{1} and~\eqref{initial2}, with $T_s = 1$, as introduced in Sec.~\ref{subsection C}. This discretization is then used to compute  
\begin{subequations}
\begin{equation}
R^{\text{d}}(i+1) =  (\gamma + u(i)) \hat{I}(i) + R^{\text{d}}(i), i=0, 1, \cdots, k-1,
\label{integral1}
\end{equation}
\begin{equation}
S^{\text{d}}(j) = 1 - R^{\text{d}}(j) - \hat{I}(j), j=0, 1, \cdots, k,
\label{integral2}
\end{equation}\label{integral}\end{subequations}
where $R^{\text{d}}(0) = R(0)$, and the superscript $\text{d}$ denotes the values derived from the integral step.
In the third step, the physics-informed training, we form the following new loss function 
\begin{equation}
\begin{aligned}
    &\text{Loss}^{\SI}({\hat S}, {\hat I}, {\hat \beta},S^{\text{d}},\tilde{I}_i,R^{\text{d}}) =\\ 
    &\text{Loss}^{\SI}_{\Data2}({\hat S}, {\hat I},S^{\text{d}},\tilde{I}_i,R^{\text{d}})+ \text{Loss}^{\SI}_{\Res}(\hat{S}, {\hat I},\hat{\beta}),
\end{aligned}\label{SI-LOSS}
\end{equation}
where derived data term $\text{Loss}^{\SI}_{\Data2}$, which incorporates the derived data $S^{\text{d}}$ and $R^{\text{d}}$, is given by
\begin{equation}
\begin{aligned}
    \text{Loss}^{\SI}_{\Data2}({\hat S},S^{\text{d}},R^{\text{d}}) = \alpha_3 \text{MSE}^S_{\Data}  +\alpha_4 \text{MSE}^R_{\Data}, \label{SIdata2}
\end{aligned}
\end{equation}
with
\begin{equation*}
\begin{aligned}
\text{MSE}^S_{\Data} &= \frac{1}{k+1} \sum_{i=0}^{k} \left| \hat{S}(i) - S^{\text{d}}(i) \right|^2 \text{ and}\\ 
\text{MSE}^R_{\Data} &= \frac{1}{k+1} \sum_{i=0}^{k} \left| \hat{R}(i) - R^{\text{d}}(i) \right|^2.
\end{aligned}
\end{equation*}
The residual term $\text{Loss}^{\SI}_{\Res}$, with $NN_{\hat I}$ and $NN_{\hat u}$ fixed in the data regression step, is defined as
\begin{equation}
\text{Loss}^{\SI}_{\Res}({\hat S}, {\hat \beta}) = \alpha_5 \text{MSE}^S_{\Res} + \alpha_6 \text{MSE}^I_{\Res} + \alpha_7 \text{MSE}^R_{\Res}, \label{SPRes1}
\end{equation}
where $\text{MSE}^S_{\Res}$, $\text{MSE}^I_{\Res}$,  and $\text{MSE}^R_{\Res}$ are defined in \eqref{RES_MSE}. Now we introduce Algorithm \ref{algorithmic2}. 
By minimizing \eqref{SI-LOSS}, the MPC-SI-PINNs optimize $NN_{\hat S}$ and $NN_{\hat \beta}$. Note that the initial condition terms, which are included in the $SIR$ data term, are omitted in~\eqref{SI-LOSS}. The details of MPC-SI-PINNs are in Algorithm \ref{algorithmic2}. 
\begin{remark}
Compared to the MPC-PINNs described in Sec.~\ref{Sec3.2}, MPC-SI-PINNs reconstruct the system states $S^{\text{d}}$ and $ R^{\text{d}}$ for physics-informed training, leading to improved performance as demonstrated in Sec.~\ref{results}. Initializing the neural networks with weights from the previous training provides a well-informed starting point, which reduces the training cost and enhances the computational efficiency in real-time operation. Figure~\ref{fig:PINNs_flowcharts} summarizes the workflows of MPC-PINNs, MPC-LS-PINNs and MPC-SI-PINNs.     
\end{remark}
\begin{algorithm}
\caption{MPC-SI-PINNs for the $k^{\text{th}}$ time step in the PINNs-based MPC framework} 
\label{algorithmic2}
\begin{algorithmic}[1]
\State \textbf{Input:} Noisy infected states $\tilde{I}_i$ for $i=0,1,\dots,k$, implemented control inputs $u_j$ for $j=0,1,\dots,k-1$, and the recovery rate $\gamma$.
\State \textbf{Output:} Estimated current states $\hat{S}(k)$, $\hat{I}(k)$, $\hat{R}(k)$, transmission rate $\hat{\beta}(k)$, and the optimal control input $u(k)$.
\State \textbf{Initialization:} Load the previously saved weights of $NN_{\hat{S}}, NN_{\hat{I}},$ and $NN_{\hat{\beta}}$ from the $k-1^{\text{th}}$ time step.
\State \textbf{Data regression:} Train $NN_{\hat I}$ and $NN_{\hat u}$ by minimizing $\text{Loss}^{\SI}_{\Data1}$ in~\eqref{Eq_Loss_SI_Data}. 
\State \textbf{Integral operation:} Compute $S^{\text{d}}(i)$ and $R^{\text{d}}(i)$ from the output $\hat I(i)$ of $NN_{\hat I}$, $i=0,\dots,k$, using~\eqref{integral1}--\eqref{integral2}.
\State \textbf{Physics-informed training:} With fixed $NN_{\hat I}$ and $NN_{\hat u}$, optimize $NN_{\hat S}$ and $NN_{\hat \beta}$ by minimizing $\text{Loss}^{\SI}$ in~\eqref{SI-LOSS}. Save the training results of all NN weights.

\State \textbf{MPC solver:}
\Statex \algindent $NN_{\hat{S}}$ and $NN_{\hat{I}}$ generate $\hat{S}(k)$ and $\hat{I}(k)$.
\Statex \algindent Compute $\hat{R}(k) = 1 - \hat{S}(k) - \hat{I}(k)$.
\Statex \algindent $NN_{\hat{\beta}}$ produces $\hat{\beta}(k)$.
\Statex \algindent MPC uses:
\Statex \algindenta $\hat{S}(k)$, $\hat{I}(k)$, and $\hat{R}(k)$ as current states.
\Statex \algindenta $\hat{\beta}(k)$ and the known $\gamma$ as current parameters.
\Statex \algindent Solve \eqref{Eq_MPC_OP} to obtain:
\Statex \algindenta Optimal control input $u(k)$.
\Statex \algindenta Predicted states for $k$ to $k + N_{\mathrm{p}} - 1$.

\State This process is repeated at each time step $k$ within the MPC framework.
\end{algorithmic}
\end{algorithm}

\begin{figure}
    \centering
    \subfloat[]{\includegraphics[width=0.99\linewidth]{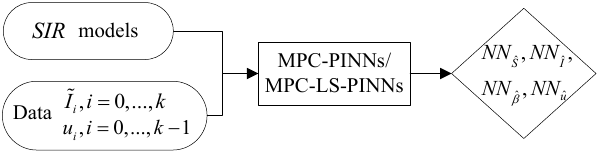}}\\
    \subfloat[]{\includegraphics[width=0.99\linewidth]{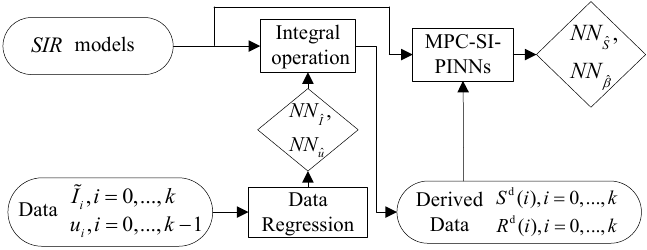}}
    \caption{Implementation flowcharts of different PINNs: 
    (a) MPC-PINNs/MPC-LS-PINNs implementation process; 
    (b) MPC-SI-PINNs implementation process.}
    \label{fig:PINNs_flowcharts}
\end{figure}

Compared to our MPC-LS-PINNs algorithm, the proposed MPC-SI-PINNs algorithm in Algorithm~\ref{algorithmic2} does not include the loss term $\text{LRE}^I_{\Data}$ in~\eqref{LRE} for noisy infected states. Instead, it employs a conventional SISO neural network in the data regression stage to process the observed data directly. Careful tuning of key hyperparameters (e.g., number of epochs, batch size, and regularization) helps reduce the effect of noise and prevents both data distortion and overfitting. However, the data regression performed by $NN_{\hat I}$ may still introduce errors that depend on the chosen hyperparameters. These errors may propagate through the subsequent integral operation, leading to accumulated error over time. This effect becomes particularly noticeable when observation noise is high (see Sec.~\ref{Sec7.1}).

\subsection{GENERALIZED PHYSICS-INFORMED NEURAL NETWORKS}
\label{Sec6}
The previous sections focus on the scenario when noisy infected states are available and the recovery rate $\gamma$ is known (Assumption~\ref{assumption1}). In this section, we investigate the estimation of all states and parameters when both the transmission rate $\beta$ and the recovery rate $\gamma$ are unknown. Following Assumption~\ref{assumption2}, we assume that the basic reproduction number $\mathcal{R}_0 = \frac{\beta}{\gamma}$ is known. Under this setting, we adapt the proposed MPC-PINNs, MPC-LS-PINNs, and MPC-SI-PINNs algorithms to the new conditions. We begin by examining the conditions for existence and uniqueness of solutions, where a solution is defined as a parameter–state pair $(\beta, \gamma, S(t), R(t))$ that satisfies the system in~\eqref{1} and is consistent with the observed trajectories of $I(t)$ and $u(t)$.
\begin{theorem} \label{Theorem}
Consider the system in~\eqref{1} with initial conditions given in~\eqref{initial2}. Suppose the trajectories of the infected proportion $I(t)$ and control input $u(t)$ are known for $t \in [0, k]$. Then, there exists a family of  solutions $\{(\beta^\dagger, \gamma^\dagger, S^\dagger(t), R^\dagger(t))\}$ that satisfy~\eqref{1} over $[0, k]$.
\end{theorem}
\begin{proof}
The proof is provided in the appendix.
\end{proof}

\begin{corollary} \label{corollary}
Consider the system in~\eqref{1} with initial conditions given in~\eqref{initial2}. Suppose the trajectories of the infected proportion $I(t)$ and control input $u(t)$ are known for $t \in [0, k]$, and the basic reproduction number $\mathcal{R}_0$ is known. Then, there exists a unique solution $(\beta, \gamma, S(t), R(t))$ that satisfies~\eqref{1} over $[0, k]$.
\end{corollary}
\begin{proof}
The proof is provided in the appendix.
\end{proof}

Theorem~\ref{Theorem} and Corollary~\ref{corollary} justify the need for an additional assumption on the basic reproduction number when both $\beta$ and $\gamma$ are unknown. With the guarantee provided by Corollary~\ref{corollary}, under Assumption~\ref{assumption2}, we adapt the MPC-PINNs and MPC-LS-PINNs algorithms to handle this scenario. In particular, we replace the SISO neural network $NN_{\hat{\beta}}$ (originally used to estimate the transmission rate) with a SISO neural network $NN_{\hat{\gamma}}$ to estimate the recovery rate. The transmission rate is then computed from the known $\mathcal{R}_0$ as ${\hat{\beta}} = \mathcal{R}_0 , {\hat{\gamma}}$. 
Based on Assumption \ref{assumption2}, we change the $SIR$ residual loss function to
\begin{subequations}
\begin{equation}
\begin{aligned}
\text{Loss}^{\mathcal{R}_0}_{\Res} =
\lambda_3\, \text{MSE}^{S(\mathcal{R}_0)}_{\Res}
+ \lambda_4\, \text{MSE}^{I(\mathcal{R}_0)}_{\Res}
+ \lambda_5\, \text{MSE}^{R(\mathcal{R}_0)}_{\Res},
\end{aligned} \label{LOSS_R0_RES}
\end{equation}
\text{where}
\begin{equation}
\begin{aligned}
\text{MSE}^{S(\mathcal{R}_0)}_{\Res}
= \frac{1}{N_c} \sum_{j=1}^{N_c}
\left| \frac{d\hat{S}(t_j)}{dt}
+ \mathcal{R}_0 \hat{\gamma} \hat{I}(t_j) \hat{S}(t_j) \right|^2
\end{aligned}
\end{equation}
\vspace{-2mm}
\begin{equation}
\begin{aligned}
\text{MSE}^{I(\mathcal{R}_0)}_{\Res}
= \frac{1}{N_c} \sum_{j=1}^{N_c}
\biggl|
\frac{d\hat{I}(t_j)}{dt}
- \Bigl(
\mathcal{R}_0 \hat{\gamma} \hat{I}(t_j) \hat{S}(t_j) \\[-2.5ex]
- (\hat{\gamma} + \hat{u}(t_j)) \hat{I}(t_j)
\Bigr)
\biggr|^2
\end{aligned}
\end{equation}
\vspace{-2mm}
\begin{equation}
\begin{aligned}
\text{MSE}^{R(\mathcal{R}_0)}_{\Res}
= \frac{1}{N_c} \sum_{j=1}^{N_c}
\left| \frac{d\hat{R}(t_j)}{dt}
- (\hat{\gamma} + \hat{u}(t_j)) \hat{I}(t_j) \right|^2\!\!.
\end{aligned}
\end{equation} \label{Res_R0_SIR}
\end{subequations}\!\!\!By replacing $\text{Loss}_{\Res}$ in the total loss function of the MPC-PINNs (see~\eqref{PINN_LOSS}) with $\text{Loss}^{\mathcal{R}0}_{\Res}$ from~\eqref{LOSS_R0_RES}, we obtain the generalized MPC-PINNs, which employ $\text{MSE}^I_{\Data}$ in~\eqref{data1c}. Likewise, replacing $\text{Loss}_{\Res}$ in~\eqref{LS_PINN_loss} with $\text{Loss}^{\mathcal{R}0}_{\Res}$ yields the generalized MPC-LS-PINNs, which employ $\text{LRE}^I_{\Data}$ in~\eqref{LRE}.

\begin{algorithm}[!t]
\caption{Generalized MPC-S-PINNs for the $k^{\text{th}}$ time step in the PINNs-based MPC framework}
\label{algorithm3}
\begin{algorithmic}[1]
\State \textbf{Input:} Noisy infected states $\tilde{I}_i$ for $i = 0,1,\dots,k$, implemented control inputs $u_j$ for $j = 0,1,\dots,k-1$, and the known basic reproduction number $\mathcal{R}_0$.
\State \textbf{Output:} Estimated current states $\hat{S}(k)$, $\hat{I}(k)$, $\hat{R}(k)$, parameters $\hat{\beta}(k)$ and $\hat{\gamma}(k)$, and the optimal control input $u(k)$.
\State \textbf{Initialization:} Load the previously saved weights of $NN_{\hat{S}}$, $NN_{\hat{I}}$, and $NN_{\hat{\beta}}$ from the $k{-}1^\text{th}$ time step. 
\State \textbf{Data regression:} Train $NN_{\hat{I}}$ and $NN_{\hat{u}}$ by minimizing $\text{Loss}^{\Split}_{\Data}$ in~\eqref{Split_data}.
\State \textbf{Physics-informed training:} With $NN_{\hat{I}}$ and $NN_{\hat{u}}$ fixed, optimize $NN_{\hat{S}}$ and $NN_{\hat{\gamma}}$ by minimizing $\text{Loss}^{\Split}$ in~\eqref{spilt_loss}. Save the trained weights of all NNs.
\State \textbf{MPC solver:}
\Statex \algindent $NN_{\hat{S}}$ and $NN_{\hat{I}}$ generate $\hat{S}(k)$ and $\hat{I}(k)$.
\Statex \algindent Compute $\hat{R}(k) = 1 - \hat{S}(k) - \hat{I}(k)$.
\Statex \algindent $NN_{\hat{\gamma}}$ produces $\hat{\gamma}(k)$.
\Statex \algindent Compute $\hat{\beta}(k) = \mathcal{R}_0 \, \hat{\gamma}(k)$.
\Statex \algindent MPC uses:
\Statex \algindenta $\hat{S}(k)$, $\hat{I}(k)$, and $\hat{R}(k)$ as current states.
\Statex \algindenta $\hat{\beta}(k)$ and $\hat{\gamma}(k)$ as current parameters.
\Statex \algindent Solve~\eqref{Eq_MPC_OP} to obtain:
\Statex \algindenta Optimal control input $u(k)$.
\Statex \algindenta Predicted states for $k$ to $k + N_{\mathrm{p}} - 1$.
\State This process is repeated at each time step $k$ within the MPC framework.
\end{algorithmic}
\end{algorithm}
\begin{table*}[!t]
\centering
\caption{Summary of the proposed PINNs-based MPC frameworks, main features, and conditions.}
\label{tab:proposed_PINNs_summary}

\begin{tabularx}{\textwidth}{@{} l Y p{2.4cm} @{}} 
\toprule
\textbf{Method} & \textbf{Main Features} & \textbf{Condition} \\
\midrule

MPC-PINNs
& Baseline PINNs-based MPC under Assumption~\ref{assumption1}; jointly estimates $S$, $I$, $R$, and $\beta$ from noisy data $\tilde{I}_i$ and known $\gamma$; integrates estimation into the MPC loop.
& \multirow{3}{2.4cm}{\raggedright Assumption~\ref{assumption1}: the initial conditions, observed infected states $\tilde{I}(t)$, and the recovery rate $\gamma$ are known.} \\
\cmidrule(lr){1-2}
MPC-LS-PINNs
& Replaces infection data MSE with a LRE term to emphasize relative errors, avoiding large-value bias and preserving learning when $I$ is small; robust to scale‑correlated noise (e.g., Poisson).
& \\
\cmidrule(lr){1-2}
MPC-SI-PINNs
& Splits data regression and physics‑informed training; derives $S^{\text{d}}$ and $R^{\text{d}}$ via integrals from $\hat I$ and $u$ to enrich training information. 
& \\
\midrule

Generalized MPC-PINNs
& Baseline PINNs-based MPC under Assumption~\ref{assumption2}; Under known $R_0$, replace $NN_{\hat \beta}$ with $NN_{\hat \gamma}$; compute $\beta$ via $\beta=R_0\hat \gamma$; adapt $SIR$ residual loss to $R_0$‑based form.
& \multirow{3}{2.4cm}{\raggedright Assumption~\ref{assumption2}: the initial conditions, observed infected states $\tilde{I}(t)$, and the basic reproduction number $\mathcal{R}_0$ are known.} \\
\cmidrule(lr){1-2}
Generalized MPC-LS-PINNs
& Extension of MPC-LS-PINNs under Assumption~\ref{assumption2}; retains LRE for noise robustness and uses $R_0$‑based $SIR$ residual loss.
& \\
\cmidrule(lr){1-2}
Generalized MPC-S-PINNs
& Splits data regression and physics‑informed training without the integral step; fix $\hat I$ and $\hat u$ from data regression, then train $\hat S$ and $\hat\gamma$ with $R_0$‑based $SIR$ residual loss and initial condition loss. 
& \\
\bottomrule
\end{tabularx}
\end{table*}

Furthermore, when both $\beta$ and $\gamma$ are unknown, the integral operation in the MPC-SI-PINNs algorithm (Algorithm~\ref{algorithmic2}) cannot be applied. By removing this step and adjusting the remaining components, we obtain the generalized split-PINNs, referred to as \textit{generalized MPC-S-PINNs}, which consist of two steps. In the first step, with $\omega_\ast$ denoting the weights in the generalized split-PINNs and using~\eqref{data1c}, data regression is performed with the loss function
\begin{equation}
    \text{Loss}^{\Split}_{\Data}({\hat I}, {\hat u},\tilde{I}_i,u_j) = \omega_1\text{MSE}^I_{\Data} + \omega_2\text{MSE}^u_{\Data}, \label{Split_data}
\end{equation}
which is used to train $NN_{\hat I}$ and $ NN_{\hat u}$. After fixing  $NN_{\hat I}$ and $ NN_{\hat u}$, the second step, called physics-informed training, is different from the MPC-SI-PINNs. Based on the definitions in \eqref{PINN_IC} and \eqref{Res_R0_SIR}, with fixed $NN_{\hat I}$ and $ NN_{\hat u}$, this step directly trains the loss 
\begin{equation}
\begin{aligned}
&\text{Loss}^{\Split} ({\hat S} ,{\hat \gamma})= \omega_3\text{MSE}^{S(\mathcal{R}_0)}_{\Res} + 
\omega_4\text{MSE}^{I(\mathcal{R}_0)}_{\Res}+ \\
&\omega_5\text{MSE}^{R(\mathcal{R}_0)}_{\Res}+
\omega_6\text{MSE}^S_{\IC} + \omega_7\text{MSE}^R_{\IC},
\end{aligned} \label{spilt_loss}
\end{equation}
where the initial condition terms are added again after removing the integral operation step. The generalized MPC-S-PINNs algorithm is summarized in Algorithm~\ref{algorithm3}. By separating the data regression from the physics-informed constraints, the generalized MPC-S-PINNs in Algorithm~\ref{algorithm3} improve training efficiency compared to the MPC-PINNs (see results in Sec.~\ref{Sec7.3}). Note that if the control input is removed and the recovery rate $\gamma$ is known instead of the basic reproduction number $\mathcal{R}_0$, the generalized MPC-S-PINNs reduce to the split-PINNs framework in\cite{millevoi2024physics}.

\vspace{-0.1cm}
\section{EXPERIMENTS AND COMPARISONS}
\label{results}

\begin{figure*}[!t]
    \centering
    \begin{subfigure}[b]{0.24\linewidth}
        \includegraphics[width=\linewidth]{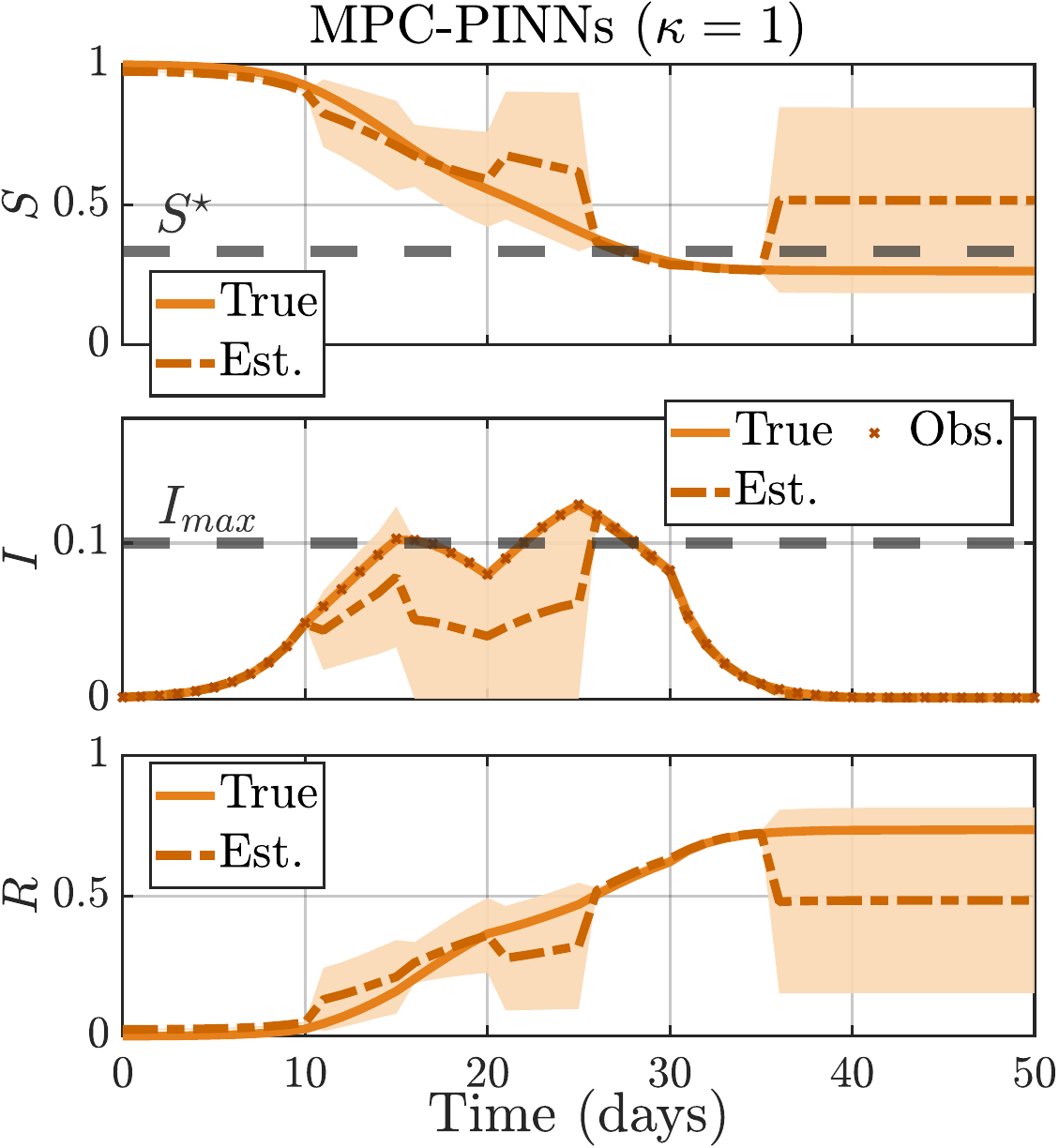}
        \caption{}
        \label{fig:PINNs_LS_SI_a}
    \end{subfigure}
    \hfill
    \begin{subfigure}[b]{0.24\linewidth}
        \includegraphics[width=\linewidth]{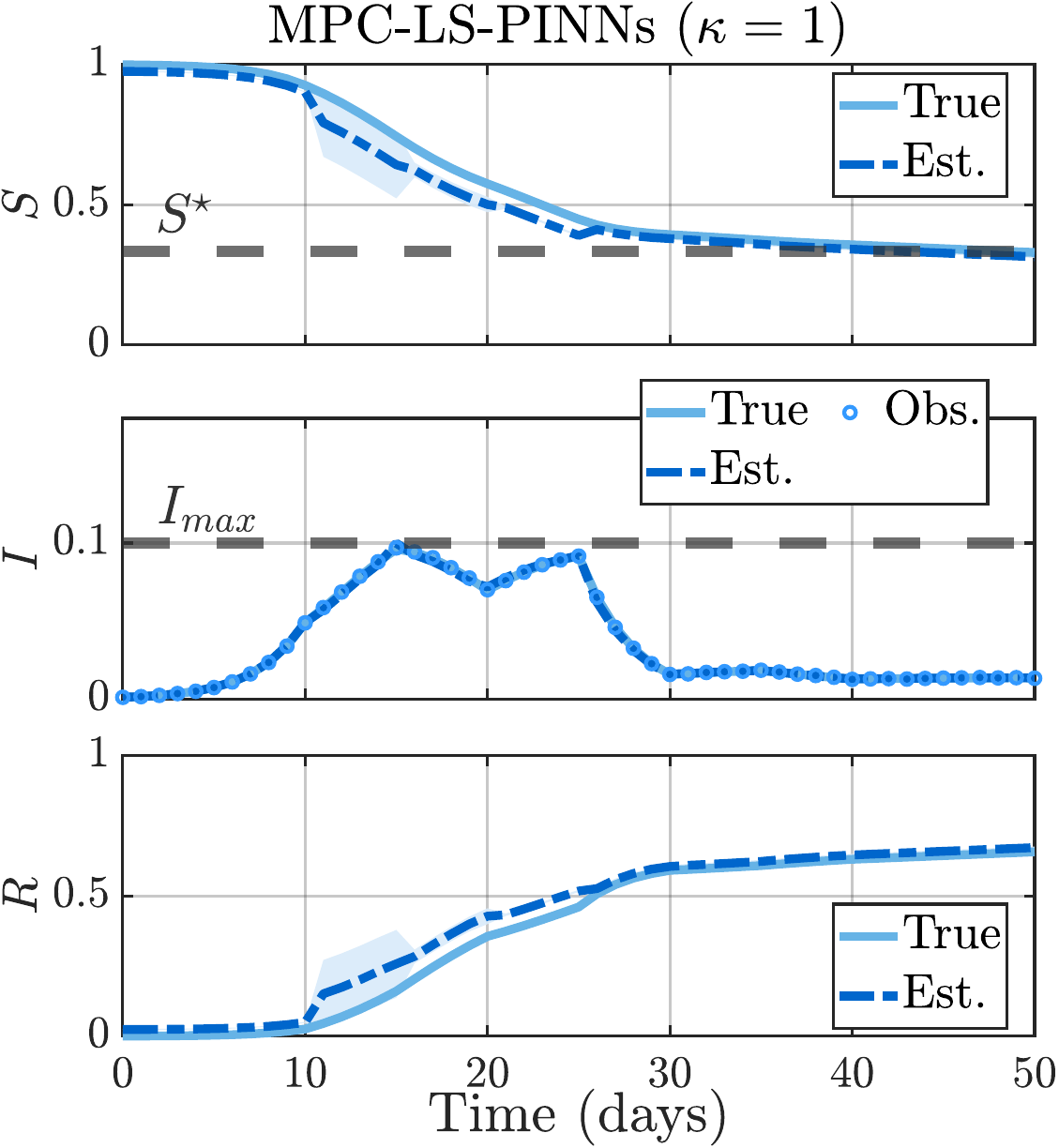}
        \caption{}
        \label{fig:PINNs_LS_SI_b}
    \end{subfigure}
    \hfill
    \begin{subfigure}[b]{0.24\linewidth}
        \includegraphics[width=\linewidth]{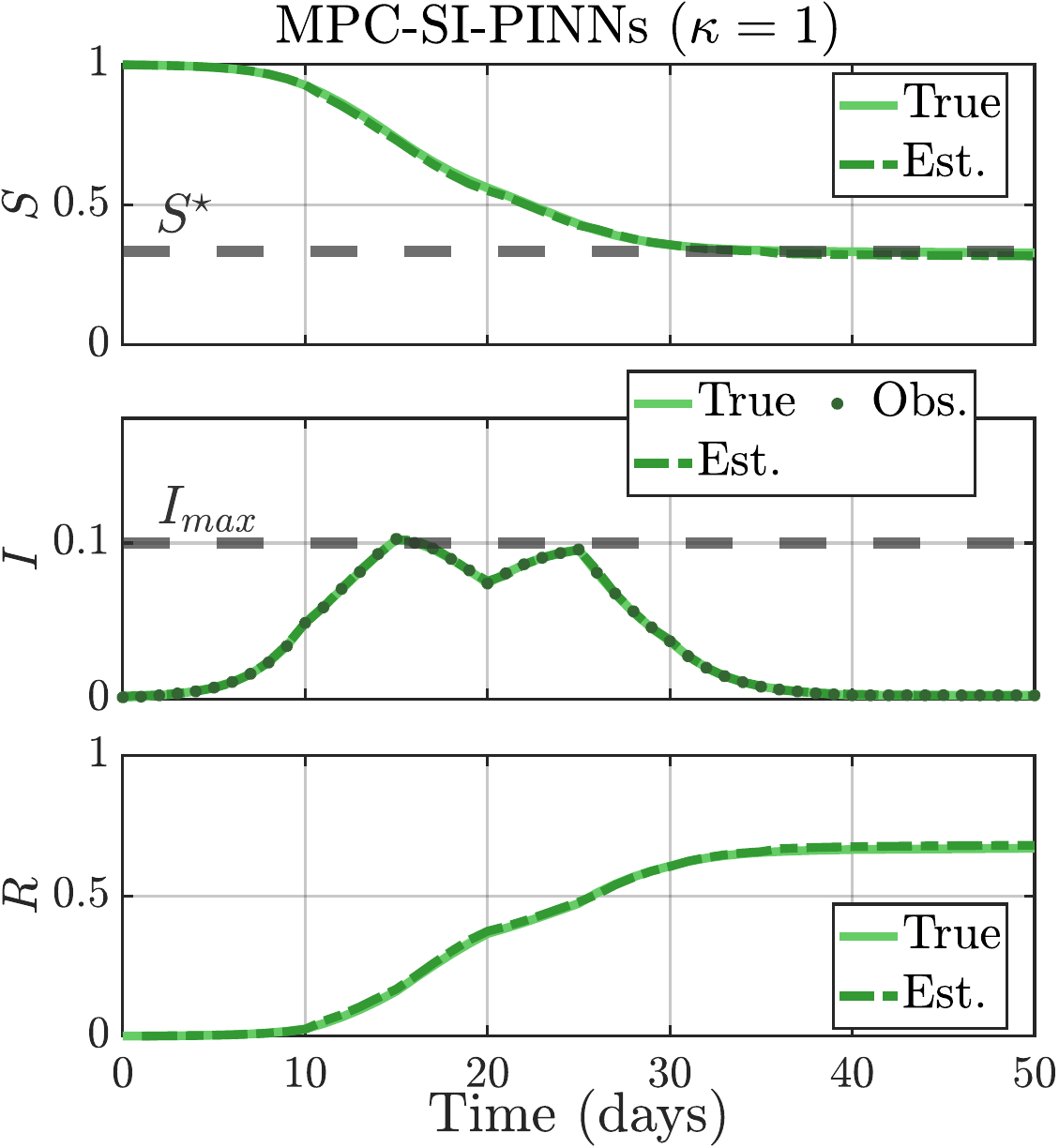}
        \caption{}
        \label{fig:PINNs_LS_SI_c}
    \end{subfigure}
    \hfill
    \begin{subfigure}[b]{0.24\linewidth}
        \includegraphics[width=\linewidth]{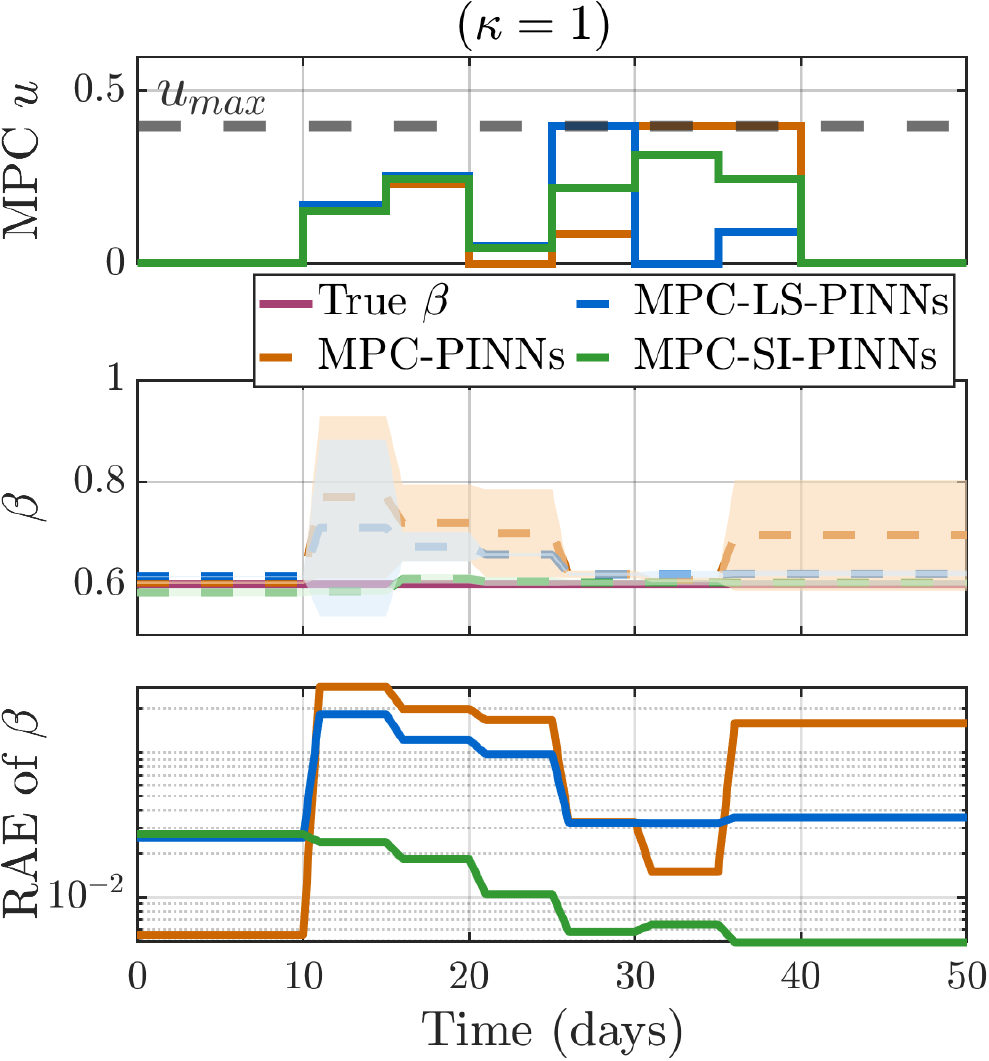}
        \caption{}
        \label{fig:PINNs_LS_SI_d}
    \end{subfigure}

    \vspace{0.3cm}

    \begin{subfigure}[b]{0.24\linewidth}
        \includegraphics[width=\linewidth]{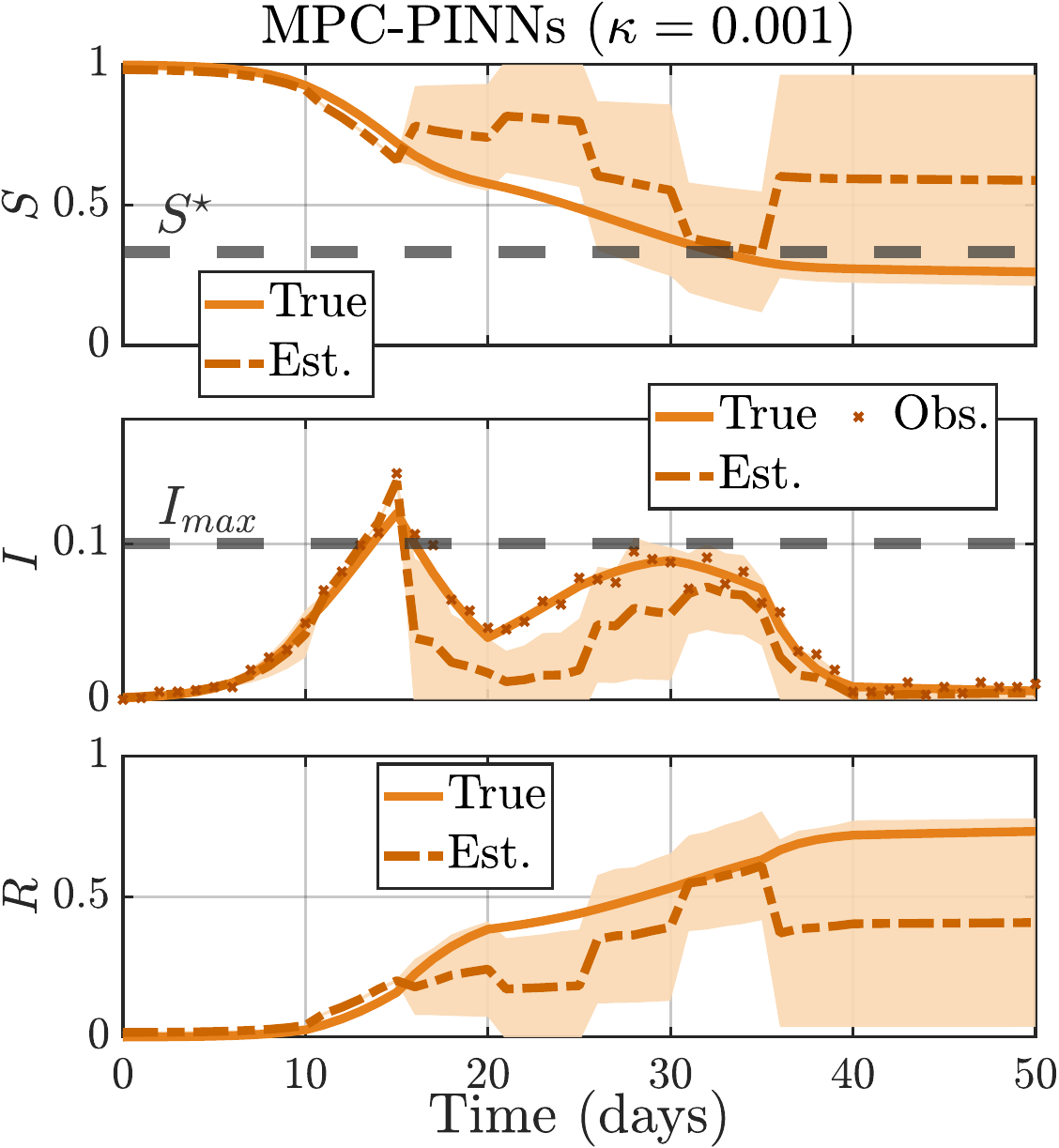}
        \caption{}
        \label{fig:PINNs_LS_SI_e}
    \end{subfigure}
    \hfill
    \begin{subfigure}[b]{0.24\linewidth}
        \includegraphics[width=\linewidth]{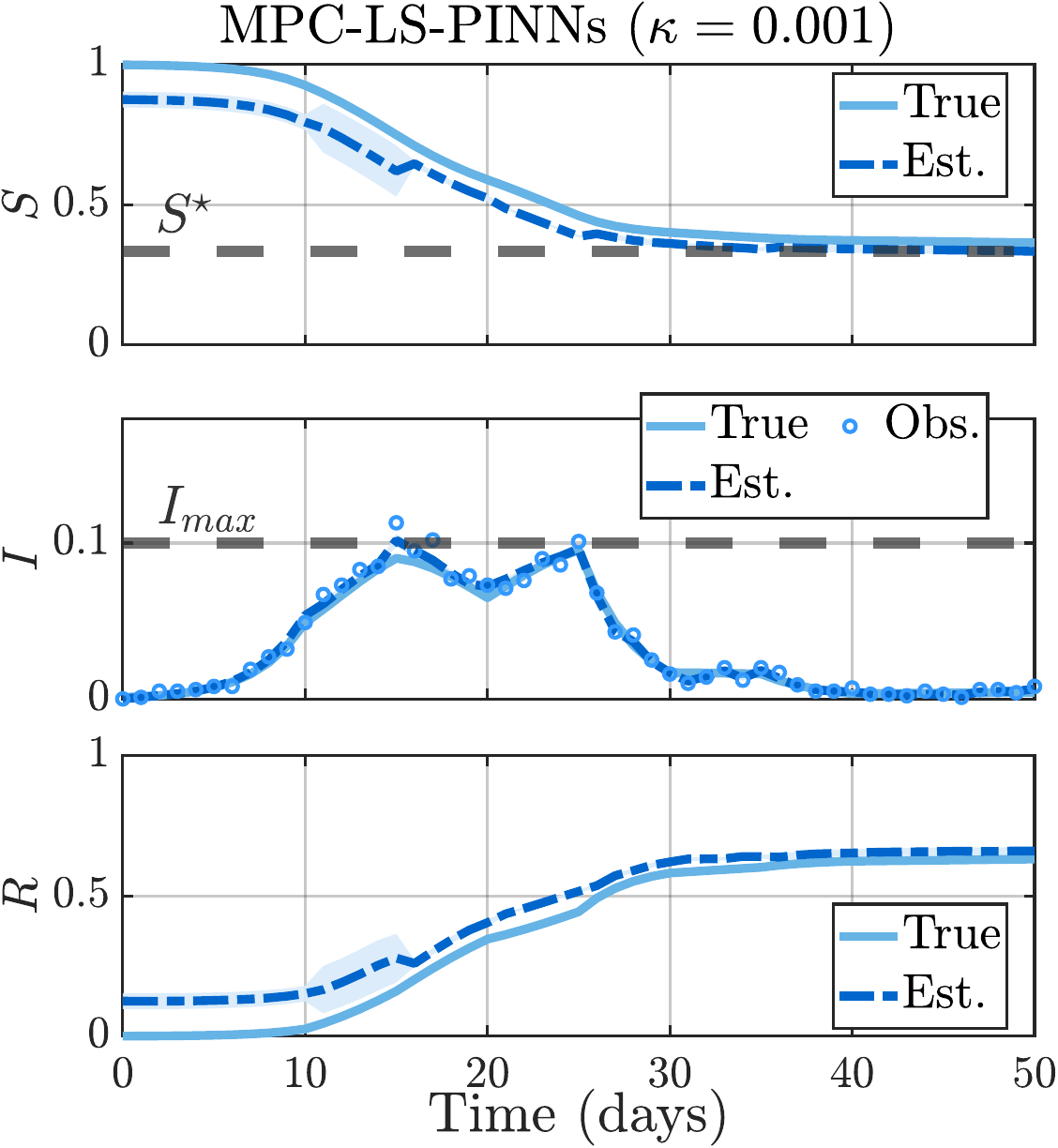}
        \caption{}
        \label{fig:PINNs_LS_SI_f}
    \end{subfigure}
    \hfill
    \begin{subfigure}[b]{0.24\linewidth}
        \includegraphics[width=\linewidth]{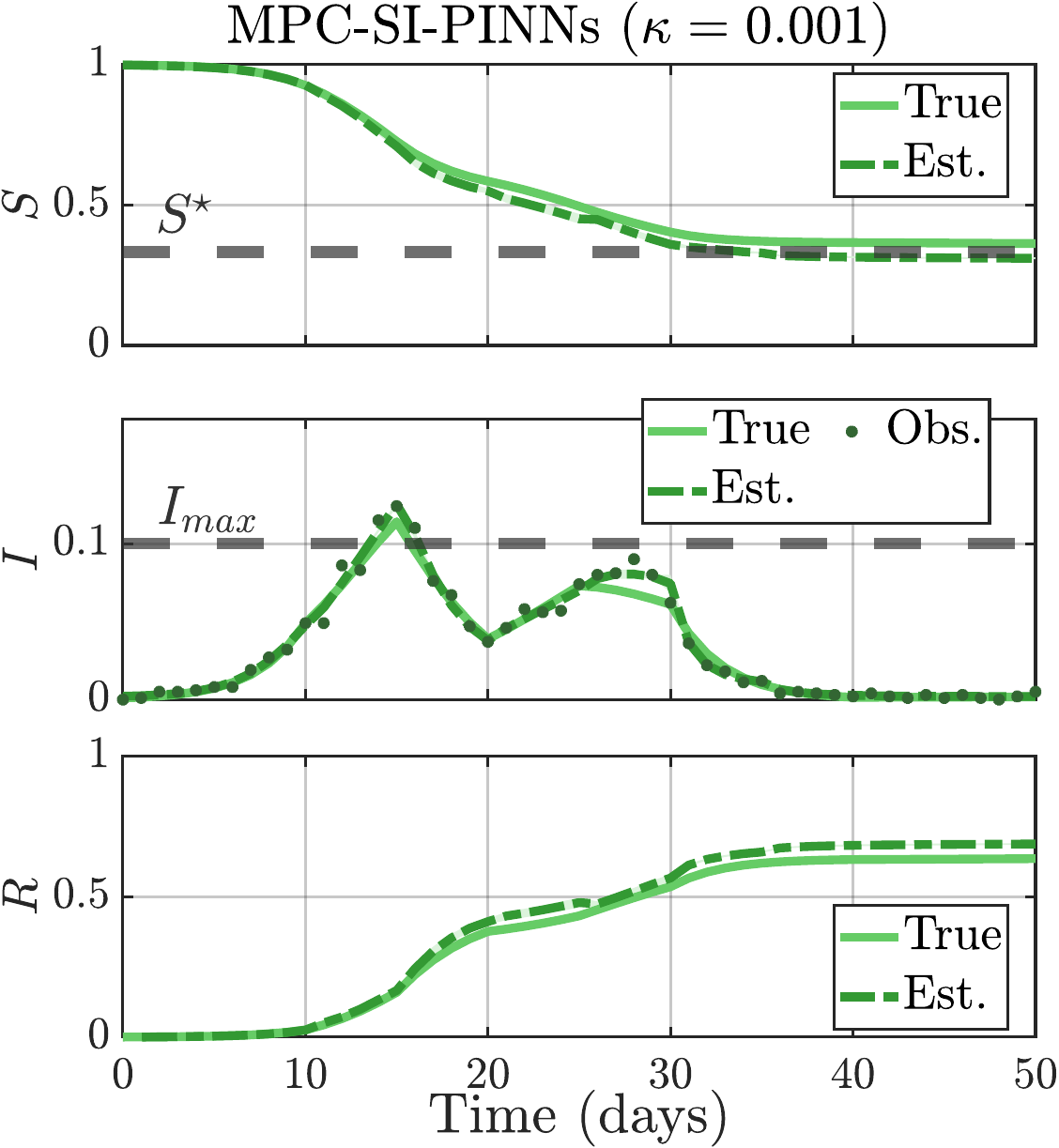}
        \caption{}
        \label{fig:PINNs_LS_SI_g}
    \end{subfigure}
    \hfill
    \begin{subfigure}[b]{0.24\linewidth}
        \includegraphics[width=\linewidth]{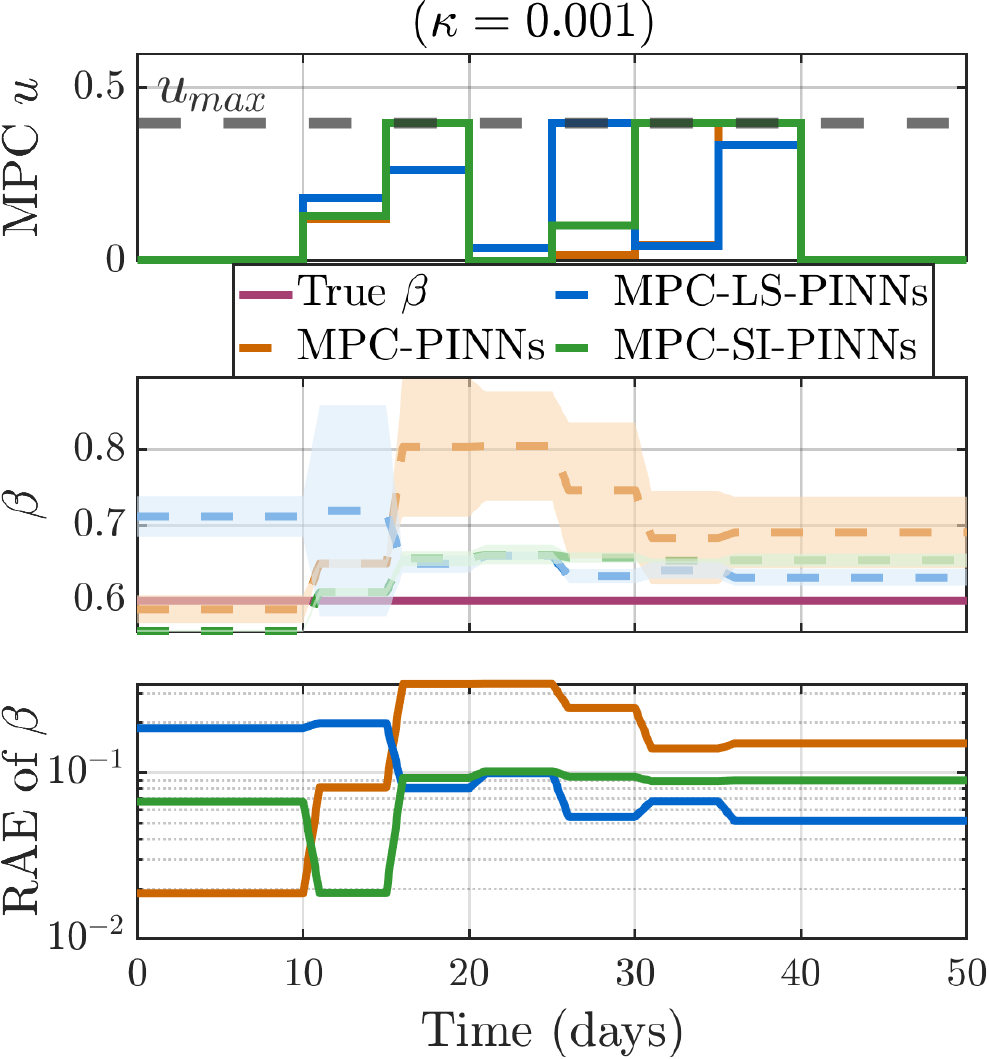}
        \caption{}
        \label{fig:PINNs_LS_SI_h}
    \end{subfigure}

    \caption{MPC-based control of epidemic dynamics using the MPC-PINNs, MPC-LS-PINNs, and MPC-SI-PINNs under noise with relatively low ($\kappa = 1$) and high variance ($\kappa = 0.001$) in \eqref{Poisson}: (a)–(c) and (e)–(g) present the true states $S,I,R$ and the estimates $\hat S, \hat I, \hat R$, and (d) and (h) show the control input $u$, the true transmission rate $\beta$, the estimate $\hat{\beta}$, and the RAE of $\hat{\beta}$. Dashed lines represent means, and shaded bands indicate one-standard-deviation intervals across multiple runs.}
    \label{fig:PINNs_LS_SI}
\end{figure*}


Table~\ref{tab:proposed_PINNs_summary} summarizes the proposed algorithms within the PINNs-based MPC framework. It highlights their main features and the conditions under which they are applicable. Sec.~\ref{sec4.1} then introduces the experimental data design and explains the rationale for adopting Poisson noise modeling.

Sec.~\ref{Sec7.1} evaluates MPC-LS-PINNs and MPC-SI-PINNs for the joint estimation of $S,I,R$ and $\beta$ under Assumption~\ref{assumption1} (with $\gamma$ known), considering both estimation accuracy and control performance. For these experiments, we use the MPC-PINNs from Sec.~\ref{Sec3.2} as the baseline. When the control input is set to zero, this baseline reduces to the algorithm in~\cite{millevoi2024physics}.

Sec.~\ref{Sec7.3} extends the comparison to the case where both $\beta$ and $\gamma$ are unknown under Assumption~\ref{assumption2} (with known $R_0$). Here, we use the generalized MPC-PINNs from Sec.~\ref{Sec6} as the baseline for comparison with the other proposed generalized algorithms. 
The experimental setup and code for this section are provided in the supporting material and supplementary code (available at \url{https://github.com/AipingZhong/PINNs-Based-MPC-for-SIR-Epidemics}).

We use the relative mean square error (rMSE) against a reference solution to evaluate our algorithms, defined as
\begin{equation*}
    \text{rMSE}_{\hat v}= \frac{\sum_{i=0}^{k}(\hat{v}(i) - v(i))^2}{\sum _{i=0}^{k}v(i)^2}. 
\end{equation*}
The $L_2$ relative error is 
\begin{equation*}
    \frac{1}{k}\sum^{k}_{i=0}
\frac{\left\| \hat y(i) - y(i)  \right\|_2}
{\left\|   y(i)  \right\|_2}, 
\end{equation*}
where $\hat y(i)$ is the solution vector at time step $i$, defined as $\hat y(i) = [\hat S(i), \hat I(i), \hat R(i), \hat \beta(i)]^\top$ when $\gamma$ is known, or $\hat y(i) = [\hat S(i), \hat I(i), \hat R(i), \hat \beta(i), \hat \gamma(i)]^\top$ when both $\beta$ and~$\gamma$ need to be estimated, where
 $i \in [0,1,\dots,k]$. For the parameters $\beta$ and~$\gamma$, we also use the relative absolute error (RAE), at time $k$, 
\begin{equation*}
    e_k = \frac{|\hat{v}(k) - v(k) |}{|v(k) |}. 
\end{equation*}
We evaluate the algorithms using the rMSE for the cumulative error of a single SISO NN, the $L_2$ relative error for the overall neural network performance, and the RAE for comparisons at specific time points.

We set a control period $T_p = 5$ days, during which the control strategy remains fixed within each period to reflect practical constraints. We train PINNs on a set of $N_{\tra}$ selected training time points, denoted by $\{k_1, k_2, \dots, k_{N_{\tra}}\}$. The value of $N_{\tra}$ varies across experiments, and the specific method for selecting these training points is described below. These training time points include the control start time, integer multiples of the control period $T_p$ (i.e., control start time $+ T_p$, $+ 2T_p$, ...) until reaching the control end time, and one manually-selected, final epidemic time near the end of the epidemic. 
To illustrate this setup, we take the experiments of Sec.~\ref{Sec7.1} as an example. Here, the control period is $T_p = 5$ days. Control is applied starting from Day $10$ up to Day $40$. A manually selected final epidemic time point at Day $50$ is also included for training. Thus, the set of $N_{\tra}$ selected training time points $\{k_1, k_2, \dots, k_{N_{\tra}}\}$ becomes 
$\{\text{Day } 10 \text{ (start)},\allowbreak\ 15,\allowbreak\ 20,\allowbreak\ 25,\allowbreak\ 30,\allowbreak\ 35,\allowbreak\ 40 \text{ (end of control)},\allowbreak\ 50 \text{ (final observation)}\}$.


\subsection{EXPERIMENTAL DATA DESIGN AND NOISE MODELING RATIONALE} \label{sec4.1}
This study focuses on dynamic model identification under a rolling control framework, where control inputs vary over time. Unlike static identification tasks, this setting requires estimating disease dynamics while accounting for continuously changing interventions. Real-world epidemic datasets (e.g., COVID-19 or influenza) typically correspond to specific historical control policies, making them unsuitable for testing algorithms under diverse or hypothetical strategies. Synthetic data, by contrast, allow systematic evaluation across a wide range of scenarios, which is essential for validating the proposed control-estimation closed-loop framework.

In this study, observation errors are modeled using the Poisson noise model in~\eqref{Poisson}, a standard choice~\cite{rashid2025poisson}. This model captures jump-like variations in reported cases, such as spikes from outbreak clusters or reporting changes, which Gaussian noise cannot represent due to its smooth, symmetric distribution. Its variance naturally scales with the mean, reflecting the greater uncertainty associated with higher case counts, whereas Gaussian noise assumes a constant variance.
\subsection{UNKNOWN $\beta$ UNDER NOISE WITH RELATIVELY LOW AND HIGH VARIANCE}\label{Sec7.1}

We first run the experiments shown in Fig.~\ref{fig:PINNs_LS_SI} under Assumption~\ref{assumption1}. We focus on two key aspects: (1) the estimation performance of the MPC-PINNs, MPC-LS-PINNs, and MPC-SI-PINNs and (2) the control performance under the three PINNs-based MPC frameworks. We first discuss the estimation performance across different noise levels, followed by the analysis of the corresponding controllers.

\textbf{Estimation Performance.} 
We first compare the MPC-PINNs, MPC-LS-PINNs, and MPC-SI-PINNs under Poisson noise with relatively low variance ($\kappa = 1$) and high variance ($\kappa = 0.001$) as defined in~\eqref{Poisson}. Overall, the MPC-SI-PINNs perform best under low noise, the MPC-LS-PINNs outperform the MPC-PINNs once control is applied, and the MPC-PINNs are reliable only before control is introduced. High-variance noise degrades all the methods, with the MPC-SI-PINNs particularly affected by the error propagation from the integral operation.

For the MPC-PINNs, the state estimates closely match the true trajectories before control ($t \leq 10$) for both $\kappa = 1$ and $\kappa = 0.001$, with $L_2$ relative errors of $2.88\times10^{-2}$ and $2.26\times10^{-2}$, respectively. After control begins, the estimation variance increases in $\hat S$, $\hat I$, and $\hat R$, and $\hat{\beta}$ fluctuates significantly under $\kappa = 0.001$ (Figs.~\ref{fig:PINNs_LS_SI_e}, \ref{fig:PINNs_LS_SI_h}), reducing the reliability of the MPC.

For the MPC-LS-PINNs, pre-control accuracy is slightly lower than the MPC-PINNs. Under $\kappa = 1$, the rMSE for $\hat{\beta}$ is $6.71\times 10^{-4}$ compared to $3.03\times 10^{-5}$ for the MPC-PINNs, and the $L_2$ relative error is $3.19\times10^{-2}$ vs. $2.88\times10^{-2}$. Once control is applied, the MPC-LS-PINNs reduce bias more effectively than the MPC-PINNs, and the estimation variance decreases over time. The estimates converge to the true trajectories after $t \geq 25$ for both $\kappa = 1$ and $\kappa = 0.001$.

For the MPC-SI-PINNs, under $\kappa = 1$ the method achieves the most consistent tracking across all states and parameters, with an $L_2$ relative error of $1.37\times 10^{-2}$. Under $\kappa = 0.001$, however, errors in $\hat I$ propagate via the integral operation, causing $\hat S$ and $\hat R$ to drift away from the true trajectories for $t\ge 36$ (Fig.~\ref{fig:PINNs_LS_SI_g}). Table~\ref{tab:comparison_LS_PINNs_SI} summarizes the performance of the MPC-PINNs, MPC-LS-PINNs, and MPC-SI-PINNs over $t \in [0, 50]$ under Poisson noise with relatively low variance ($\kappa = 1$) and high variance ($\kappa = 0.001$) as defined in~\eqref{Poisson}.

\begin{table}[!t]
    \centering
    \caption{Performance comparison of the MPC-PINNs (Sec.~\ref{Sec3.2}), MPC-LS-PINNs (Sec.~\ref{Sec4}), and MPC-SI-PINNs (Sec.~\ref{Sec5}) under noise with relatively low ($\kappa = 1$) and high ($\kappa = 0.001$) variance in~\eqref{Poisson}.}
    \label{tab:comparison_LS_PINNs_SI}
    \resizebox{\linewidth}{!}{
    \begin{tabular}{@{}p{0.6cm}|@{\hskip 2pt}l@{\hskip 4pt}c@{\hskip 4pt}c@{\hskip 4pt}c@{\hskip 4pt}c@{\hskip 4pt}c@{}}
    \toprule
    $\kappa$ & \textbf{Method} & $\text{rMSE}_{\hat{S}}$ & $\text{rMSE}_{\hat{I}}$ & $\text{rMSE}_{\hat{R}}$ & $\text{rMSE}_{\hat{\beta}}$ & $L_2$ rel. error \\
    \midrule
    \multirow{3}{*}{$1$}
        & MPC-PINNs     & $5.97e{-2}$ & $1.41e{-1}$ & $7.85e{-2}$ & $2.23e{-2}$ & $1.81e{-1}$ \\
        & MPC-LS-PINNs        & $5.43e{-3}$ & $4.96e{-4}$ & $1.05e{-2}$ & $6.47e{-3}$ & $6.54e{-2}$ \\
        & \textbf{MPC-SI-PINNs} & $\mathbf{1.44e{-4}}$ & $\mathbf{1.08e{-4}}$ & $\mathbf{2.39e{-4}}$ & $\mathbf{2.77e{-4}}$ & $\mathbf{1.37e{-2}}$ \\
    \midrule
    \multirow{3}{*}{$0.001$}
        & MPC-PINNs     & $1.11e{-1}$ & $1.61e{-1}$ & $1.60e{-1}$ & $3.79e{-2}$ & $2.65e{-1}$ \\
        & MPC-LS-PINNs        & $1.58e{-2}$ & $\mathbf{4.49e{-3}}$ & $3.11e{-2}$ & $1.44e{-2}$ & $1.15e{-1}$ \\
        & \textbf{MPC-SI-PINNs} & $\mathbf{3.44e{-3}}$ & $1.01e{-2}$ & $\mathbf{6.65e{-3}}$ & $\mathbf{6.89e{-3}}$ & $\mathbf{7.15e{-2}}$ \\
    \bottomrule
    \end{tabular}
    }
\end{table}


\begin{table}
    \centering
    \caption{Performance comparison of the MPC-PINNs, MPC-LS-PINNs, and MPC-SI-PINNs under noise with relatively high ($\kappa = 0.001$) variance in~\eqref{Poisson} over different time intervals. The upper part reports results for $t \in [0, 35]$ (early-to-mid stage), and the lower part for $t \in [36, 50]$ (late stage).}
    \label{tab:kappa001_timed_comparison}
      \resizebox{\linewidth}{!}{\begin{tabular}{@{}lccccc@{}}
    \toprule
    \multicolumn{6}{c}{\textbf{Time Interval: $t \in [0, 35]$}} \\
    \midrule
    \textbf{Method} & $\text{rMSE}_{\hat{S}}$ & $\text{rMSE}_{\hat{I}}$ & $\text{rMSE}_{\hat{R}}$ & $\text{rMSE}_{\hat{\beta}}$ & $L_2$ rel. error \\
    \midrule
    MPC-PINNs& $3.44e{-2}$ & $1.61e{-1}$ & $9.78e{-2}$ & $4.43e{-2}$ & $1.78e{-1}$ \\
    MPC-LS-PINNs  & $1.68e{-2}$ & $\mathbf{4.37e{-3}}$ & $6.83e{-2}$ & $1.93e{-2}$ & $1.39e{-1}$ \\
    \textbf{MPC-SI-PINNs}  & $\mathbf{1.67e{-3}}$ & $1.01e{-2}$ & $\mathbf{6.68e{-3}}$ & $\mathbf{6.42e{-3}}$ & $\mathbf{6.13e{-2}}$ \\
    \midrule
    \multicolumn{6}{c}{\textbf{Time Interval: $t \in [36, 50]$}} \\
    \midrule
    \textbf{Method} & $\text{rMSE}_{\hat{S}}$ & $\text{rMSE}_{\hat{I}}$ & $\text{rMSE}_{\hat{R}}$ & $\text{rMSE}_{\hat{\beta}}$ & $L_2$ rel. error \\
    \midrule
    MPC-PINNs& $1.42e{+0}$ & $1.80e{-1}$ & $1.95e{-1}$ & $2.26e{-2}$ & $4.73e{-1}$ \\
    \textbf{MPC-LS-PINNs}  & $\mathbf{6.69e{-3}}$ & $\mathbf{3.65e{-2}}$ & $\mathbf{2.35e{-3}}$ & $\mathbf{2.63e{-3}}$ & $\mathbf{5.60e{-2}}$ \\
    MPC-SI-PINNs  & $2.03e{-2}$ & $6.16e{-2}$ & $6.62e{-3}$ & $8.02e{-3}$ & $9.60e{-2}$ \\
    \bottomrule
    \end{tabular}}
\end{table}\vspace{-0.1cm}

\begin{figure*}
    \centering
    \begin{subfigure}[b]{0.235\linewidth}
        \includegraphics[width=\linewidth]{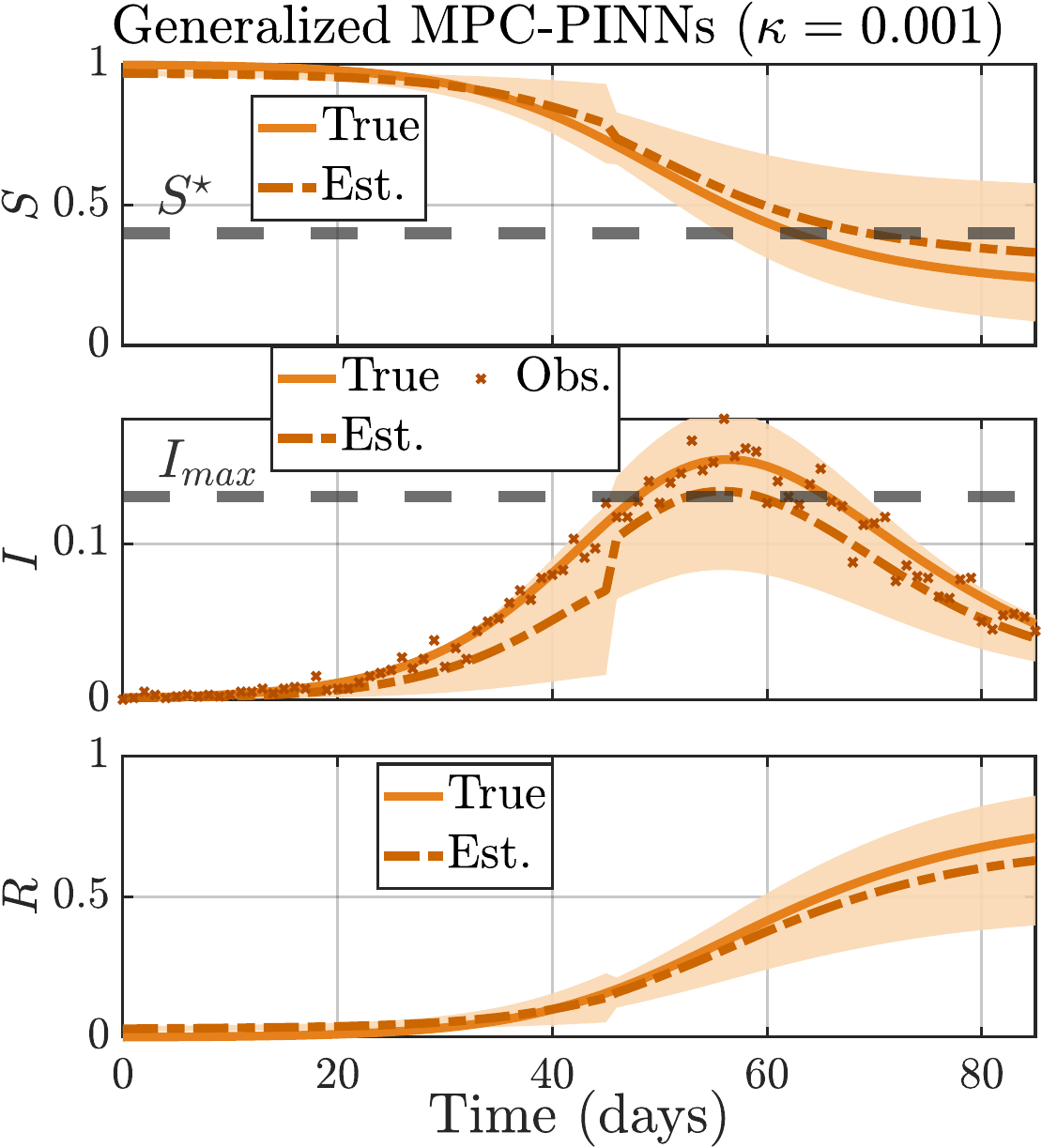}
        \caption{}
        \label{fig:generalized_PINNs_comparison_a}
    \end{subfigure}
    \hfill
    \begin{subfigure}[b]{0.235\linewidth}
        \includegraphics[width=\linewidth]{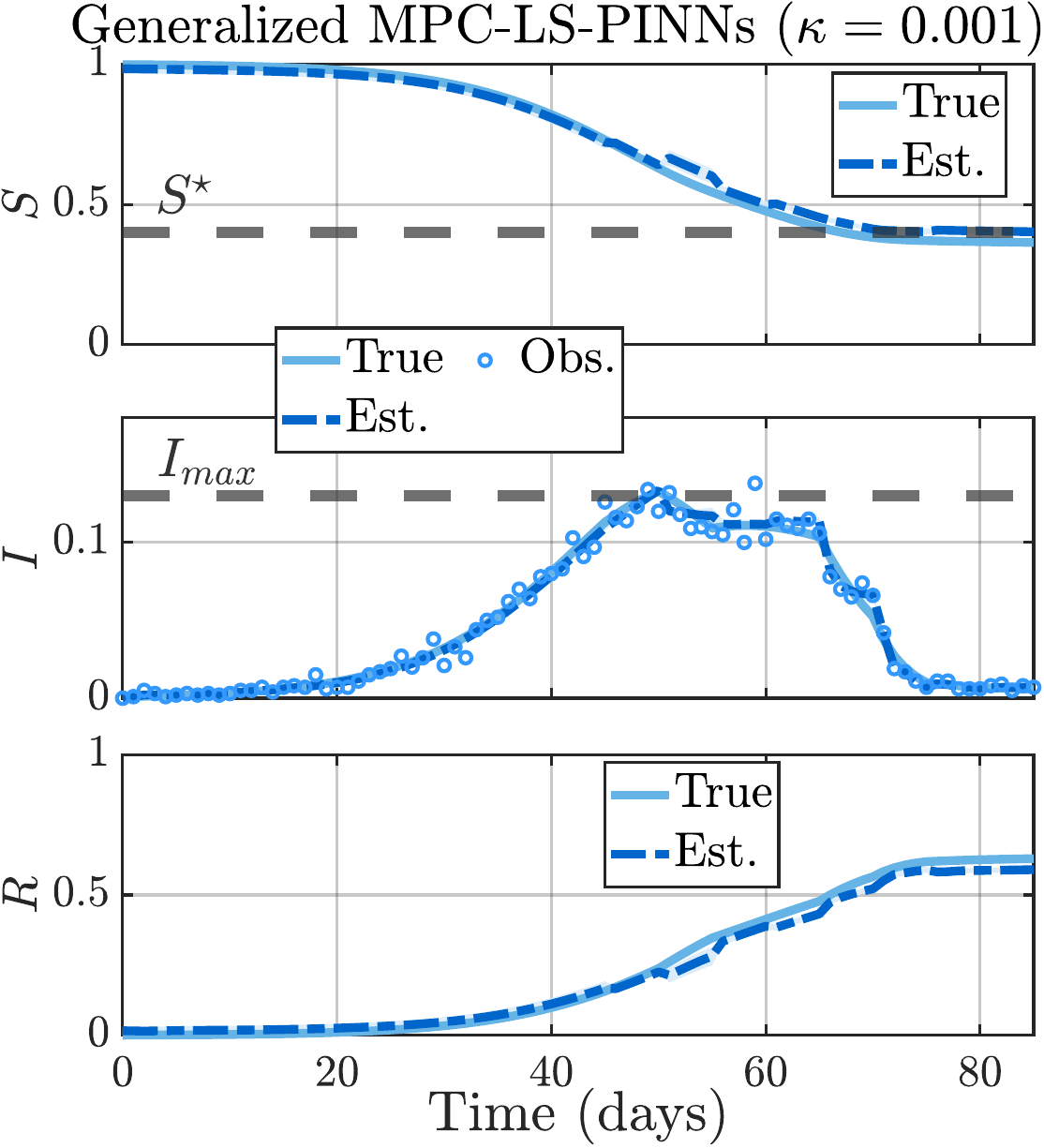}
        \caption{}
        \label{fig:generalized_PINNs_comparison_b}
    \end{subfigure}
    \hfill
    \begin{subfigure}[b]{0.235\linewidth}
        \includegraphics[width=\linewidth]{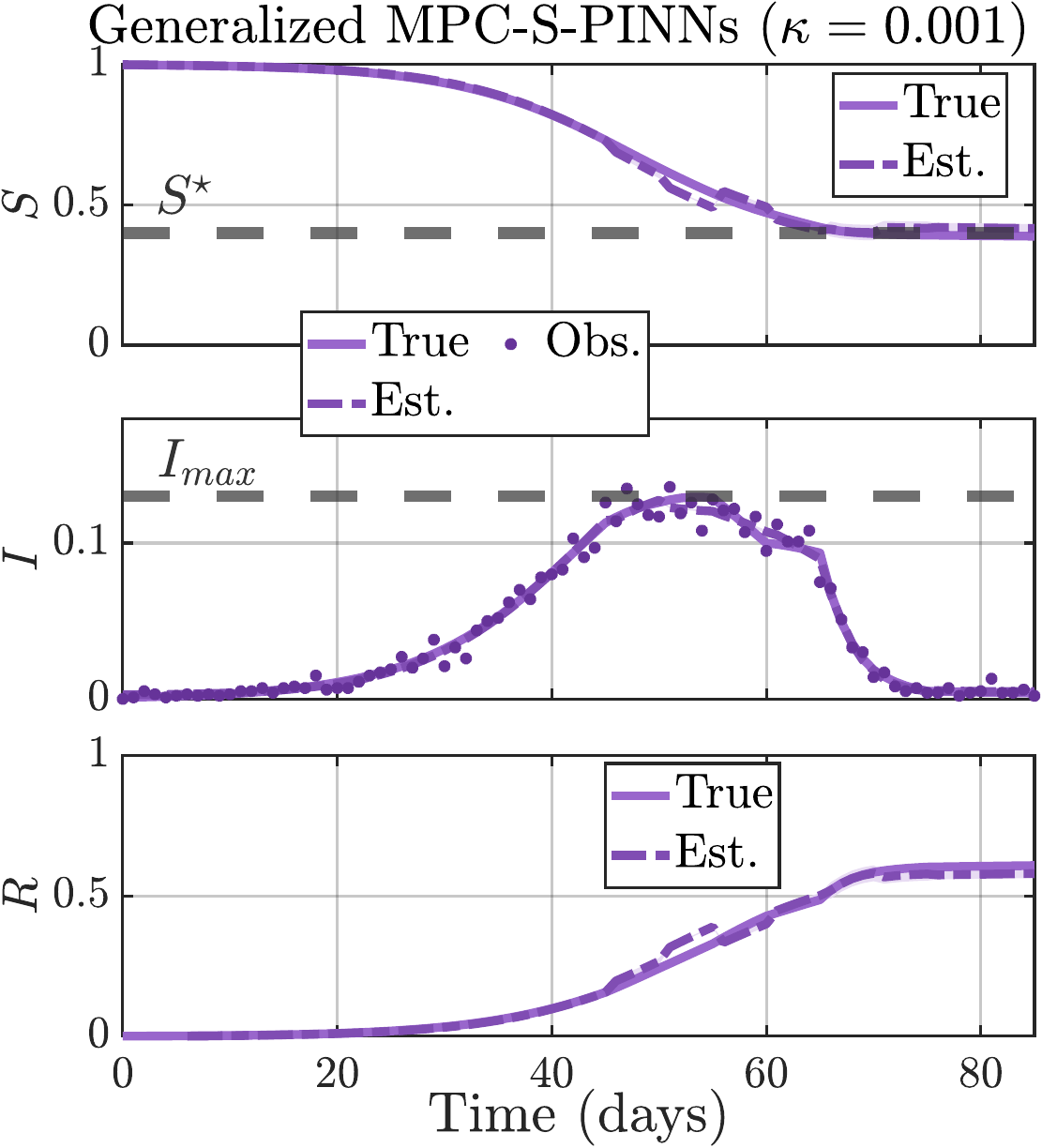}
        \caption{}
        \label{fig:generalized_PINNs_comparison_c}
    \end{subfigure}
    \hfill
    \begin{subfigure}[b]{0.26\linewidth}
        \includegraphics[width=\linewidth]{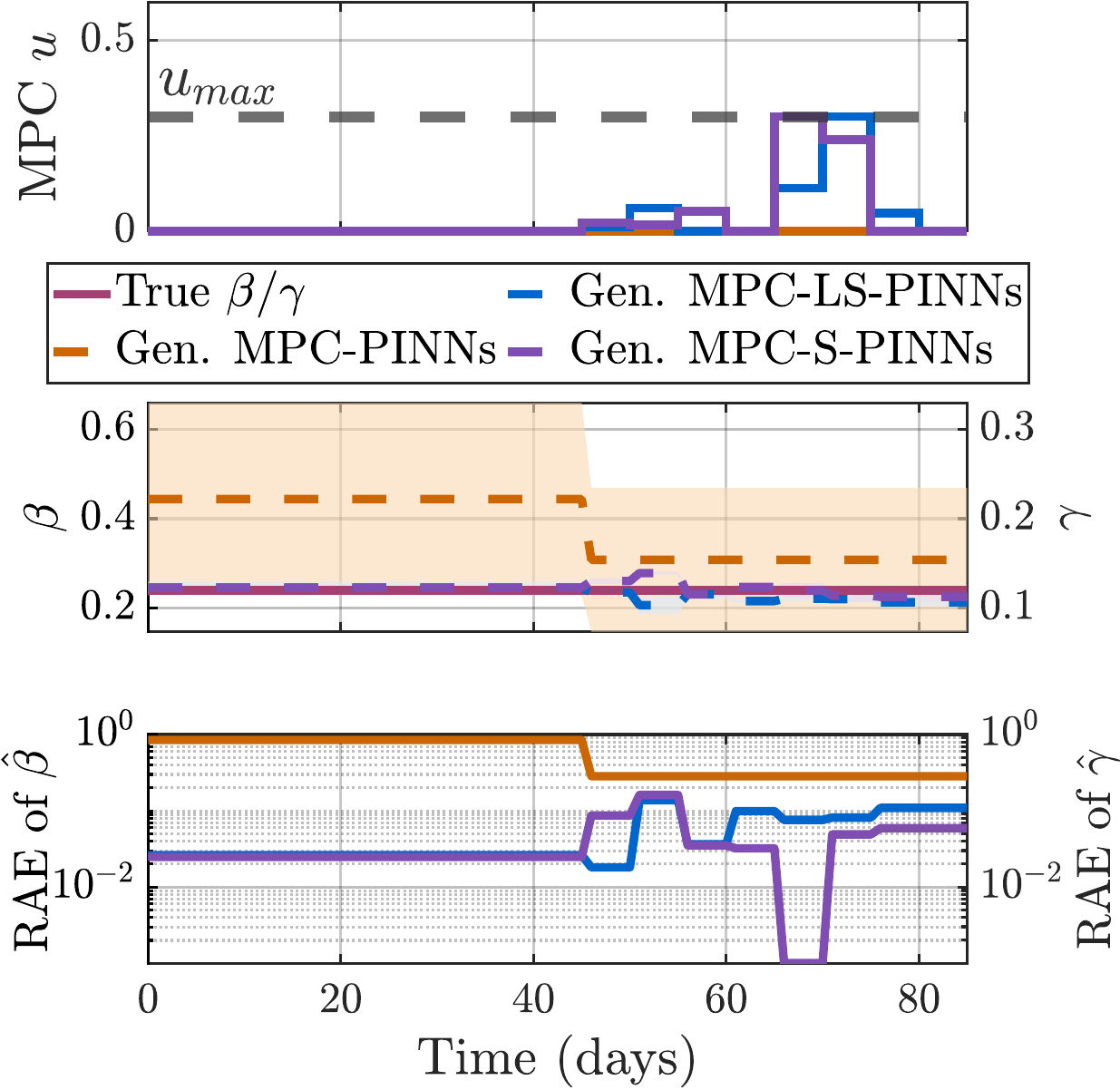}
        \caption{}
        \label{fig:generalized_PINNs_comparison_d}
    \end{subfigure}

    \caption{MPC-based control of epidemic dynamics using the  generalized MPC-PINNs, MPC-LS-PINNs, and MPC-S-PINNs in the PINNs-based MPC framework under noise with relatively high ($\kappa = 0.001$) variance in~\eqref{Poisson}: (a)–(c) present the true states $S, I, R$, and state estimates $\hat S, \hat I, \hat R$; (d) shows the MPC-derived control input $u$, the true parameters $\beta$ and $\gamma$, and the parameter estimates $\hat{\beta}$ and $\hat{\gamma}$, along with the RAE of $\hat{\beta}$ and $\hat{\gamma}$ across time. Dashed lines represent mean estimates, and shaded bands indicate one-standard-deviation confidence intervals over multiple runs.}
    \label{fig:generalized_PINNs_comparison}
\vspace{-0.05cm}\end{figure*}

To further compare the estimation performance across time, Table~\ref{tab:kappa001_timed_comparison} summarizes the estimation errors during the early-to-mid stage ($t \in [0, 35]$) and the late stage ($t \in [36, 50]$) under noise with relatively high ($\kappa = 0.001$) variance in~\eqref{Poisson}. The data reveals distinct, phase-dependent advantages for the proposed methods.
In the early stage ($t \in [0, 35]$), the MPC-SI-PINNs outperform both the MPC-LS-PINNs and the MPC-PINNs in estimating $\hat S$, $\hat R$, and $\hat{\beta}$. The rMSE values are reduced by about $0.5$–$1$ order of magnitude compared to MPC-LS-PINNs. These results indicate that enriching the physics-informed training with derived data (i.e., $S^{\mathrm{d}}$ and $R^{\mathrm{d}}$ obtained via the integral operation in Algorithm~\ref{algorithmic2}) in the MPC-SI-PINNs leads to faster convergence and improved estimation accuracy.  However, the MPC-SI-PINNs' $\hat I$ estimates have an rMSE about $0.36$ orders of magnitude higher than the MPC-LS-PINNs, due to the absence of ODE residual regularization.

The performance dynamics shift in the late stage ($t \in [36, 50]$). the MPC-LS-PINNs yield the lowest errors across all variables, reducing rMSE by roughly $0.2$–$0.5$ orders of magnitude relative to the MPC-SI-PINNs. The performance drop in the MPC-SI-PINNs during this stage stems from error propagation in the ``integral operation'' step (Algorithm~\ref{algorithmic2}), where inaccuracies in $\hat{I}(t)$ accumulate in the derived states $S^{\mathrm{d}}(t)$ and $R^{\mathrm{d}}(t)$, degrading subsequent predictions. The success of the MPC-LS-PINNs in this noisy, data-rich environment is significant. It confirms that the real-time data updating mechanism in the PINNs-based MPC framework enables continuous adaptation and improved estimation accuracy. Such performance is possible only when the learning algorithm is sufficiently robust, a capability the baseline MPC-PINNs lacks. 

Overall, both the MPC-LS-PINNs and MPC-SI-PINNs substantially improve over MPC-PINNs. 
In the early stage, $L_2$ relative errors are reduced by approximately $65.6\%$ (MPC-SI-PINNs) and $21.9\%$ (MPC-LS-PINNs) relative to MPC-PINNs. In the late stage, reductions are $88.2\%$ (MPC-LS-PINNs) and $79.7\%$ (MPC-SI-PINNs). Thus, MPC-SI-PINNs are more effective with limited data, while MPC-LS-PINNs gain advantage as more data become available.

\begin{table}[!t]
    \centering
    \caption{Comparison of control performance for the MPC-PINNs (Sec.~\ref{Sec3.2}), MPC-LS-PINNs (Sec.~\ref{Sec4}), and MPC-SI-PINNs (Sec.~\ref{Sec5}) under noise with relatively low ($\kappa = 1$) and high ($\kappa = 0.001$) variance in~\eqref{Poisson}. Indices include the infection peak $I_{\max}$, the CCT, the EET, and the final susceptible proportion $S_{\text{final}}$. Empty entries are denoted by ‘---’.}
    \label{tab:control_performance}
      \resizebox{\linewidth}{!}{\begin{tabular}{@{}c|lcccc@{}}
    \toprule
    $\kappa$ & \textbf{Method} & $I_{\max}$ & CCT& EET & $S_{\text{final}}$ \\
    \midrule
    \multirow{3}{*}{$1$}
        & MPC-PINNs & $0.1245$ & $6.3938$ & $37.00$ & $0.2660$ \\
        & MPC-LS-PINNs    & $0.0961$ & $4.8497$ & ---      & --- \\
        & \textbf{MPC-SI-PINNs} & $0.1024$ & $6.1801$ & $37.00$ & $0.3349$ \\
    \midrule
    \multirow{3}{*}{$0.001$}
        & MPC-PINNs & $0.1193$ & $4.9448$ & ---      & --- \\
        & MPC-LS-PINNs    & $0.0955$ & $6.3382$ & $48.00$  & $0.3664$ \\
        & \textbf{MPC-SI-PINNs} & $0.1140$ & $7.1782$ & $40.00$  & $0.3663$ \\
    \bottomrule
    \end{tabular}}
    \end{table} 

\textbf{Control Performance.} We evaluate the control performance using four indices: (1) infection peak $I_{\max}$, (2) epidemic end time (EET), (3) final susceptible proportion $S_{\text{final}}$, and (4) cumulative control cost (CCT). Since safety constraints are not included in our MPC design, we impose a $3\%$ tolerance margin for the $I_{\max}$ target, requiring $I_{\max} < 1.03 I^\star_{\max} = 0.1030$. EET is defined as the first time $I_{\text{true}} < 0.0050$, provided $S_{\text{true}} - S^\star < 0.03 S^\star$ for $\kappa = 1$ or $S_{\text{true}} - S^\star < 0.10 S^\star$ for $\kappa = 0.001$, ensuring the epidemic is deemed over only when $S$ is near the herd immunity threshold. $S_{\text{final}}$ denotes the  uninfected proportion at the epidemic’s end, with higher values indicating better outcomes.

Table~\ref{tab:control_performance} compares the control performance of the MPC-PINNs, MPC-LS-PINNs, and MPC-SI-PINNs under relative low ($\kappa = 1$) and high ($\kappa = 0.001$) noise. For $\kappa = 1$, the MPC-LS-PINNs and MPC-SI-PINNs satisfy the $I_{\max}$ constraint ($\leq 0.1030$), while the MPC-PINNs exceed it. the MPC-LS-PINNs achieve the lowest CCT but do not reach the epidemic end, whereas the MPC-SI-PINNs end earlier ($t=37$) with higher $S_{\text{final}}$ (0.3349) and lower CCT than the MPC-PINNs. For $\kappa = 0.001$, the MPC-LS-PINNs achieve the best $I_{\max}$ (0.0955) and highest $S_{\text{final}}$ (0.3664) but later EET ($t=48$), while the MPC-SI-PINNs end earlier ($t=40$) but with the highest CCT (7.1782) and $I_{\max}$ above the constraint (0.1140). The latter is linked to early-stage overestimation of $\hat{\beta}(t)$ in the MPC-LS-PINNs, leading to a more conservative control that offsets other estimation errors. Overall, higher estimation accuracy generally improves control outcomes: the MPC-SI-PINNs perform best for $\kappa = 1$, while the MPC-LS-PINNs are more robust for $\kappa = 0.001$.

\subsection{UNKNOWN $\beta$ AND $\gamma$ UNDER NOISE WITH RELATIVELY HIGH VARIANCE}\label{Sec7.3}
In this section, we present the experiments in Fig.~\ref{fig:generalized_PINNs_comparison} under Assumption~\ref{assumption2}, evaluating the generalized MPC-PINNs, generalized MPC-LS-PINNs, and generalized MPC-S-PINNs with high noise variance ($\kappa = 0.001$) in~\eqref{Poisson}, when both $\beta$ and $\gamma$ are unknown. This experiment assesses the estimation accuracy of $\hat S$, $\hat I$, and $\hat R$, the parameters $\hat \beta$ and $\hat \gamma$, and the control performance achieved by each PINNs-based MPC framework.

\textbf{Estimation performance.} We analyze the estimation performance under high noise variance ($\kappa = 0.001$) in Fig.~\ref{fig:generalized_PINNs_comparison}. For generalized MPC-PINNs (Fig.~\ref{fig:generalized_PINNs_comparison_a}), large errors appear even before the control begins ($t \leq 45$), with rMSE values of $1.43 \times 10^{-1}$ for $\hat{I}(t)$, $1.91 \times 10^{-1}$ for $\hat{R}(t)$, and $8.22 \times 10^{-4}$ for $\hat{S}(t)$. the parameter estimates $\hat{\beta}(t)$ and $\hat{\gamma}(t)$ also deviate substantially (rMSE $= 7.21 \times 10^{-1}$), resulting in an overall $L_2$ relative error of $1.86 \times 10^{-1}$. These inaccuracies lead the corresponding MPC framework to produce zero control inputs throughout (Fig.~\ref{fig:generalized_PINNs_comparison_d}), and errors persist even after control starts ($t \geq 46$).  

In contrast, the generalized MPC-LS-PINNs (Fig.~\ref{fig:generalized_PINNs_comparison_b}) and generalized MPC-S-PINNs (Fig.~\ref{fig:generalized_PINNs_comparison_c}) yield accurate and stable estimates of $\hat{S}(t)$, $\hat{I}(t)$, and $\hat{R}(t)$ across the entire horizon, with only mild fluctuations after $t \geq 46$. During the controlled stage, the generalized MPC-LS-PINNs achieve a $75.8\%$ lower $L_2$ relative error than the generalized MPC-PINNs ($4.50 \times 10^{-2}$ vs. $1.86 \times 10^{-1}$), while the generalized MPC-S-PINNs improve it by $85.0\%$ ($2.80 \times 10^{-2}$). For $\hat{\beta}(t)$ and $\hat{\gamma}(t)$, the generalized MPC-LS-PINNs perform slightly better up to $t \leq 60$, but the generalized MPC-S-PINNs achieve higher accuracy for $t \geq 61$.

\begin{table}
    \centering
    \caption{Performance comparison of the generalized MPC-PINNs, MPC-LS-PINNs, and MPC-S-PINNs proposed in Sec.~\ref{Sec6}, under noise with relatively high variance ($\kappa = 0.001$) in \eqref{Poisson} over $t \in [0, 85]$.}
    \label{tab:comparison_LS_Split_PINNs}
    \resizebox{\linewidth}{!}{
    \begin{tabular}{@{}l@{\hskip 4pt}c@{\hskip 4pt}c@{\hskip 4pt}c@{\hskip 4pt}c@{\hskip 4pt}c@{}}
    \toprule
    \textbf{Method} & $\text{rMSE}_{\hat{S}}$ & $\text{rMSE}_{\hat{I}}$ & $\text{rMSE}_{\hat{R}}$ & $\text{rMSE}_{\hat{\beta}}$ & $L_2$ rel. error \\
    \midrule
    Gen. MPC-PINNs     & $4.58e{-3}$ & $3.92e{-2}$ & $1.30e{-2}$ & $4.23e{-1}$ & $1.86e{-1}$ \\
    Gen. MPC-LS-PINNs        & $1.21e{-3}$ & $2.61e{-3}$ & $6.50e{-3}$ & $4.24e{-3}$ & $4.50e{-2}$ \\
    \textbf{Gen. MPC-S-PINNs} & $\mathbf{6.17e{-4}}$ & $\mathbf{1.41e{-3}}$ & $\mathbf{3.66e{-3}}$ & $\mathbf{2.97e{-3}}$ & $\mathbf{2.80e{-2}}$ \\
    \bottomrule
    \end{tabular}
    }
\end{table}
Table~\ref{tab:comparison_LS_Split_PINNs} further shows that under $\kappa = 0.001$, the generalized MPC-PINNs produce near-zero control and large estimation errors. the generalized MPC-LS-PINNs improve accuracy by approximately one order of magnitude, while the generalized MPC-S-PINNs, benefiting from split training, achieve $1.6$ times lower $L_2$ error than the MPC-LS-PINNs.

\textbf{Control Performance.}
Table~\ref{tab:control_performance_generalized} summarizes the control performance under high-variance noise ($\kappa = 0.001$) with both $\beta$ and $\gamma$ unknown. For the infection peak $I_{\max}$, both the generalized MPC-LS-PINNs ($0.1329$) and MPC-S-PINNs ($0.1295$) satisfy the $3\%$ tolerance constraint ($\leq 0.1339$), while MPC-PINNs exceed it ($0.1539$). In terms of CCT, the MPC-S-PINNs incur the highest cost ($3.1483$), the MPC-LS-PINNs are lower ($2.6496$), and the MPC-PINNs have zero cost, which reflects ineffective control rather than optimality due to poor estimation accuracy. For EET, the MPC-S-PINNs suppress the epidemic earlier ($t = 75$) than the MPC-LS-PINNs ($t = 84$), while the MPC-PINNs fail to meet the epidemic termination criterion ($I(t)$ never drops below $0.005$).

For the final susceptible proportion $S_{\text{final}}$, the MPC-S-PINNs achieve the highest value ($0.3922$), followed by the MPC-LS-PINNs ($0.3651$), both outperforming the MPC-PINNs. In the case of MPC-PINNs, the epidemic is not suppressed over the experiment horizon: as shown in the middle panel of Fig.~\ref{fig:generalized_PINNs_comparison}, $I(t)$ remains nonzero at $t=85$; and from the upper panel, one can see that $S(t)$ is still decreasing at $t=85$. Hence, a final $S_{\text{final}}$ cannot be determined in the fixed simulation length. Overall, the MPC-S-PINNs strike the best balance between early epidemic mitigation and preserving susceptibles, albeit at a higher control cost, while the MPC-LS-PINNs offer more conservative but still effective control. the MPC-PINNs perform poorly under high-variance noise, underscoring the need for improved estimation accuracy to enable reliable PINNs-based MPC.

\begin{table}
    \centering
    \caption{Comparison of control performance for the generalized MPC-PINNs, MPC-LS-PINNs, and MPC-S-PINNs, as described in Sec.~\ref{Sec6}, under noise with relatively high ($\kappa=0.001$) variance in~\eqref{Poisson} with unknown $\beta$ and $\gamma$. Indices include the constraint of infection peak $I_{\max}$, the CCT, the EET, and the final susceptible proportion $S_{\text{final}}$. Empty entries are denoted by ‘---’.}
    \label{tab:control_performance_generalized}
        \resizebox{\linewidth}{!}{\begin{tabular}{@{}c|lcccc@{}}
    \toprule
    $\kappa$ & \textbf{Method} & $I_{\max}$ & CCT & EET & $S_{\text{final}}$ \\
    \midrule
    \multirow{3}{*}{$0.001$}
        & Gen. MPC-PINNs& $0.1539$ & $0.0000$ & ---     & --- \\
        & Gen. MPC-LS-PINNs& $0.1329$ & $2.6496$ & $84.00$ & $0.3651$ \\
        & Gen. MPC-S-PINNs& $0.1295$ & $3.1483$ & $75.00$ & $0.3922$ \\
    \bottomrule
    \end{tabular}}
\end{table}

\section{CONCLUSION} 
\label{section5}
This paper presents a PINNs-based MPC closed-loop framework for real-time epidemic control under noisy infected-state observations, introducing two novel algorithms: MPC-LS-PINNs and MPC-SI-PINNs. the MPC-LS-PINNs enhance noise resilience through a log-relative error term, while the MPC-SI-PINNs decouple data regression from physics-informed training, leveraging integral-derived states to improve convergence. We further generalize these frameworks to the case where both the transmission and recovery rates are unknown. Experimental results demonstrate that the MPC-LS-PINNs and MPC-SI-PINNs significantly improve estimation accuracy over the MPC-PINNs across different noise levels. the MPC-SI-PINNs perform best in early epidemic stages with limited data, whereas the MPC-LS-PINNs excel in later stages as more data become available. When both $\beta$ and $\gamma$ are unknown, the generalized MPC-S-PINNs formulation further improves performance by decoupling data regression and physics-informed training. Future work will extend these methods to other models of spreading dynamics.

\section*{APPENDIX}

\subsection{Proofs of Theorem \ref{Theorem} and Corollary \ref{corollary}} \label{Appendix_Theorem_corollary}
\noindent \textbf{Statement of Theorem \ref{Theorem}.}
Consider the system in~\eqref{1} with initial conditions given in~\eqref{initial2}. Suppose the trajectories of the infected proportion $I(t)$ and control input $u(t)$ are known for $t \in [0, k]$. Then, there exists a family of  solutions $\{(\beta^\dagger, \gamma^\dagger, S^\dagger(t), R^\dagger(t))\}$ that satisfy~\eqref{1} over $[0, k]$.

\begin{proof}
From~\eqref{1}, we obtain
\begin{equation}
    S(t) = \frac{\frac{dI(t)}{dt} + (\gamma+u(t))I(t)}{\beta I(t)}.  \label{proof1}
\end{equation}
Differentiating both sides of the equation in \eqref{1b} with respect to $t$ yields
\begin{equation}
    \frac{d^2 I(t)}{dt^2} = \beta \frac{d}{dt} (I(t) S(t)) -  (\gamma+u) \frac{dI(t)}{dt}. \label{proof2}
\end{equation}
Rewrite $\frac{d}{dt} (I(t) S(t))$ in terms of $I(t)$ and $\frac{dI(t)}{dt}$ as
\begin{equation}
    \frac{d}{dt} (I(t) S(t)) = S(t) \frac{dI(t)}{dt} + I(t) \frac{dS(t)}{dt}. \label{proof3}
\end{equation}
Substituting \eqref{1a} into \eqref{proof3} gives
\begin{equation}
\begin{aligned}
    \frac{d}{dt} (I(t) S(t)) &= S(t) \frac{dI(t)}{dt} + I(t) (- \beta I(t) S(t)) \\
    &   = S(t) \left[ \frac{dI(t)}{dt} - \beta I(t)^2 \right].
\end{aligned} \label{proof4}
\end{equation}
By substituting  \eqref{proof1} into  \eqref{proof4} and subsequently incorporating the obtained result into  \eqref{proof2}, we can derive the following expression
\begin{equation}
\begin{aligned}
    \frac{d^2 I(t)}{dt^2} = - \beta (\gamma+u(t))I(t)^2 
    -  \beta I(t) \frac{dI(t)}{dt} + \frac{\left(\frac{dI(t)}{dt}\right)^2}{I(t)}.
\end{aligned} \label{proof5}
\end{equation}
According to \eqref{proof5}, if the trajectories of $I(t)$ and $u(t)$ are known for $t \in [0, k]$, for any given $\beta^	\dagger$, there exists a unique $\gamma^	\dagger$ satisfying
\begin{equation}
    \gamma^	\dagger\! = \!-\frac{1}{\beta^	\dagger  I(t)^2}\!\! \left[ \!\frac{d^2 I(t)}{dt^2} \!+\!  \beta^	\dagger I(t) \frac{dI(t)}{dt} \!-\! \frac{\left(\frac{dI(t)}{dt}\right)^2}{I(t)} \!\right]\!\! - u(t).\label{proof6}
\end{equation}
The system in~(1) can be written in vector form as 
\[
\dot{\chi}(t) = F(\chi(t), u(t), \beta, \gamma),
\]
where \(\chi(t) = [S(t), I(t), R(t)]^\top\). Since \(F(\cdot)\) is continuously differentiable in \(\chi(t)\) and parameters, and \(u(t)\) is piecewise continuous, given $\beta^\dagger, \gamma^\dagger$ and the initial conditions in \eqref{initial2}, by the Picard–Lindel\"{o}f Theorem~\cite{hartman2002ordinary}, a unique solution of $S^\dagger(t)$ and $R^\dagger(t)$ exists locally for $t$ in the interval $[0, k]$. Thus, the system admits a family of solutions $\{\beta^	\dagger, \gamma^	\dagger, S^	\dagger(t), R^	\dagger(t)\}$ for $t \in [0, k]$.
\end{proof}

\noindent \textbf{Statement of Corollary \ref{corollary}.}
Consider the system in~\eqref{1} with initial conditions given in~\eqref{initial2}. Suppose the trajectories of the infected proportion $I(t)$ and control input $u(t)$ are known for $t \in [0, k]$, and the basic reproduction number $\mathcal{R}_0$ is known. Then, there exists a unique solution $(\beta, \gamma, S(t), R(t))$ that satisfies~\eqref{1} over $[0, k]$.

\begin{proof}
From $\mathcal{R}_0 = \frac{\beta}{\gamma}$, we obtain $\gamma = \frac{\beta}{\mathcal{R}_0}$. Substituting it into \eqref{proof6} yields
\begin{equation}
\beta\! =\! -\frac{\mathcal{R}_0}{\beta  I(t)^2} \!
\left[ \!\frac{d^2 I(t)}{dt^2}\! + \!\beta I(t) \frac{dI(t)}{dt} \!-\! \frac{\left(\frac{dI(t)}{dt}\right)^2}{I(t)}\! \right] \!\!
-\! \mathcal{R}_0 u(t).
\label{fixed_point_beta}
\end{equation}
This forms a fixed-point equation for $\beta$, which admits a unique solution over $t \in [0, k]$ under the Lipschitz continuity condition with the trajectories of $I(t)$ and $u(t)$ known for $t \in [0, k]$. Then, based on the proof of Theorem~\ref{Theorem}, by the Picard–Lindel\"{o}f Theorem~\cite{hartman2002ordinary}, a unique solution of $S(t)$ and $R(t)$ exists locally for $t \in [0, k]$.
\end{proof}

\bibliographystyle{IEEEtran}
\bibliography{Ref}

\newpage
\setcounter{table}{0}
\setcounter{figure}{0}
\setcounter{equation}{0}
\setcounter{section}{0}
\setcounter{subsection}{0}

\renewcommand{\thetable}{S\arabic{table}}
\renewcommand{\thefigure}{S\arabic{figure}}
\renewcommand{\theequation}{S\arabic{equation}}
\renewcommand{\thesection}{S-\Roman{section}}
\renewcommand{\thesubsection}{S-\Roman{section}-\Alph{subsection}}

\section*{Supporting Material}
\addcontentsline{toc}{section}{Supporting Material} 

This supporting material provides a detailed description of the experimental settings, model configurations, and additional analyses for the proposed Physics-Informed Neural Networks (PINNs)-based Model Predictive Control (MPC) framework for $SIR$ epidemics.  Sec.~\ref{SM1} presents the complete experiment setup, including software–hardware configurations, neural network architectures, training strategies, and specific parameter choices for each case study in the main paper. Sec.~\ref{SM2} compares alternative neural network architectures within three PINNs-based MPC frameworks, highlighting their impact on estimation accuracy, noise robustness, and computational efficiency. Sec.~\ref{SM3} reports extensive parameter sensitivity analyses under varying epidemic conditions, noise levels, and parameter identifiability scenarios, thereby complementing and reinforcing the main experimental findings. Together, these materials offer transparency, reproducibility, and deeper insights into the performance and robustness of the proposed approaches.

\section{Experiment Setup}\label{SM1}
We implement the experiments in Python with TensorFlow $2.10.0$, Keras $2.10.0$, SciANN $0.7.0.1$, and CasADi $3.6.7$. We conduct all experiments on a single NVIDIA RTX $4090$ GPU paired with an AMD EPYC $7502$ CPU. The activation functions are sigmoids. We use an Adam optimizer to train the NN parameters. The LRE term in~(12) is scaled by $0.01$ to ensure its magnitude is comparable to the other terms in~(14), and the eigenvalues of NTK~\cite{wang2022and} dynamically adjust the weights of the loss function. During the PINNs training, we normalize the time variable $t$.

\textbf{The Iterative Approach}. We now introduce the iterative approach used in the experiments between the PINNs and MPC, where the procedure cycles between a PINNs initialization stage and an MPC control stage. We divide the experiment into two stages: a PINNs initialization stage and an MPC control stage. 
1) In the PINNs initialization stage, PINNs are trained at each selected time point $k_i$ using all available noisy infected states $\tilde{I}_i,i=0,\dots,k_i$ and control inputs $u_j,j=0,\dots,k_i-1$. PINNs then provide estimates of the current system states and parameters $\hat{S}(k_i)$, $\hat{I}(k_i)$, $\hat{R}(k_i)$, and $\hat{\beta}(k_i)$ (and $\hat{\gamma}(k_i)$). In addition, for evaluation and visualization purposes, PINNs also generate estimates of state and parameter over the interval $[k_{i-1}+1, k_i - 1]$ (with $k_0 = 0$). These intermediate estimates are not used in the MPC module, but are included in the figures to evaluate estimation performance. Only the estimates at time $k_i$ are passed to the MPC module to compute the control input $u(k_i)$.
2) In the control stage, at each selected time point $k_i$, the MPC module uses the estimated states and parameters provided by PINNs (from the initialization stage at $k_i$) as the current system condition. Using the system dynamics in~(1) and a predefined prediction horizon $N_{\p}$, the MPC module then constructs a prediction model in (5b) over $[k_i, k_i + N_{\p} - 1]$ and solves the optimization problem in~(5) to obtain the optimal control input $u(k_i)$. No additional PINNs training is performed during this stage.

\subsection{Experiment Setup for Section~IV-A}\label{Setup7.1}
For Sec.~IV-A, the transmission rate is $\beta=0.6$ and the recovery rate is $\gamma=0.2$. The total population is $N=1{,}000{,}000$, and the observed infected states are generated with Poisson noise as defined in~(3).
Initial conditions are $S_0=0.999$ and $I_0=0.001$.
MPC parameters include a prediction horizon $N_{\p}=14$ days, cost weights $q_1=10^3$ and $q_2=1$ (prioritizing $S^\star$), maximum infected proportion $I^\star_{\max}=0.1$, maximum control input $u_{\max}=0.4$, and target susceptible proportion $S^\star=\frac{1}{3}$.
The number of collocation points is $N_c=5000$. The initial physics-informed training uses $5000$ epochs, and the data regression stage of  the MPC-SI-PINNs is trained for $4500$ epochs. In subsequent iterations, both training phases use only one-fifth of their initial epochs to reduce computation cost.
Control begins at Day $10$ and ends at Day $40$. Each experiment is run independently ten times. We remove the maximum and minimum values from each set as outliers and then compute the confidence intervals within one standard deviation of the mean. Each neural network for $NN_{\hat S}$ and $NN_{\hat I}$ has four layers with $50$ neurons, while $NN_{\hat\beta}$ uses SciANN’s built-in functions.

\subsection{Experiment Setup for Section~IV-B}\label{Setup7.3}
For Sec.~IV-B, the transmission rate is $\beta=0.24$ and the recovery rate is $\gamma=0.12$. The initial susceptible and infected proportions are $S_0=0.999$ and $I_0=0.001$, respectively. The total population size $N=1{,}000{,}000$ is the same as in previous sections.
The control constraints include a maximum control input $u_{\max}=0.3$ and a maximum infected proportion $I^\star_{\max}=0.13$. The target susceptible proportion is $S^\star=1/2.5$. The control starts at day $45$ and ends at day $80$, with a prediction horizon $N_{\p}=14$ days. The cost function weights $q_1$ and $q_2$ are $10^3$ and $1$, respectively.
Regarding the neural network settings, $NN_{\hat S}$ uses four layers with $100$ neurons, $NN_{\hat I}$ and $NN_{\hat u}$ adopt four layers with $50$ neurons, and $NN_{\hat\gamma}$ uses SciANN’s built-in functions.
The initial physics-informed training uses $6000$ epochs, and the data regression stage is trained for $2500$ epochs. In subsequent iterations, both training phases use only one-fifth of their initial epochs to reduce computation cost. We sample $6000$ collocation points for enforcing the physics residuals.

\subsection{Experiment Setup for Section~IV-C} \label{Setup7.3}
For Sec.~IV-C, the transmission rate is $\beta = 0.24$ and the recovery rate is $\gamma = 0.12$. The initial susceptible and infected proportions are set as $S_0 = 0.999$ and $I_0 = 0.001$, respectively. The total population size $N=1,000,000$ is the same as in previous sections.  
The control constraints include a maximum control input $u_{\max} = 0.3$ and a maximum infected proportion $I^\star_{\max} = 0.13$. The target susceptible proportion is $S^\star = 1/2.5$. The control starts at day $45$ and ends at day $80$, with a prediction horizon $N_{\p} = 14$ days. The cost function weights $q_1$ and $q_2$ are $10^3$ and $1$, respectively. 

Regarding the neural network settings, the $NN_{\hat{S}}$ utilizes four layers with $100$ neurons, the $NN_{\hat{I}}$ and $NN_{\hat{u}}$ adopt four layers with $50$ neurons, and $NN_{\hat{\gamma}}$ uses SciANN’s built-in functions.
The initial physics-informed training uses $6000$ epochs, and the data regression stage is trained for $2500$ epochs. In subsequent iterations, both training phases use only one-fifth of their respective initial epochs to reduce computation cost. We sample $6000$ collocation points for enforcing the physics residuals.

\section{Comparison of Different NN Architectures for Section~III-A} \label{SM2}
This section evaluates the impact of different neural network architectures, as described in Sec.~III-A, under the three PINNs-based MPC frameworks, i.e., MPC-PINNs, MPC-LS-PINNs, and MPC-SI-PINNs. 
Specifically, we compare our proposed architecture from Sec.~III-A, which employs four separate SISO neural networks for estimations ($\hat{S}$, $\hat{I}$, $\hat{u}$, and $\hat{\beta}/\hat \gamma$), with three alternative designs:
\begin{enumerate}
\item \textbf{Architecture 1 (MPC-PINNs\_1):} This design, illustrated in Fig.~\ref{fig:NN_architecture_2}, uses three MISO  neural networks. However, each of these single-output networks takes both time~$t$ and the control input $u(t)$ as inputs to estimate each state and parameter. Unlike our separate NNs architecture, it does not employ a separate $NN_{\hat{u}}(t)$.  \label{Architecture_1} 
    \item \textbf{Architecture 2 (MPC-PINN\_2):} This approach, shown in Fig.~\ref{fig:NN_architecture_1}, utilizes a single SIMO neural network. This network takes only time $t$ as input and simultaneously generates all the estimates ($\hat{S}$, $\hat{I}$, $\hat{u}$, and $\hat{\beta}/\hat \gamma$) as its multiple outputs. \label{Architecture_2}
    \item \textbf{Architecture 3 (MPC-PINN\_3):} This approach, shown in Fig.~\ref{fig:NN_architecture_1}, utilizes a single MIMO neural network. This network takes both time $t$ and the control input $u(t)$ as inputs, and simultaneously generates all estimates ($\hat{S}$, $\hat{I}$, and $\hat{\beta}/\hat \gamma$). \label{Architecture_3}
\end{enumerate}
For the MPC-LS-PINNs and MPC-SI-PINNs, we similarly define MPC-LS-PINNs\_$1$ and MPC-SI-PINNs\_$1$ as the versions that adopt Architecture~\ref{Architecture_1} described above (i.e., using three single-output networks with both time $t$ and control $u$ as inputs, as in Fig.~\ref{fig:NN_architecture_2}).

\begin{figure}
    \centering
    \begin{subfigure}[b]{0.23\textwidth}
        \centering
        \includegraphics[width=\textwidth]{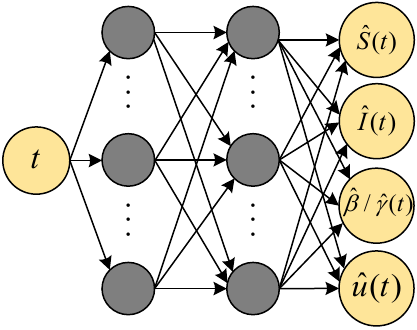}
        \caption{}
        \label{fig:NN_architecture_1}
    \end{subfigure}
    \hfill
    \begin{subfigure}[b]{0.23\textwidth}
        \centering
        \includegraphics[width=\textwidth]{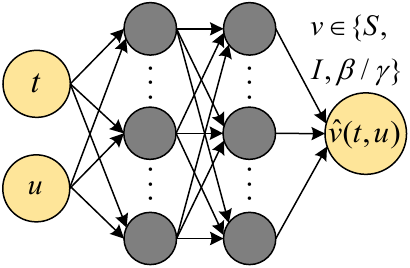}
        \caption{}
        \label{fig:NN_architecture_2}
    \end{subfigure}
    \hfill
    \begin{subfigure}[b]{0.23\textwidth}
        \centering
        \includegraphics[width=\textwidth]{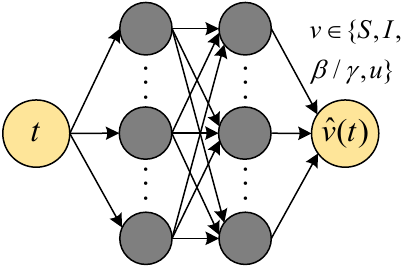}
        \caption{}
        \label{fig:NN_architecture_3}
    \end{subfigure}
    \hfill
    \begin{subfigure}[b]{0.23\textwidth}
        \centering
        \includegraphics[width=\textwidth]{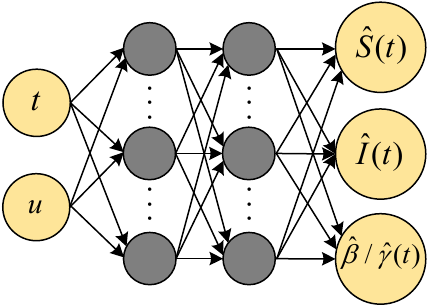}
        \caption{}
        \label{fig:NN_architecture_4}
    \end{subfigure}

    \caption{Four neural network architectures: 
    (a) A single SIMO neural network with input $t$;
    (b) Three MISO neural networks with inputs $t$ and $u$;
    (c) Our proposed four SISO $NN_{\hat{v}}(t)$, where $v\in\{S,I,\beta/\gamma,u\}$;
    (d) A single MIMO NN with inputs $t$ and $u$.}
    \label{fig:NN_architecture}
\end{figure}

\begin{table*}[!t]
\centering
\caption{Network architecture and training epochs for each method. Dimensions are given as [layers $\times$ neurons]. Empty entries are marked with ‘---’.}
\label{tab:architecture_setting}
\begin{tabular}{@{}lccccc@{}}
\toprule
\textbf{Method} & $NN_{\hat{S}}$ & $NN_{\hat{I}}$ & $NN_{\hat{u}}$ & Data Epoch & ODEs Epoch \\
\midrule
MPC-PINNs       & $4\times50$ & $4\times50$ & $4\times50$ & --- & $5000$ \\
MPC-PINNs\_$1$    & $4\times50$ & $4\times50$ & $4\times50$ & --- & $5000$ \\
MPC-PINN\_$2$& \multicolumn{3}{c}{$8\times150$}         & --- & $4000$ \\
MPC-PINN\_$3$ & \multicolumn{2}{c}{$8\times150$}& --- & --- & $5000$\\
\midrule
MPC-LS-PINNs    & $4\times50$ & $4\times50$ & $4\times50$ & --- & $5000$ \\
MPC-LS-PINNs\_$1$& $4\times50$ & $4\times50$ & ---    & --- & $5000$ \\
\midrule
MPC-SI-PINNs    & $4\times50$ & $4\times50$ & $4\times50$ & $4500$ & $5000$ \\
MPC-SI-PINNs\_$1$ & $4\times50$ & $4\times50$ & $4\times50$ & $4500$ & $5000$ \\
\bottomrule
\end{tabular}
\end{table*}

\begin{table*}[!t]
    \centering
    \caption{Performance comparison of different NN architectures. Columns report rMSE over $t\in[0,50]$ for each variable, the overall $L_2$ relative error, and runtime in seconds. }
    \label{tab:neural_networks_architecture}
    \begin{tabular}{c|lccccc c}
        \toprule
        $\kappa$ & \textbf{Method} & $\text{rMSE}_{\hat{S}}$ & $\text{rMSE}_{\hat{I}}$ & $\text{rMSE}_{\hat{R}}$ & $\text{rMSE}_{\hat{\beta}}$ & $L_2$ rel. error & Runtime (s) \\
        \midrule
        \multirow{4}{*}{$1$}
            & \textbf{MPC-PINNs} & $\mathbf{2.18e{-5}}$ & $6.10e{-5}$ & $\mathbf{3.49e{-5}}$ & $2.51e{-4}$ & $\mathbf{1.05e{-2}}$ & $1725.70$ \\
            & MPC-PINNs\_$1$     & $2.58e{-4}$ & $\mathbf{5.27e{-5}}$ & $4.24e{-4}$ & $\mathbf{7.43e{-5}}$ & $1.12e{-2}$ & $\mathbf{1581.07}$ \\
            & MPC-PINN\_$2$    & $5.55e{-4}$ & $2.53e{-4}$ & $9.01e{-4}$ & $5.48e{-3}$ & $4.23e{-2}$ & $2812.48$ \\
            & MPC-PINN\_$3$    & $7.61e{-2}$ & $2.34e{-2}$ & $9.01e{-1}$ & $2.34e{-1}$ & $3.53e{-1}$ & $3993.02$\\
        \midrule
        \multirow{2}{*}{$0.01$}
            & \textbf{MPC-LS-PINNs} & $\mathbf{2.80e{-4}}$ & $\mathbf{1.65e{-3}}$ & $\mathbf{5.07e{-4}}$ & $\mathbf{5.87e{-4}}$ & $\mathbf{2.13e{-2}}$ & $1673.62$ \\
            & MPC-LS-PINNs\_$1$     & $2.62e{-2}$ & $1.83e{-3}$ & $4.43e{-2}$ & $3.04e{-3}$ & $1.00e{-1}$ & $\mathbf{1568.75}$ \\
        \midrule
        \multirow{2}{*}{$0.01$}
            & \textbf{MPC-SI-PINNs} & $\mathbf{8.40e{-5}}$ & $\mathbf{1.28e{-3}}$ & $\mathbf{1.25e{-4}}$ & $\mathbf{7.38e{-7}}$ & $\mathbf{6.22e{-3}}$ & $1963.13$ \\
            & MPC-SI-PINNs\_$1$     & $1.64e{-3}$ & $1.34e{-3}$ & $2.66e{-3}$ & $1.47e{-3}$ & $4.09e{-2}$ & $\mathbf{1954.42}$ \\
        \bottomrule
    \end{tabular}%
\end{table*}

In this experiment, we employ an ideal MPC setup, where the control strategy is precomputed by solving the MPC problem using the true $SIR$ states and parameters over the entire time horizon. This ideal control input is directly applied to $SIR$ models without iterative interaction between the PINNs and MPC. Each method is trained using the complete set of noisy infected states and the ideal control trajectory across the full time interval. In other words, all methods receive identical full-sample training data and control inputs, to focus on the influence of the neural network architecture on the estimation accuracy and computational efficiency (runtime). 
The detailed network configurations for each architecture are listed in Table~\ref{tab:architecture_setting}. Each experiment is repeated independently three times, and the final evaluation is based on the average performance across all runs. For all architectures, $\hat{R}(t)$ is obtained as $1 - {\hat{S}} (t)- {\hat{I}}(t)$.

\begin{figure*}
    \centering
    \begin{subfigure}[t]{0.98\textwidth}
        \centering
        \includegraphics[width=\linewidth]{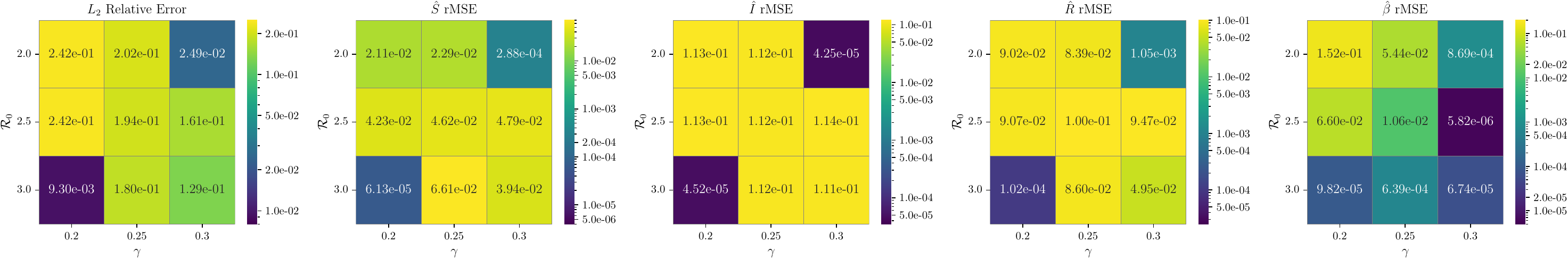}
        \caption{MPC-PINNs ($\kappa = 1$)}
        \label{fig:para_scene1_a}
    \end{subfigure}

    \vspace{0.01cm}

    \begin{subfigure}[t]{0.98\textwidth}
        \centering
        \includegraphics[width=\linewidth]{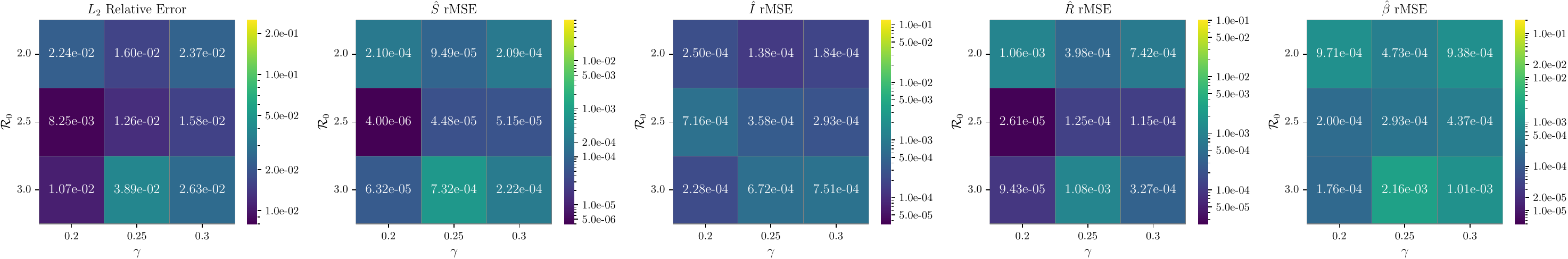}
        \caption{MPC-LS-PINNs ($\kappa = 1$)}
        \label{fig:para_scene1_b}
    \end{subfigure}

    \vspace{0.01cm}

    \begin{subfigure}[t]{0.98\textwidth}
        \centering
        \includegraphics[width=\linewidth]{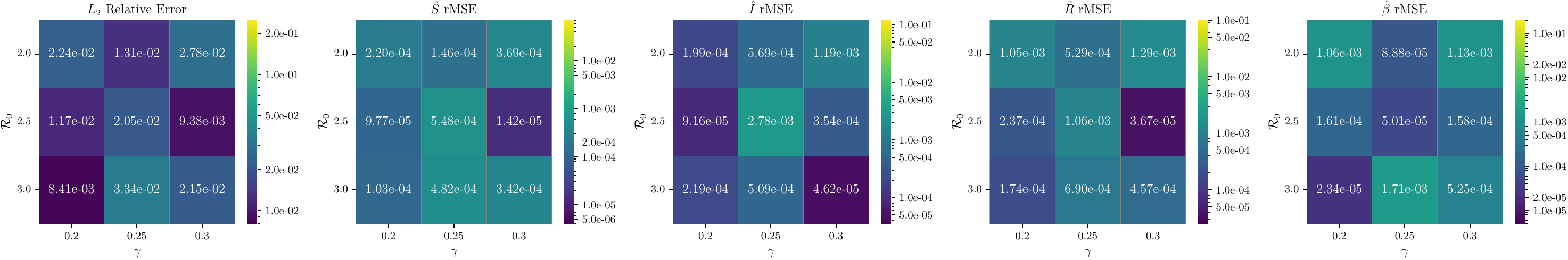}
        \caption{MPC-SI-PINNs ($\kappa = 1$)}
        \label{fig:para_scene1_c}
    \end{subfigure}

    \caption{Heatmaps of $L_2$ relative errors and rMSE for $\hat S$, $\hat I$, $\hat R$, and $\hat \beta$ across parameter pairs $(\gamma, \mathcal{R}_0)$ under noise with relatively low ($\kappa=1$) variance in~(3) with unknown $\beta$: (a) shows results for the MPC-PINNs, (b) for the MPC-LS-PINNs, and (c) for the MPC-SI-PINNs.}
    \label{fig:para_scene1}
\end{figure*}

Table~\ref{tab:neural_networks_architecture} presents the performance comparison of different neural network architectures under the MPC-PINNs, MPC-LS-PINNs, and MPC-SI-PINNs frameworks. The evaluation includes the rMSE of each variable, the overall $L_2$ relative error, and runtime.
For the MPC-PINNs under noise with relatively low ($\kappa=1$) variance, the architecture with separate single-input single-output neural networks achieves the lowest $L_2$ relative error ($1.05 \times 10^{-2}$) and the smallest rMSE for $\hat{S}(t)$ and $\hat{R}(t)$. In comparison, the MPC-PINNs\_$1$ and MPC-PINN\_$2$ exhibit higher $L_2$ relative errors ($1.12 \times 10^{-2}$ and $4.23 \times 10^{-2}$, respectively), while MPC-PINN\_$3$ shows a significantly larger $L_2$ relative error of $3.53 \times 10^{-1}$. MPC-PINN\_$3$ shows particularly reduced accuracy across all state estimates. Notably, the separate single-input single-output architecture (MPC-PINNs) also achieves lower runtime ($1725.70$ s) than MPC-PINN\_$2$ ($2812.48$ s) and MPC-PINN\_$3$ ($3993.02$ s), though marginally slower than MPC-PINNs\_$1$ ($1581.07$ s). These results indicate that a modular network design enhances estimation accuracy while maintaining computational efficiency.

Under noise with relatively high ($\kappa=0.01$) variance for the MPC-LS-PINNs, the separate architecture again achieves superior performance, with an $L_2$ relative error of $2.13 \times 10^{-2}$, substantially lower than the $1.00 \times 10^{-1}$ achieved by MPC-LS-PINNs\_$1$. This architecture also yields lower rMSE values across all variables, suggesting enhanced noise robustness and more accurate estimation.
Regarding computational aspects, the runtime for the MPC-LS-PINNs ($1673.62$\,s) is slightly higher than that of MPC-LS-PINNs\_$1$ ($1568.75$\,s). This modest increase in runtime reflects the added complexity of setting a separate $NN_{\hat{u}}$ for the control input, but the significant accuracy gains justify the additional computational cost.

Similarly, under noise with relatively high ($\kappa=0.01$) variance, the MPC-SI-PINNs outperforms the MPC-SI-PINNs\_$1$, achieving an $L_2$ relative error of $6.22\times 10^{-3}$ versus $4.09\times 10^{-2}$. The MPC-SI-PINNs also shows consistently lower rMSEs across $\hat{S}(t)$, $\hat{I}(t)$, $\hat{R}(t)$, and $\hat{\beta}(t)$, with particularly notable improvements in $\hat{S}(t)$ and $\hat{\beta}(t)$. The runtime difference between the MPC-SI-PINNs and the MPC-SI-PINNs\_$1$ is minimal ($1963.13$ s vs. $1954.42$ s), indicating that the separate SISO neural networks architecture achieves significant accuracy improvement without sacrificing computational efficiency.

These results demonstrate that using separate single-input single-output neural networks improves estimation accuracy across all PINNs-based frameworks, while maintaining acceptable computational overhead. This design enables each neural network to specialize in learning its target variable, avoiding potential interference among outputs that may arise in multi-output or multi-input architectures. The findings further validate the architectural choice introduced in Sec.~III-A.

\begin{figure*}[!t]
    \centering
    \begin{subfigure}[t]{0.98\textwidth}
        \centering
        \includegraphics[width=\linewidth]{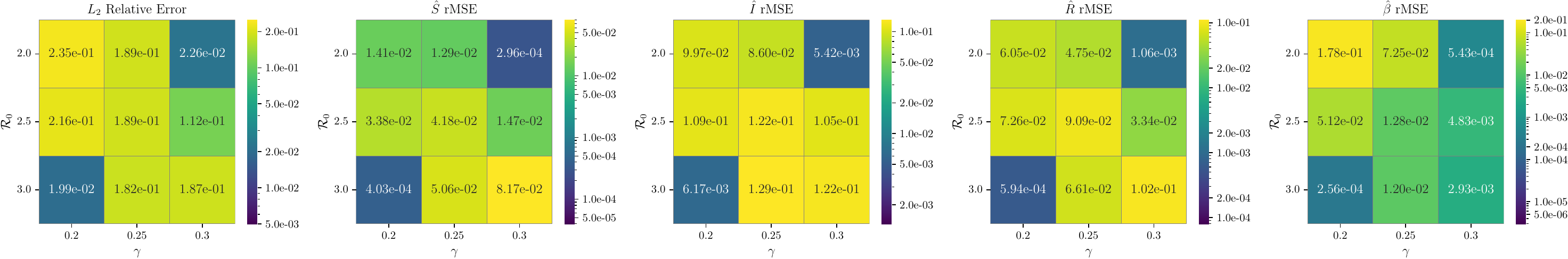}
        \caption{MPC-PINNs ($\kappa = 0.001$)}
        \label{fig:para_scene2_a}
    \end{subfigure}

    \vspace{0.01cm}

    \begin{subfigure}[t]{0.98\textwidth}
        \centering
        \includegraphics[width=\linewidth]{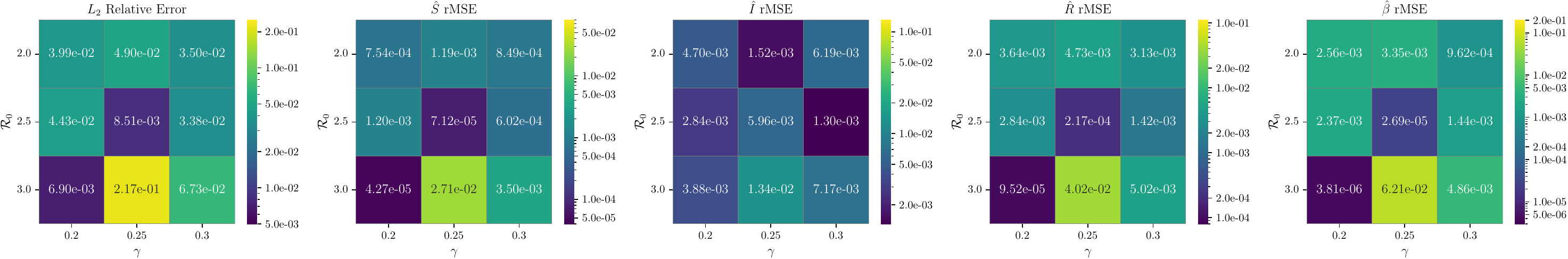}
        \caption{MPC-LS-PINNs ($\kappa = 0.001$)}
        \label{fig:para_scene2_b}
    \end{subfigure}

    \vspace{0.01cm}

    \begin{subfigure}[t]{0.98\textwidth}
        \centering
        \includegraphics[width=\linewidth]{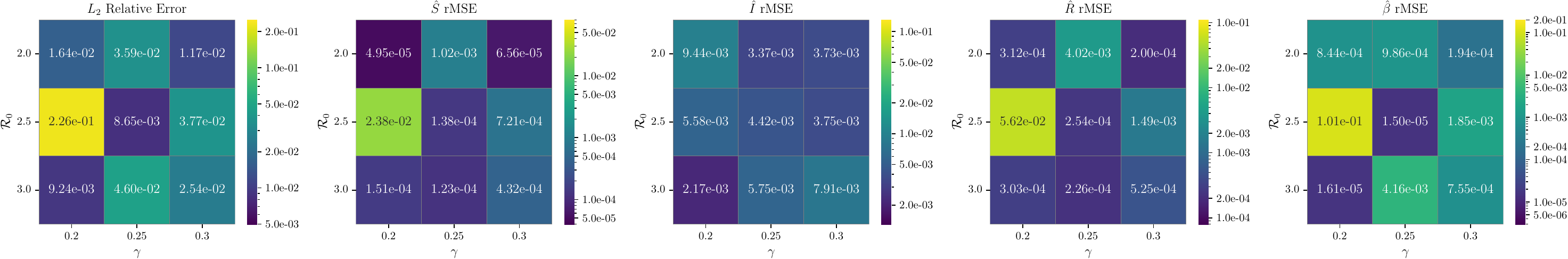}
        \caption{MPC-SI-PINNs ($\kappa = 0.001$)}
        \label{fig:para_scene2_c}
    \end{subfigure}

    \caption{Heatmaps of $L_2$ relative errors and rMSE for $\hat S$, $\hat I$, $\hat R$, and $\hat \beta$ across parameter pairs $(\gamma, \mathcal{R}_0)$ under noise with relatively high ($\kappa=0.001$) variance in~(3) with unknown $\beta$: (a) shows results for the MPC-PINNs, (b) for the MPC-LS-PINNs, and (c) for the MPC-SI-PINNs.}
    \label{fig:para_scene2}
\end{figure*}

\section{Parameter Sensitivity Analysis}\label{SM3}
To evaluate the MPC-PINNs, MPC-LS-PINNs, and MPC-SI-PINNs under varying epidemic conditions in Assumption~1, and their generalized versions with unknown $\beta$ and $\gamma$ in Assumption~2, we conduct experiments across different parameter combinations. Constrained by computational costs, we employ ideal MPC with all the observed infected states and control inputs, averaging results over three training runs. The details of the ideal MPC are presented in Sec.~3. We select the recovery rates $\gamma \in \{ 0.2,0.25, 0.3\}$ and basic reproduction numbers $\mathcal{R}_0 \in \{2.0, 2.5, 3.0\}$ with noise with relatively low ($\kappa=1$) and high ($\kappa=0.001$) variance in~(3). Increasing $\mathcal{R}_0$ reflects faster epidemic spread, while smaller $\gamma$ extends the duration of the epidemic. These combinations cover a broad range of realistic scenarios.

The $SIR$ models, MPC, and PINNs parameters largely follow Sec.~IV-A, with the different settings described as follows. The control start time (introducing control) is set to $t=0$, and the control end time (removing control) is defined as the final epidemic time minus $5$ days. This final epidemic time is determined adaptively according to the epidemic trajectory under the specific $SIR$ model parameters ($\gamma$ and $\mathcal{R}_0$), and corresponds approximately to the point when the epidemic has effectively ended (i.e., the number of infected individuals $I(t)$ has decayed close to zero and remains negligible thereafter). The rationale is: 1) to ensure that the training interval fully captures the entire dynamics of the epidemic without truncating it prematurely; and 2) to avoid unnecessarily extending the interval far beyond the epidemic’s conclusion, where $I(t)$ is already near zero and $S(t)$ and $R(t)$ evolve negligibly, which would dilute the focus on the dominant dynamics.
Final epidemic times and data regression epochs vary with parameter pairs, as summarized in Table~\ref{Parameter_setup}. The neural network $\hat{S}(t)$ uses four layers with $50$ neurons when only $\beta$ is unknown, and $100$ neurons when both $\beta$ and $\gamma$ are unknown. The $S^\star$ is set as $\frac{1}{\mathcal{R}_0}$ according to the values of basic reproduction numbers $\mathcal{R}_0 \in \{2.0, 2.5, 3.0\}$. 

\begin{table}
\centering
\caption{Final epidemic times, data regression epochs, and network settings under different $(\gamma,\mathcal{R}_0)$ pairs.}
\label{Parameter_setup}
\begin{tabular}{cccc}
\toprule
$\gamma$ & $\mathcal{R}_0$ & Final time $t$ & Regression epochs \\
\midrule
0.20 & 2.0 & 60 & 4500 \\
0.20 & 2.5 & 55 & 4500 \\
0.20 & 3.0 & 50 & 3000 \\
0.25 & 2.0 & 55 & 2500 \\
0.25 & 2.5 & 45 & 2500 \\
0.25 & 3.0 & 45 & 2500 \\
0.30 & 2.0 & 50 & 2500 \\
0.30 & 2.5 & 40 & 2500 \\
0.30 & 3.0 & 40 & 4500 \\
\bottomrule
\end{tabular}
\end{table}

\subsection{Parameter Sensitivity Analysis for Unknown $\beta$}
\begin{figure*}[!t]
    \centering
    \begin{subfigure}[t]{0.98\textwidth}
        \centering
        \includegraphics[width=\linewidth]{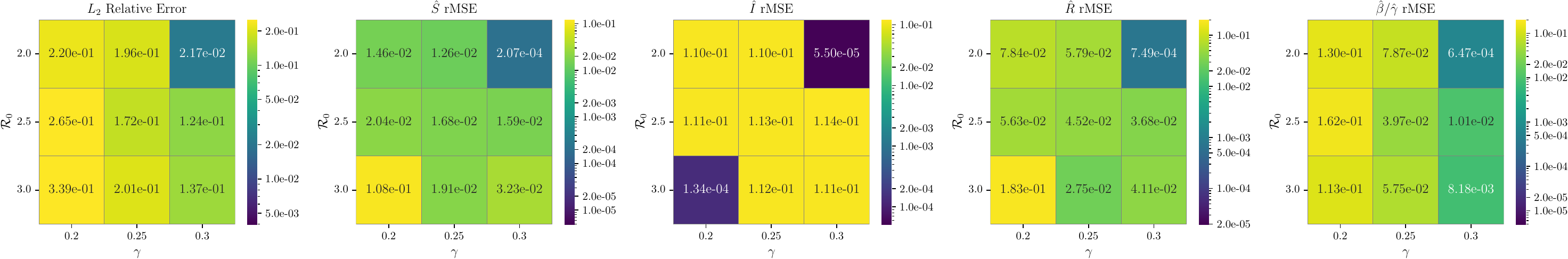}
        \caption{Generalized MPC-PINNs ($\kappa = 1$)}
        \label{fig:para_scene3_a}
    \end{subfigure}

    \vspace{0.01cm}

    \begin{subfigure}[t]{0.98\textwidth}
        \centering
        \includegraphics[width=\linewidth]{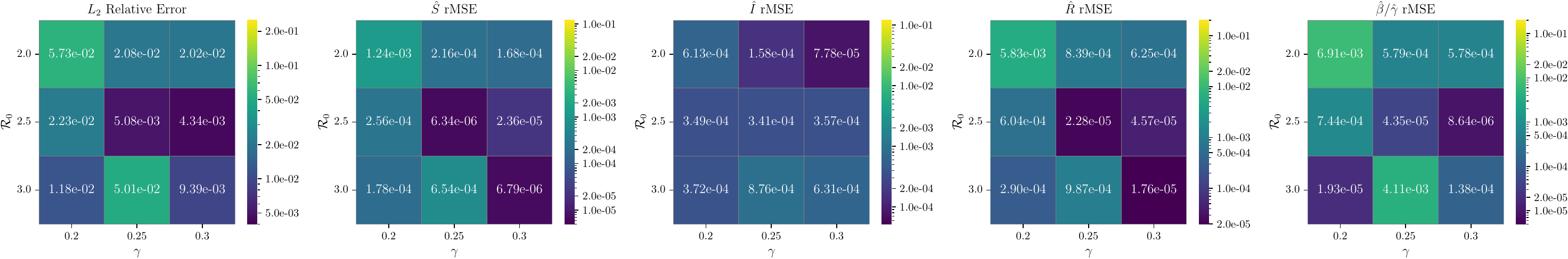}
        \caption{Generalized MPC-LS-PINNs ($\kappa = 1$)}
        \label{fig:para_scene3_b}
    \end{subfigure}

    \vspace{0.01cm}

    \begin{subfigure}[t]{0.98\textwidth}
        \centering
        \includegraphics[width=\linewidth]{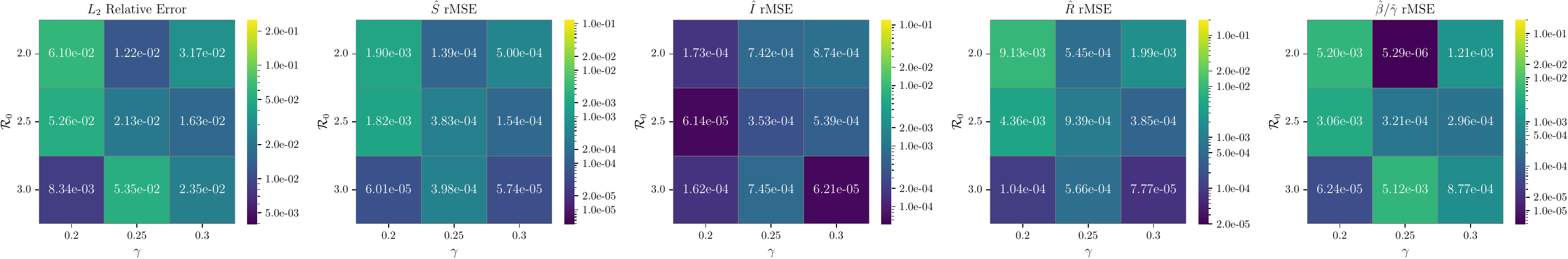}
        \caption{Generalized MPC-S-PINNs ($\kappa = 1$)}
        \label{fig:para_scene3_c}
    \end{subfigure}

    \caption{Heatmaps of $L_2$ relative errors and rMSE for $\hat S$, $\hat I$, $\hat R$, $\hat \beta$, and $\hat \gamma$ across parameter pairs $(\gamma, \mathcal{R}_0)$ under noise with relatively low ($\kappa=1$) variance in~(3) with unknown $\beta$ and $\gamma$: (a) shows results for the generalized MPC-PINNs, (b) for the generalized MPC-LS-PINNs, and (c) for the generalized MPC-S-PINNs.}
    \label{fig:para_scene3}
\end{figure*}

\begin{figure}[t]
    \centering
    \includegraphics[width=0.38\textwidth]{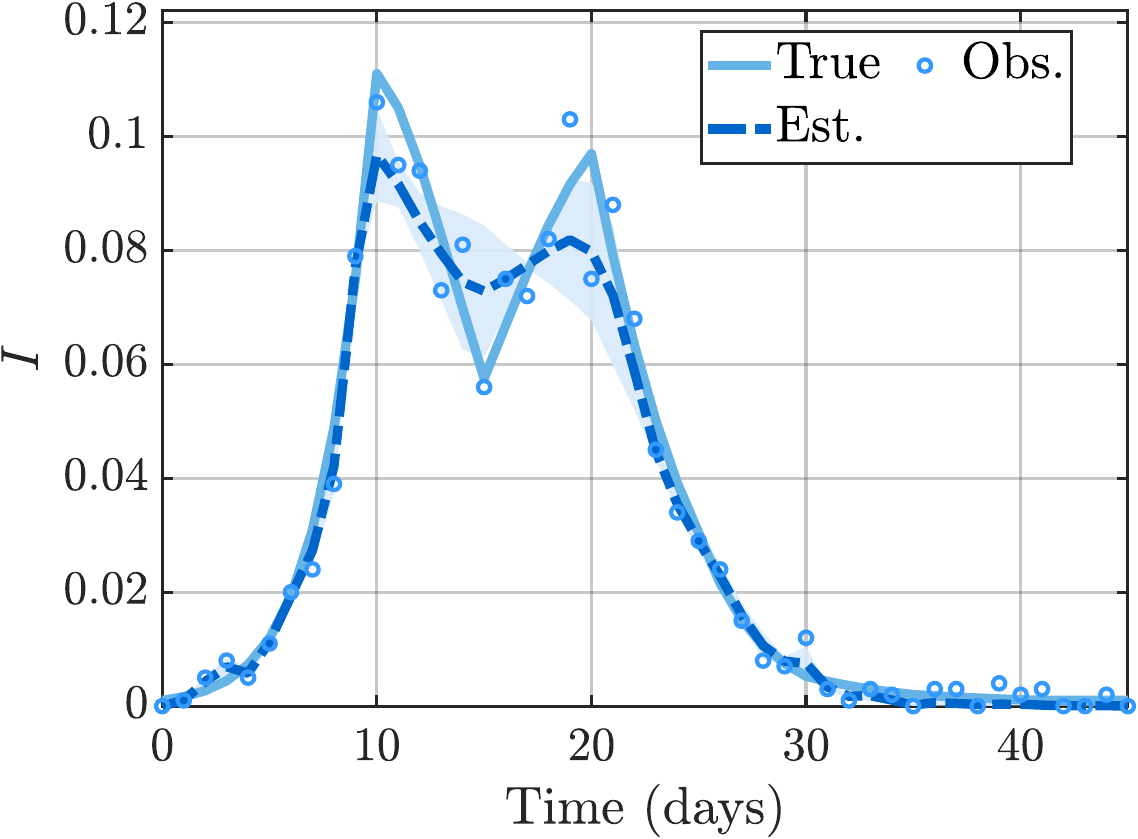}
    \caption{Infected state trajectory of the MPC-LS-PINNs at $(\gamma = 0.25, \mathcal{R}_0 = 3.0)$ under high-variance noise ($\kappa = 0.001$). 
    The trajectory exhibits a sharp double-peak with irregular noise, corresponding to the outlier reported in Fig.~\ref{fig:para_scene2}.}
    \label{fig:double_peak_case}
\end{figure}

Fig.~\ref{fig:para_scene1} compares the estimation performance of the MPC-PINNs, MPC-LS-PINNs, and MPC-SI-PINNs with $\kappa = 1$ and unknown $\beta$, across combinations of $\gamma \in \{0.2, 0.25, 0.3\}$ and $\mathcal{R}_0 \in \{2.0, 2.5, 3.0\}$. Each panel reports the $L_2$ relative error and rMSE values for $\hat{S}$, $\hat{I}$, $\hat{R}$, and $\hat{\beta}$.
The MPC-PINNs (Fig.~\ref{fig:para_scene1_a}) show large variability. In the first panel of Fig.~\ref{fig:para_scene1_a}, the $L_2$ relative error typically ranges from $1.29 \times 10^{-1}$ to $2.42 \times 10^{-1}$. Only two parameter settings lead to lower errors: $9.30 \times 10^{-3}$ at $(\gamma=0.2, \mathcal{R}_0=3.0)$ and $2.49 \times 10^{-2}$ at $(\gamma=0.3, \mathcal{R}_0=2.0)$. In other cases, large errors persist across all outputs.
The MPC-LS-PINNs (Fig.~\ref{fig:para_scene1_b}) perform in a more stable manner. In the first panel of Fig.~\ref{fig:para_scene1_b}, the $L_2$ errors range from $8.25 \times 10^{-3}$ to $3.89 \times 10^{-2}$. The second to fifth panels show that rMSE values for all outputs mostly lie between $10^{-5}$ and $10^{-4}$.
The MPC-SI-PINNs (Fig.~\ref{fig:para_scene1_c}) give the most consistent results. In the first panel of Fig.~\ref{fig:para_scene1_c}, the $L_2$ relative errors stay below $3.34 \times 10^{-2}$. The other panels show rMSE values between $10^{-5}$ and $10^{-4}$, with occasional increases up to $10^{-3}$. These results confirm the robustness and accuracy of MPC-SI-PINNs across different settings under noise with relatively low ($\kappa=1$) variance in~(3).

\begin{figure}[t]
    \centering
    \includegraphics[width=0.38\textwidth]{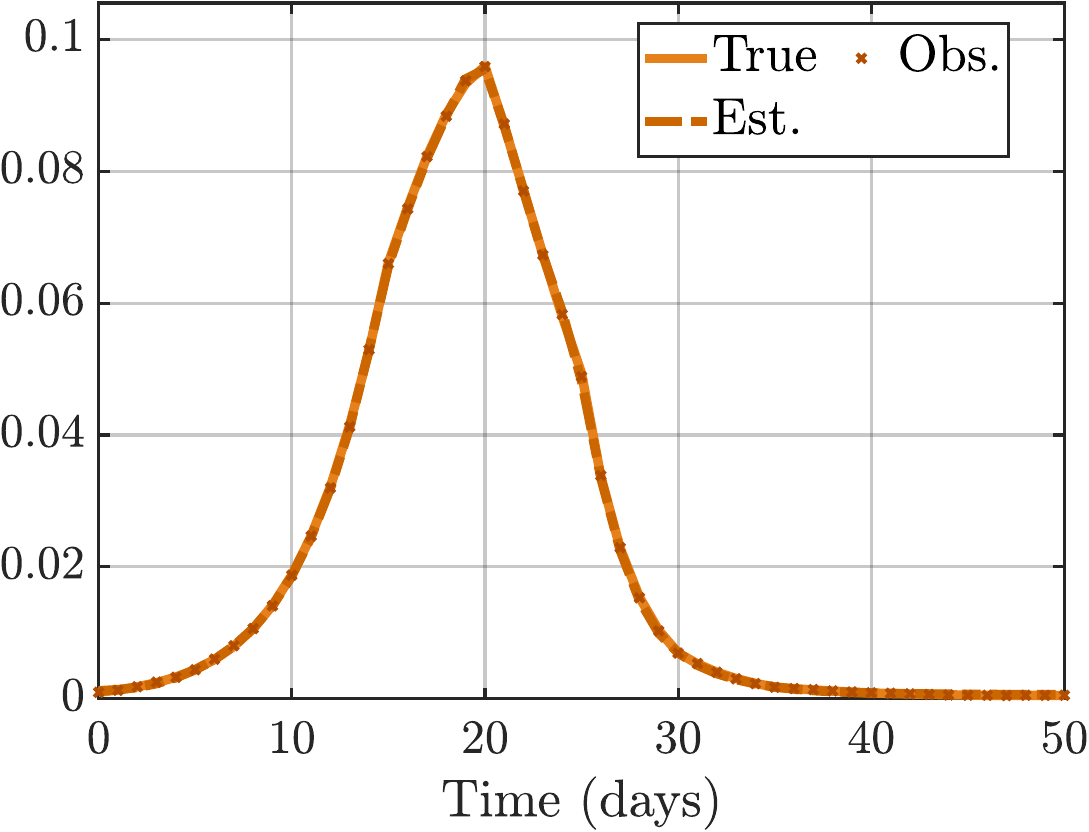}
    \caption{Infected state trajectory of the generalized MPC-PINNs at $(\gamma = 0.3, \mathcal{R}_0 = 2.0)$ under low-variance noise ($\kappa = 1$). 
    This trajectory exhibits a sharp single peak with a rapid rise and fall, which corresponds to the exceptionally low error reported in Fig.~\ref{fig:para_scene3}.}
    \label{fig:low_error_case}
\end{figure}

\begin{figure*}[!t]
    \centering
    \begin{subfigure}[t]{0.98\textwidth}
        \centering
        \includegraphics[width=\linewidth]{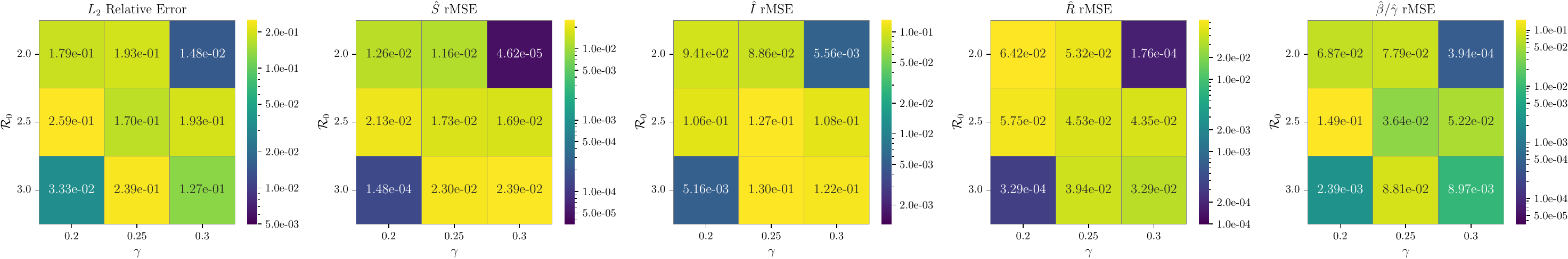}
        \caption{Generalized MPC-PINNs ($\kappa = 0.001$)}
        \label{fig:para_scene4_a}
    \end{subfigure}

    \vspace{0.01cm}

    \begin{subfigure}[t]{0.98\textwidth}
        \centering
        \includegraphics[width=\linewidth]{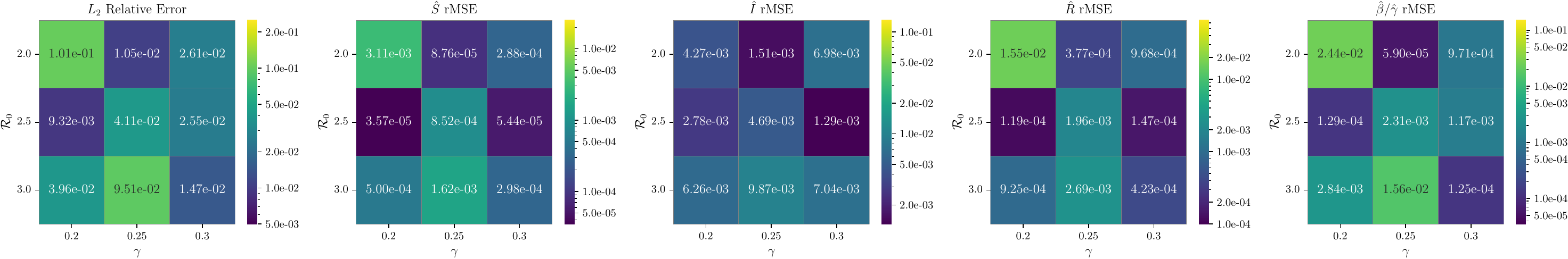}
        \caption{Generalized MPC-LS-PINNs ($\kappa = 0.001$)}
        \label{fig:para_scene4_b}
    \end{subfigure}

    \vspace{0.01cm}

    \begin{subfigure}[t]{0.98\textwidth}
        \centering
        \includegraphics[width=\linewidth]{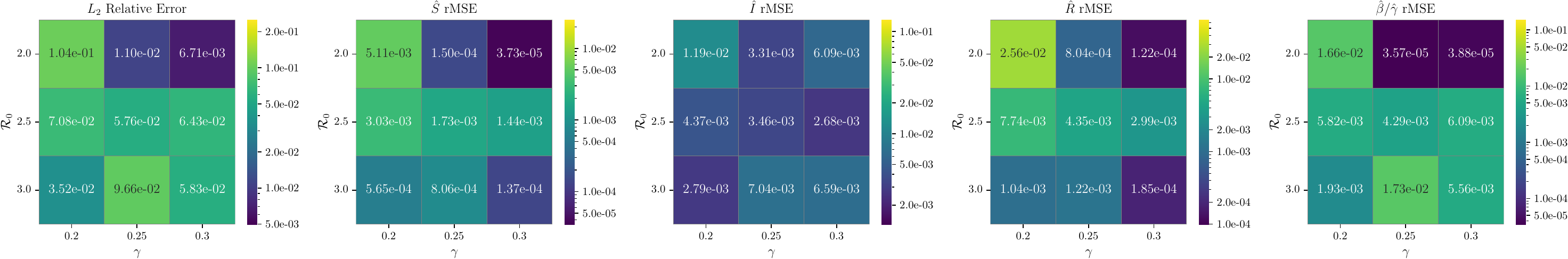}
        \caption{Generalized MPC-S-PINNs ($\kappa = 0.001$)}
        \label{fig:para_scene4_c}
    \end{subfigure}

    \caption{Heatmaps of $L_2$ relative errors and rMSE for $\hat S$, $\hat I$, $\hat R$, $\hat \beta$ and $\hat \gamma$ across parameter pairs $(\gamma, \mathcal{R}_0)$ under noise with relatively high ($\kappa=0.001$) variance in~(3) with unknown $\beta$ and $\gamma$: (a) shows results for the generalized MPC-PINNs, (b) for the generalized MPC-LS-PINNs, and (c) for the generalized MPC-S-PINNs.}
    \label{fig:para_scene4}
\end{figure*}

Fig.~\ref{fig:para_scene2} shows the heatmaps of estimation performance for the MPC-PINNs, MPC-LS-PINNs, and MPC-SI-PINNs with $\kappa = 0.001$ and unknown $\beta$. Rows represent different values of $\mathcal{R}_0 \in \{2.0, 2.5, 3.0\}$, and columns correspond to $\gamma \in \{0.2, 0.25, 0.3\}$.
The MPC-PINNs (Fig.~\ref{fig:para_scene2_a}) show degraded performance under this noise level. In the first panel, the $L_2$ relative errors range from $1.99 \times 10^{-2}$ to $2.35 \times 10^{-1}$. The second to fourth panels reveal large rMSE values for the states, often exceeding $10^{-2}$. Only two parameter combinations yield visibly lower errors. In the fifth panel, the rMSE for $\hat{\beta}$ varies from $2.56 \times 10^{-4}$ to $1.78 \times 10^{-1}$.
The MPC-LS-PINNs (Fig.~\ref{fig:para_scene2_b}) perform in a more stable manner. Most $L_2$ relative errors in the first panel fall between $6.9 \times 10^{-3}$ and $6.73 \times 10^{-2}$, except for one outlier $2.17 \times 10^{-1}$ at $(\gamma = 0.25, \mathcal{R}_0 = 3.0)$.  The rMSE values in the remaining panels are mostly within $10^{-5}$ to $10^{-3}$, showing consistent accuracy. The outlier at $(\gamma = 0.25, \mathcal{R}_0 = 3.0)$ arises because the infected trajectory at this parameter pair exhibits a sharp double-peak with irregular noise (see Fig.~\ref{fig:double_peak_case}), which hinders the joint training of states and parameters in MPC-PINNs and MPC-LS-PINNs. By contrast, MPC-SI-PINNs first split the training for $NN_{\hat I}$ to smooth the infected states $\tilde I$ and reduce the coordination burden, and then employ integral-derived $S^{\text{d}}$ and $R^{\text{d}}$ as anchors, enabling reliable estimation despite the challenging data.
The MPC-SI-PINNs (Fig.~\ref{fig:para_scene2_c}) achieve similar performance. The first panel shows most $L_2$ relative errors between $8.65 \times 10^{-3}$ and $4.60 \times 10^{-2}$, with one exception at $(\gamma = 0.2, \mathcal{R}_0 = 2.5)$ where the error reaches $2.26 \times 10^{-1}$. This exception is mainly attributed to the late-stage error propagation in the integral operation of the MPC-SI-PINNs. In this case, inaccuracies in $\hat{I}(t)$ accumulate in the derived states $S^{\mathrm d}$ and $R^{\mathrm d}$, defined in (16), leading to degraded performance. The second to fifth panels report rMSE values mostly within $10^{-5}$ to $10^{-3}$, comparable to MPC-LS-PINNs. These results confirm that both MPC-LS-PINNs and MPC-SI-PINNs maintain high robustness, even with high levels of noise ($\kappa=0.001$ in~(3)).

In summary, the MPC-SI-PINNs and MPC-LS-PINNs demonstrate strong robustness against different levels of noise and different parameters of $SIR$ models, with overall estimation accuracy significantly better than that of MPC-PINNs. The performance gap between the MPC-SI-PINNs and the MPC-LS-PINNs is not substantial, and both show complementary advantages depending on the parameter combination. Note that the results in Fig.~\ref{fig:para_scene1} and Fig.~\ref{fig:para_scene2} are obtained using all the noisy infected states and control inputs from an ideal MPC. Therefore, the nature of  conclusions reported here differs from those in Sec.~IV-A, where model performance is evaluated in interaction with the MPC controller.

\subsection{Parameter Sensitivity Analysis for Unknown $\beta$ and $\gamma$} 
Fig.~\ref{fig:para_scene3} shows the heatmaps for the generalized MPC-PINNs, MPC-LS-PINNs, and MPC-S-PINNs when both $\beta$ and $\gamma$ are unknown.
The generalized MPC-PINNs (Fig.~\ref{fig:para_scene3_a}) perform poorly. In the first panel, most $L_2$ relative errors range from $1.24 \times 10^{-1}$ to $3.39 \times 10^{-1}$, except one low value $2.17 \times 10^{-2}$ at $(\gamma = 0.3, \mathcal{R}_0 = 2.0)$. This exceptionally low error may be due to the infected state trajectory exhibiting a single, sharp peak with a rapid rise and fall, which is relatively simple (see Fig.~\ref{fig:low_error_case}). By contrast, the other parameter settings produce relatively more complex double-peak or plateau-like patterns, with the latter arising from control actions that keep the infected state trajectory close to but not exceeding the infection peak constraint $I^\star_{\max}$. The second to fifth panels show rMSE values for $\hat{S}$, $\hat{I}$, $\hat{R}$, $\hat{\beta}$, and $\hat{\gamma}$ between $10^{-2}$ and $10^{-1}$. These results confirm limited estimation ability under low noise ($\kappa = 1$).
The generalized MPC-LS-PINNs (Fig.~\ref{fig:para_scene3_b}) achieve the best performance. The first panel shows $L_2$ relative errors between $4.34 \times 10^{-3}$ and $5.73 \times 10^{-2}$. The other panels report rMSE values mostly within $10^{-6}$ to $10^{-4}$, with a few reaching $10^{-3}$. In seven out of nine parameter combinations, the generalized MPC-LS-PINNs outperform the generalized MPC-S-PINNs in $L_2$ error.
The generalized MPC-S-PINNs (Fig.~\ref{fig:para_scene3_c}) rank second. Their $L_2$ relative errors range from $8.34 \times 10^{-3}$ to $6.10 \times 10^{-2}$ (first panel), and rMSE values fall between $10^{-6}$ and $10^{-3}$ in the remaining panels. Compared to the generalized MPC-PINNs, both the generalized MPC-LS-PINNs and MPC-S-PINNs reduce $L_2$ errors by one to two orders of magnitude in most cases.

Fig.~\ref{fig:para_scene4} shows the heatmaps for the generalized MPC-PINNs, MPC-LS-PINNs, and MPC-S-PINNs under high-variance noise ($\kappa = 0.001$) when both $\beta$ and $\gamma$ are unknown.
The generalized MPC-PINNs (Fig.~\ref{fig:para_scene4_a}) perform poorly. The $L_2$ relative errors in the first panel range from $1.48 \times 10^{-2}$ to $2.59 \times 10^{-1}$. The second to fifth panels show rMSE values for all outputs between $10^{-4}$ and $10^{-1}$, with large variations across the parameter settings.
The generalized MPC-LS-PINNs (Fig.~\ref{fig:para_scene4_b}) show notable improvement. The first panel reports $L_2$ errors mostly between $9.32 \times 10^{-3}$ and $1.01 \times 10^{-1}$. The rMSE values in other panels fall within $10^{-5}$ to $10^{-3}$, consistently outperforming the MPC-PINNs by one to two orders of magnitude.
The generalized MPC-S-PINNs (Fig.~\ref{fig:para_scene4_c}) deliver comparable or slightly better results. The $L_2$ relative errors stay within $6.71 \times 10^{-3}$ to $9.66 \times 10^{-2}$, except one flawed case reaching $1.04 \times 10^{-1}$ at $(\gamma = 0.2, \mathcal{R}_0 = 2.0)$. The exception occurs mainly because the training data at this parameter setting happen to be of particularly poor quality (this case can be directly reproduced using the corresponding random seed provided in the released code). Specifically, the infected trajectory stays on a plateau near the epidemic peak (see Fig.~\ref{fig:poor_seed_case}), where the noise is both large and asymmetric. As a result, the generalized MPC-PINNs, MPC-LS-PINNs, and MPC-S-PINNs yield less reliable estimates in this case. The MPC-S-PINNs' rMSE values are similar to those of the generalized MPC-LS-PINNs. 
Overall, both the generalized MPC-LS-PINNs and MPC-S-PINNs achieve robust and accurate estimation under high noise. The MPC-LS-PINNs yield the lowest $L_2$ errors in seven out of nine parameter combinations. Compared to the MPC-PINNs, their errors are lower in eight cases and reduced by $44\%$ to $96\%$ in the seven best-performing ones.

\begin{figure}[t]
    \centering
    \includegraphics[width=0.38\textwidth]{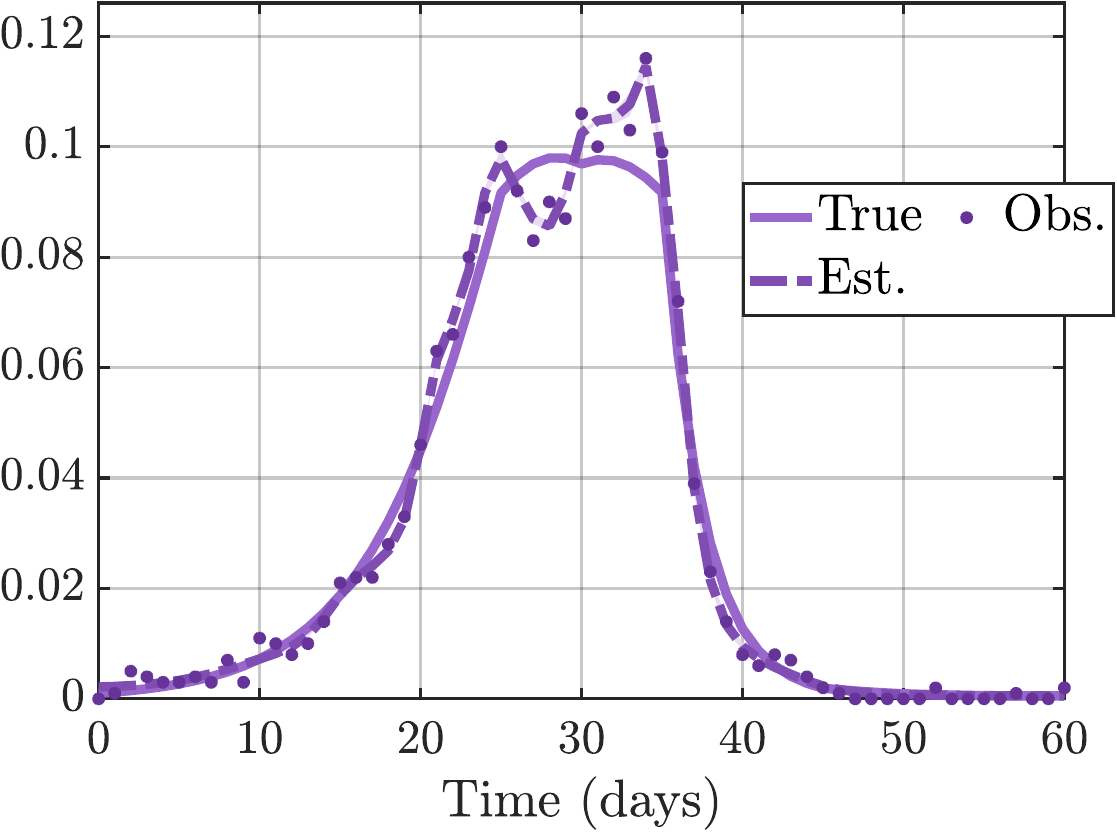}
   \caption{Infected state trajectory of the generalized MPC-S-PINNs at $(\gamma = 0.2, \mathcal{R}_0 = 2.0)$ under high-variance noise ($\kappa = 0.001$). 
    The trajectory corresponds to a poor-quality dataset generated from a random seed, leading to the flawed case in Fig.~\ref{fig:para_scene4}.}
    \label{fig:poor_seed_case}
\end{figure}

In summary, when both the transmission rate $\beta$ and recovery rate $\gamma$ are unknown, the generalized MPC-LS-PINNs and the generalized MPC-S-PINNs consistently and significantly outperform the generalized MPC-PINNs across both noise levels with relatively low ($\kappa=1$, Fig.~\ref{fig:para_scene3}) and high ($\kappa=0.001$, Fig.~\ref{fig:para_scene4}) variance. These results highlight the enhanced robustness and accuracy of the proposed the generalized MPC-LS-PINNs and the generalized MPC-S-PINNs against different levels of noise and different parameters of $SIR$ models. 
Note that the results in Fig.~\ref{fig:para_scene3} and Fig.~\ref{fig:para_scene4} are derived using all the noisy infected states and control inputs from an ideal MPC. 
Therefore, the nature of the conclusions reported here differs from those in Sec.~IV-C, which evaluate the model performance within the MPC framework.

\end{document}